\documentclass[11pt]{article}

\usepackage[T1]{fontenc}
\usepackage{lmodern,amsmath,amssymb,graphicx,booktabs,microtype}
\usepackage[margin=0.85in]{geometry}
\usepackage[font=small,labelfont=bf]{caption}
\usepackage[hidelinks]{hyperref}
\usepackage{placeins,xcolor}
\usepackage{ulem}

\newcommand{\fig}[4]{\begin{figure}[!tbp]\centering
\includegraphics[width=\linewidth,height=#1\textheight,keepaspectratio]{figures/#2.png}
\caption{#4}\label{#3}\end{figure}}
\newcommand{\E}{\mathcal E}

\title{Generative Energy-Based Modeling of Binary Data with Hypernetworks}

\author{\parbox{0.95\textwidth}{\centering
Carpio, N. A. C.\textsuperscript{1}\thanks{\href{mailto:ncarpio@estudante.ufscar.br}{\nolinkurl{ncarpio@estudante.ufscar.br}}},
Paes, A.\textsuperscript{2},
Fernandes, G. P. L. M.\textsuperscript{1,3},\\
Villas-Boas, C. J.\textsuperscript{1},
Farias, T. S.\textsuperscript{1,4}\\[0.62em]
{\small
\textsuperscript{1}Department of Computer Science, Federal University of S\~ao Carlos, S\~ao Carlos, SP, Brazil\\
\textsuperscript{2}Department of Information Science, University of North Texas, Denton, TX, USA\\
\textsuperscript{3}Wernher von Braun Advanced Research Center, Campinas, SP, Brazil\\
\textsuperscript{4}Department of Physics, University of Innsbruck, Innsbruck, Austria
}
}}

\date{}
\begin{document}
\maketitle
\begin{abstract}
Binary image codes provide a common space in which to learn an energy for generation and inference. We introduce HyperIsing, a model in which a hypernetwork computes spin fields and couplings from the configuration being scored. This dependence allows for effective interactions beyond fixed pairs. We train the energy by ranking data codes against corrupted alternatives and, for labeled data, by comparing the energies assigned to candidate classes. On MNIST, the same fitted energy supports classification, class-conditional generation,  joint image restoration and label inference. With the decoder and the number of latent proposals held fixed, search based on observed-pixel fidelity alone gives lower reconstruction error; joint inference returns a class estimate alongside the image. For unconditional cat-image generation, the method with the lowest feature likelihood divergence changes with the reference: pixel diffusion ranks first against original images, whereas HyperIsing ranks first against autoencoder reconstructions at settings selected on the evaluated grid. These results distinguish matching the distribution of decoded data from matching the original image distribution.
\end{abstract}

\section{Introduction}

Energy-based models assign a scalar energy to a configuration of variables, with lower energies corresponding to more favorable configurations \cite{ebm}. The variables can represent the observed part of a problem, unobserved variables, and candidate outputs; the energy scores these quantities jointly or conditionally. Inference compares candidate assignments while holding the observations fixed \cite{ebm}. In a probabilistic model with Boltzmann weights, energy differences also determine the relative probability of configurations at a fixed temperature \cite{ackley}. Learning the energy therefore specifies both which assignments are preferred for a given observation and, under Boltzmann weighting, the target distribution at a given temperature.

Early image models made this connection explicit. Geman and Geman combined a Gibbs prior over images with a model of the degradation process to infer the original image from a corrupted observation \cite{geman}. Neural energy models learn the energy function itself from data. The Generative ConvNet uses a convolutional network to define a generative random field over images \cite{xie}. Du and Mordatch studied neural energies for image synthesis and completion \cite{du}. Grathwohl et al.\ use classifier logits to define a joint energy over images and labels, allowing the same model to infer the label of a fixed image or generate images at a fixed label \cite{grathwohl}. These studies establish different ways to connect learning, generation, and inference through energy scores.

Many latent generative models learn a distribution over image representations and use a decoder to map samples back to image space \cite{ldm,vqvae}. The autoencoder learns a representation that retains information needed for reconstruction. The generative model then learns a distribution over those representations. Latent Diffusion performs denoising in the continuous, lower-dimensional space of a pretrained autoencoder \cite{ldm}. VQ-VAE instead represents images by discrete codebook indices and learns a distribution over those indices \cite{vqvae}. Binary codes are a special case of discrete representations, with two possible values per component. Binary Latent Diffusion learns such codes with an autoencoder and models their distribution with a diffusion process \cite{binarydiff}. HyperIsing uses a binary latent representation to match the two possible states of each Ising spin. This allows each code bit to be mapped directly to one spin.

In HyperIsing, the autoencoder represents each image with fewer variables than the original pixels, so inference and sampling operate on a smaller set of variables. The decoder maps the inferred or sampled codes back to image space (Appendix~\ref{app:binary-ae}). If we sampled the exact distribution of encoded data and used the same decoder, the resulting images would follow the distribution of autoencoder reconstructions. We therefore compare the generated images with both reconstructions and original images. The first comparison measures agreement with what the decoder produces from data codes; the second measures agreement with the original image distribution.

The choice of energy determines which dependencies can be expressed on the code. An Ising energy with fixed fields and pair couplings has degree at most two in its spin variables \cite{ising,decelle}. Higher-order visible interactions can arise by marginalizing hidden spins from a restricted Boltzmann machine \cite{decelle}. In factored three-way Boltzmann machines, multiplicative interactions let hidden units modulate local covariances in natural images \cite{ranzato}. Here we introduce \textit{HyperIsing}, a state-dependent spin energy. A hypernetwork \cite{ha} maps each candidate configuration to the fields and couplings used to score it. A pair coefficient can therefore depend on spins outside that pair, making higher-order terms possible in the scalar energy. We learn this energy by comparing data codes with their corrupted version and, when labels are available, by comparing candidate classes for the same code.

We evaluate the learned energy in two settings. On MNIST \cite{mnist}, a single fitted energy per model supports classification, class-conditional generation, and joint restoration and label inference. These tasks examine the same energy under different constraints on the image code and the label. On unlabeled images from the archived \emph{Cats Faces 64$\times$64} collection \cite{catdata}, we study generation across code lengths and sampling settings. The comparison with original images, autoencoder reconstructions, and held-out codes makes it possible to evaluate agreement with distinct reference distributions. Together, the two studies characterize how a state-dependent spin energy can be used for generation and inference, and how its generative performance depends on the representation, the sampling settings, and the reference used for evaluation.

The remainder of the paper is organized as follows. Section~\ref{sec:method} defines the spin representation, the hypernetwork, and the objectives for training and inference, and describes the experimental protocols. Section~\ref{sec:results} presents and discusses the MNIST and cat-image results. Section~\ref{sec:conclusions} summarizes the findings. The appendices give the mathematical details, evaluation protocols, and supplementary results.

\section{Methods}

In this section, we describe how HyperIsing learns a state-dependent energy over binary image codes and uses it for generation and inference. We first define the spin representation of images and class labels, where available, and introduce the energy function. We then describe the hypernetwork that predicts the fields and couplings for each spin configuration and the objectives used to train it. Finally, we explain how the learned energy is used to sample image codes, infer class labels, and restore partially observed images, and specify the experimental protocols for MNIST and the cat-image dataset.

\label{sec:method}
\subsection{Binary Representation and State-Dependent Spin Energy}

The HyperIsing energy is defined on configurations of two-state spins. To apply it to observations that are not binary, we first map each observation to a binary code. A binary autoencoder provides one way to learn this mapping from data \cite{binarydiff}. For images, we train the encoder to map each input to a code $b\in\{0,1\}^n$ and the decoder $D$ to reconstruct the image from that code. We then use $s=2b-1$ as the spin representation. The encoder and decoder are trained independently of the energy model and kept fixed during energy fitting. Appendix~\ref{app:binary-ae} describes the binary quantizer and the dataset-specific autoencoders.

\begin{figure}[!tbp]\centering
\begingroup\setlength{\unitlength}{1pt}
\definecolor{workflowpurple}{HTML}{734381}
\definecolor{workflowfill}{HTML}{F5F0F8}
\definecolor{workflowblue}{HTML}{0C719F}
\resizebox{!}{.52\textheight}{%
\begin{picture}(2175,1575)
\put(0,0){\includegraphics[width=2175pt,height=1575pt]{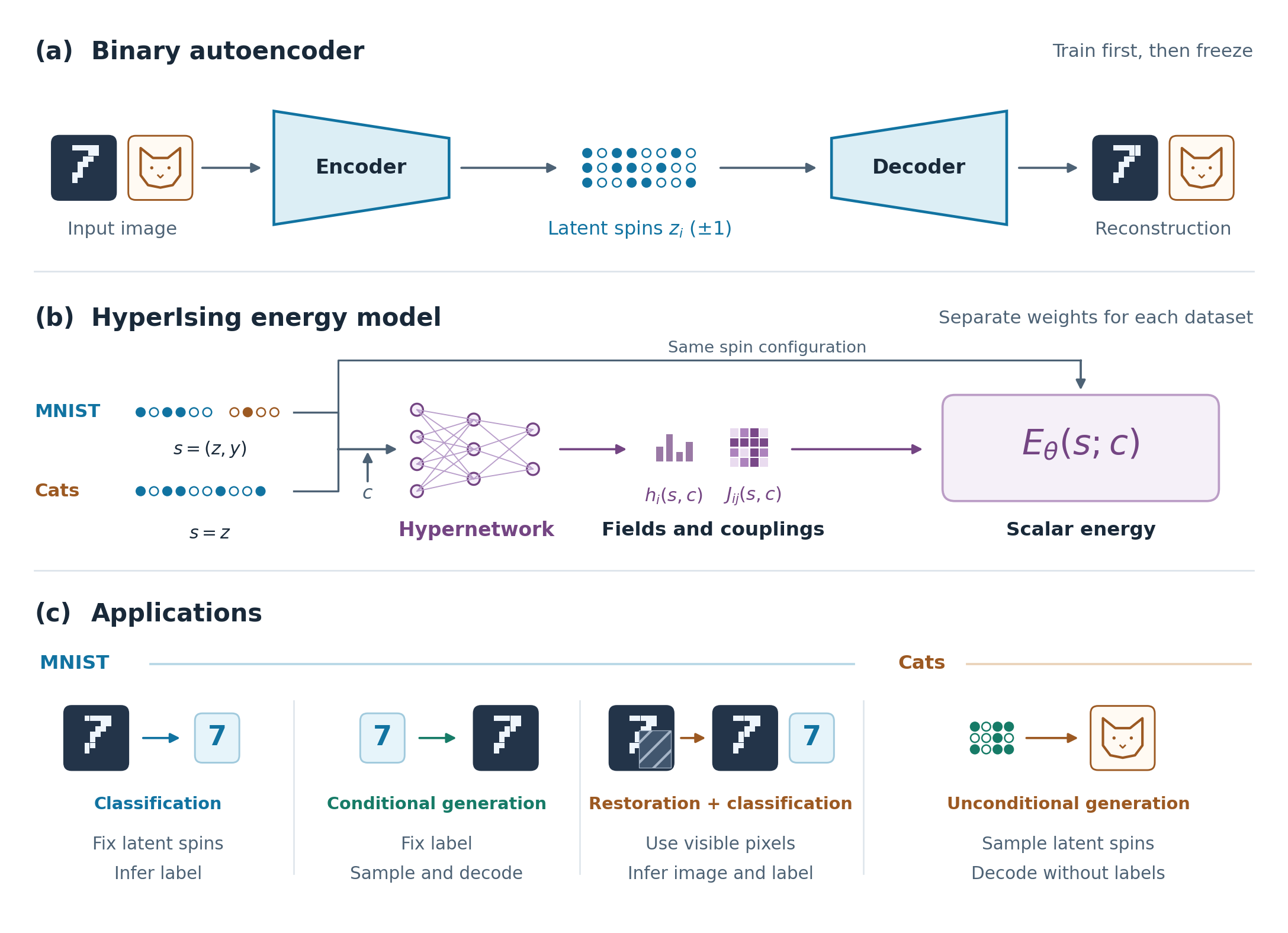}}
\put(900,1168){\color{white}\rule{365pt}{50pt}}
\put(1082,1190){\makebox(0,0){\color{workflowblue}\fontsize{30}{36}\selectfont\sffamily Binary code \(b\in\{0,1\}^n\)}}
\put(265,796){\color{white}\rule{190pt}{51pt}}
\put(355,820){\makebox(0,0){\fontsize{32}{38}\selectfont \(z=(s,\ell(y))\)}}
\put(268,657){\color{white}\rule{180pt}{48pt}}
\put(355,679){\makebox(0,0){\fontsize{32}{38}\selectfont \(z=s\)}}
\put(1060,711){\color{white}\rule{310pt}{51pt}}
\put(1201,737){\makebox(0,0){\color{workflowpurple}\fontsize{32}{38}\selectfont \(h_i(z,c)\quad J_{ij}(z,c)\)}}
\put(1635,765){\color{workflowfill}\rule{370pt}{104pt}}
\put(1825,820){\makebox(0,0){\color{workflowpurple}\fontsize{63}{72}\selectfont \(\E_\theta(z;c)\)}}
\end{picture}}
\endgroup
\caption{HyperIsing workflow. A frozen dataset-specific autoencoder maps images to bits \(b\) and decodes those bits. The energy uses spins \(s=2b-1\), with the label spins appended on MNIST. For generated or restored image codes, the inverse map \(b=(s+1)/2\) provides the decoder input. The hypernetwork computes fields and couplings from the candidate state and its corruption condition. One trained MNIST energy supports classification, conditional generation, and joint restoration and labeling. The cat dataset uses no label spins.}
\label{fig:workflow}
\end{figure}

Additional discrete information can be included as spin values. For example, in conditional generation, we have access to labels, in which can guide the model to generate data of a specific class. For MNIST, $y\in\{0,\ldots,9\}$ denotes a scalar class label (true or candidate), and $\ell(y)=2e_y-\boldsymbol 1$ denotes its ten-spin representation, where $e_y$ is the one-hot vector for class $y$. The joint state is $z=(s,\ell(y))$: the spin at class $y$ is $+1$ and the other nine are $-1$. In expressions such as $\E_\theta(s,y;c)$, the scalar label is shorthand for evaluating the energy at $z=(s,\ell(y))$. For unlabeled cat images, the state contains only the image-code spins, $z=s$. Given $z$ and a scalar corruption condition $c$, the hypernetwork predicts local fields $h_\theta(z,c)$ and pair coefficients $J_\theta(z,c)$. The energy is
\begin{equation}
\E_\theta(z;c)
=-\sum_i h_{\theta,i}(z,c)z_i
-\sum_{(i,j)\in\mathcal G,\,i<j}J_{\theta,ij}(z,c) z_i z_j,
\label{eq:energy}
\end{equation}
where $\theta$ denotes the trainable hypernetwork weights. Figure~\ref{fig:workflow} connects the binary encoder, the energy evaluation, and the subsequent decoding of sampled or inferred codes. Equation~\eqref{eq:energy} uses the field and pair terms of the Ising Hamiltonian \cite{ising}. The fields $h_{\theta,i}$ weight single spins, and the couplings $J_{\theta,ij}$ weight spin products on the edges of $\mathcal G$. A term $J_{\theta,12}(z,c)=a z_3$ contributes $-a z_1z_2z_3$ to the energy, illustrating how configuration dependence induces interactions beyond pairs. 

\subsection{Hypernetwork Architecture}

The hypernetwork computes the fields and couplings in Eq.~\eqref{eq:energy} from the full candidate spin vector $z$ and corruption condition $c$. Its input is
\begin{equation}
 d_0=[z\,;\,W_c c+b_c],
 \label{eq:hypernet-input}
\end{equation}
where $[a\,;\,b]$ denotes concatenation and the learned affine map of the scalar $c$ produces 16 features. The spin vector includes the candidate label on MNIST and contains only image-code spins on Cats. The image encoder and decoder are separate networks and are frozen during energy fitting.

A fully connected architecture inspired by U-Net~\cite{unet} processes this input through a contracting path and a corresponding expanding path. At each expansion stage, the current features are concatenated with those from the matching contraction stage. Two affine heads act on the final shared feature vector. The field head predicts one local field per spin, and the coupling head supplies the coefficients for the spin pairs in the energy function. Both outputs depend on the full input configuration and the corruption condition. Appendix~\ref{app:hypernetwork} specifies the layer widths and gives the equations for the residual blocks and skip connections.

The interaction graph specifies which spin pairs appear explicitly in the energy function. The field head emits one value per spin, so its size grows linearly with the number of spins. When the coupling head emits one coefficient per edge, its size is determined by the number of edges. A fully connected graph has one edge for every distinct pair of spins, so the number of coupling outputs grows quadratically. In a graph of fixed degree, each spin has a fixed number of neighbors, so the number of coupling outputs grows linearly. The total number of trainable weights also depends on the hidden layer widths.

For MNIST, Dense includes every pair of image-code spins, whereas Sparse8 connects each image-code spin to eight other image-code spins in a fixed graph. Both variants connect each image-code spin to all ten label spins and omit explicit couplings between label spins. With $n$ image-code spins, both variants emit $n+10$ fields. The Dense coupling head emits $n(n-1)/2+10n$ coefficients. The Sparse8 head emits $4n+10n$ because counting eight image-code neighbors per spin counts each edge twice. With 32 image-code spins, Sparse8 reduces the coupling head from 816 to 448 outputs. The reduction comes entirely from retaining fewer edges between image-code spins; the 42 field outputs and all 320 image-to-label edges are unchanged.

For Cats, there are no label spins. The field head emits $n$ values, and the coupling head emits $n^2$ raw entries. Symmetrizing the resulting matrix and setting its diagonal to zero yields $n(n-1)/2$ distinct pair coefficients. Table~\ref{tab:architecture} in Appendix~\ref{app:hypernetwork} summarizes the output dimensions for both datasets. The same appendix details the construction of the Cats coupling matrix and the energy gradients.

Each MNIST variant is fitted once per training seed, and the resulting energy is reused for classification, generation, and restoration. The cat-image experiments use a separately fitted energy for each binary code length.

\subsection{Energy Ranking and Class Supervision}

Training starts from the binary code of an image produced by the frozen encoder. We use its spin representation as the positive example and construct a negative example by flipping its spins independently or, with a small probability, replacing the entire code with independent uniform spins. We evaluate both codes at the same corruption condition and, for MNIST, with the same true label. The pairwise ranking loss uses the form associated with conditional noise-contrastive estimation~\cite{cnce}:
\begin{equation}
 \ell_{\mathrm{rank},i}=\log\!\left(1+\exp[(E_i^+-E_i^-)/\tau]\right).
 \label{eq:rank-main}
\end{equation}
Here $E_i^+$ and $E_i^-$ denote the energies of the positive and negative codes, respectively. The scale $\tau$ controls how strongly the loss responds to their energy difference. The loss decreases as the energy of the positive code falls relative to that of the negative code.

For MNIST, we also train the energy to distinguish labels for a fixed image code. We score the image-code spin vector $s_i$ with each of the ten candidate labels at $c=0$ and convert the resulting energies into conditional class probabilities:
\begin{equation}
 \pi_\theta(j\mid s_i)=
 \frac{\exp[-\E_\theta(s_i,j;0)/\tau]}
 {\sum_{k=0}^{9}\exp[-\E_\theta(s_i,k;0)/\tau]}.
 \label{eq:class-prob-main}
\end{equation}
The class cross-entropy is the negative log-probability of the true label $y_i$. At inference, the most probable class is the one with the lowest energy.

The MNIST minibatch objective combines the two losses with a squared-energy penalty:
\begin{equation}
\begin{aligned}
 \widehat{\mathcal L}_{\mathrm{MNIST}}
 &=
 \frac{1}{|\mathcal I|}\sum_{i\in\mathcal I}\ell_{\mathrm{rank},i}
 -\frac{\lambda_{\mathrm{CE}}}{B}\sum_{i=1}^{B}\log\pi_\theta(y_i\mid s_i)\\
 &\quad+\frac{\lambda_{\mathrm E}}{2B}\sum_{i=1}^{B}
 \big[(E_i^+)^2+(E_i^-)^2\big].
\end{aligned}
\label{eq:mnistloss}
\end{equation}
Here $B$ is the minibatch size, $\mathcal I$ indexes pairs whose codes differ, and the first term is set to zero if $\mathcal I$ is empty. The weights $\lambda_{\mathrm{CE}}$ and $\lambda_{\mathrm E}$ control the class loss and the energy penalty. For Cats, the objective is the mean of $\ell_{\mathrm{rank},i}$ over the full minibatch, including unchanged pairs, with $\tau=1$. Both objectives update only the hypernetwork. Appendix~\ref{app:learning} specifies the corruption kernels and gradients, and Appendix~\ref{app:objectives} gives the optimization and validation settings.

\begin{figure}[!htbp]
\centering
\includegraphics[width=\linewidth]{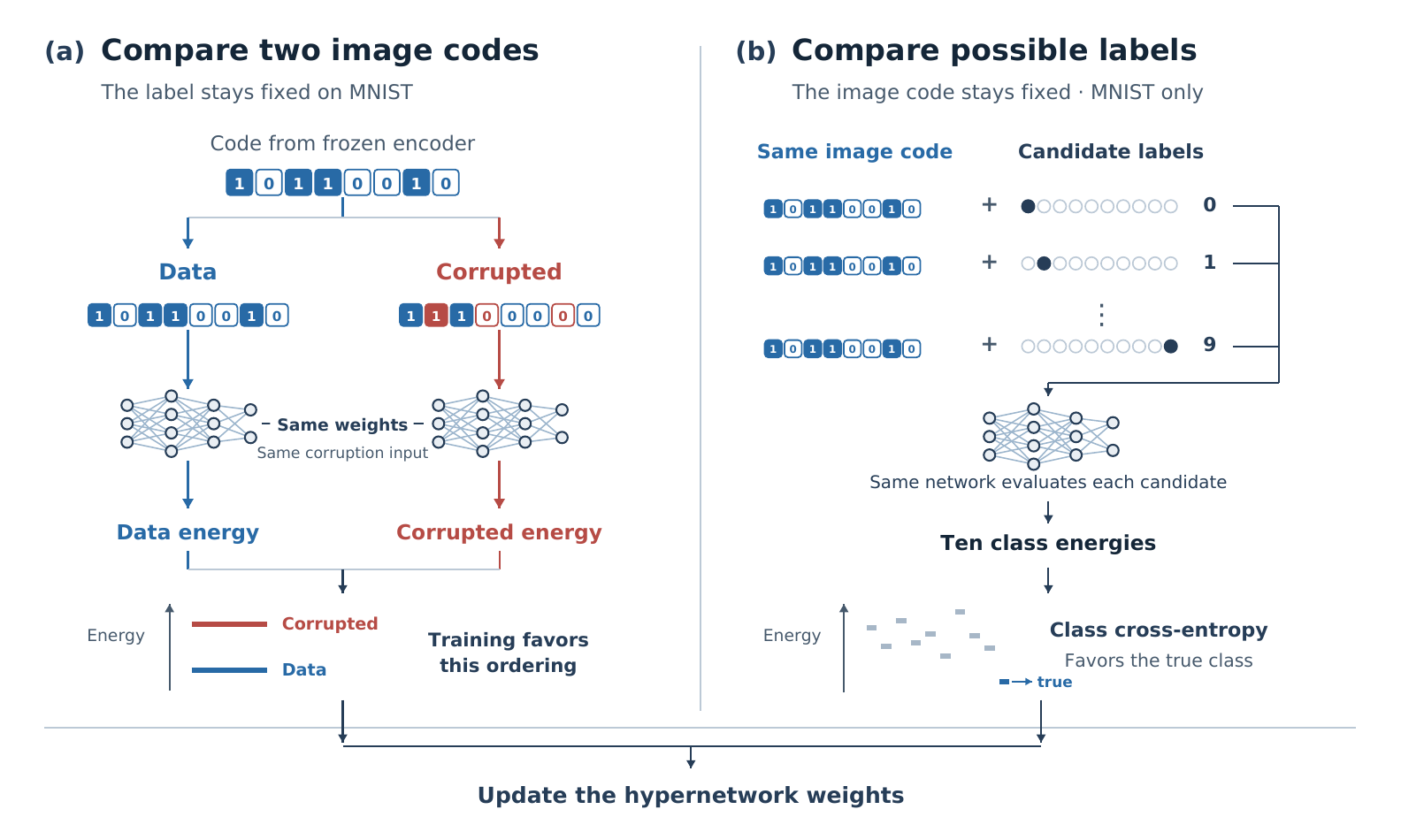}
\caption{Training the HyperIsing energy. (a) An image code from the frozen encoder is compared with a corrupted version of the same code. Both are scored at the same corruption condition and, for MNIST, at the same true label. The ranking loss favors a lower energy for the data code. (b) For MNIST, the data code is held fixed and scored with each of the ten candidate labels at $c=0$. The classification loss favors a lower energy for the correct label relative to the alternatives. The same hypernetwork computes fields and couplings separately for each candidate state; its weights are updated by the training objective. The displayed codes and energy levels are schematic.}
\label{fig:training-v2}
\end{figure}

Figure~\ref{fig:training-v2} illustrates the two comparisons used in training. Panel~(a) compares a positive code with a negative code at the same corruption condition, holding the true label fixed for MNIST. Panel~(b) holds the MNIST image code fixed and compares the energies assigned to the ten candidate labels. In both panels, the same hypernetwork recomputes the fields and couplings for each candidate state.

\subsection{Sampling and Masked Restoration}

For generation, we set $c=0$ and sample the image-code spins $s$, holding the requested MNIST class fixed. For Cats, the class input is absent and $c=0$ is the endpoint of the continuous training condition. Denote the energy with these inputs fixed by $\widetilde{\E}_\theta(s)$. At a fixed temperature $T>0$, the target distribution is
\begin{equation}
 p_T(s)=
 \frac{\exp[-\widetilde{\E}_\theta(s)/T]}
 {\sum_{\bar s\in\{-1,+1\}^{n}}\exp[-\widetilde{\E}_\theta(\bar s)/T]}.
 \label{eq:gibbs-main}
\end{equation}
The denominator sums over all image-code configurations $\bar s$. Temperature $T$ controls the relative probabilities in the sampler; $c$ is a network input and $\tau$ sets the scale of the training loss.

Single-spin Metropolis updates use energy differences so that the denominator of Eq.~\eqref{eq:gibbs-main} cancels \cite{metropolis,hastings}. A proposal $s'$ differs from the current state $s$ at one spin and is accepted with probability
\begin{equation}
 a(s,s')=\min\!\left\{1,\exp[-(\widetilde{\E}_\theta(s')-
 \widetilde{\E}_\theta(s))/T]\right\}.
 \label{eq:accept-main}
\end{equation}
Each energy is computed using the fields and couplings for its own configuration. A spin flip can therefore change coefficients beyond the flipped spin and its incident edges. Each sweep proposes one flip per image-code spin in random order. The experiments use finite runs with a cooling schedule for MNIST and constant temperature for Cats. Generated spins are converted back to bits and passed to the frozen decoder.

For masked restoration, the search objective combines the learned energy with a penalty for disagreement with observed pixels \cite{geman}. For an observation $\tilde x$ and a nonempty binary observation mask $m$, the objective is
\begin{equation}
\E_{\mathrm{inf}}(b,y\mid\tilde x,m)=
\E_\theta(2b-1,y;0)+\lambda
\frac{\sum_i m_i[D_i(b)-\tilde x_i]^2}{\sum_i m_i}.
\label{eq:rest}
\end{equation}
Here $b$ is the candidate image code, $y$ is its candidate label, and $D_i(b)$ is pixel $i$ of the decoded image. The mask selects observed pixels, and $\lambda$ sets the weight of the pixel penalty. Appendix~\ref{app:inference} derives the sampling and restoration updates and distinguishes their fixed-temperature targets from the actual finite runs. Appendix~\ref{app:metrics} defines the comparison metrics.

\subsection{Experimental Protocols}
\label{sec:protocols}

In this subsection, we describe the experimental protocols for MNIST and cat images. For MNIST, we specify the data splits, model variants, and evaluation procedures for classification, conditional generation, and image restoration with joint class inference. For cat images, we describe how we vary the binary code length and sampling settings, compare HyperIsing with the baselines, and evaluate both the generated images and the spin configurations before decoding.

\subsubsection{MNIST Classification and Conditional Generation}
We use the standard MNIST training and test sets \cite{mnist}. We reserve 10,000 of the 60,000 training images for validation and fit the models with the remaining 50,000. The separate 10,000-image test set is reserved. Four variants share the 32-bit autoencoder and add ten label spins. Dense connects all pairs of latent spins; Sparse8 uses a fixed random degree-eight graph. Both retain 320 latent-to-label edges and omit label-to-label edges. Compact hypernetworks have internal widths 256, 128, and 64; small hypernetworks use 128, 64, and 32. Each variant is trained with five seeds for 5,000 steps (batch size 64, learning rate $10^{-4}$); checkpoint selection uses validation scores. The encoder, decoder, and independent pixel CNN evaluator are fixed across the comparison.

Generation uses 64 sweeps with temperature decreasing from 2 to 1 and yields 200 images per class per run. Conditional accuracy is the fraction for which the fixed CNN recovers the requested class. We compute the squared Gaussian Fr\'{e}chet expression in the CNN's 64-dimensional feature space \cite{dowson,heusel}, using 2,000 validation originals and their reconstructions as separate references.

\subsubsection{MNIST Masked Restoration and Joint Class Inference}
For restoration we use the same 100 held-out images (ten per class) and the same corrupted observations across variants and seeds. We evaluate clean images, clipped Gaussian noise with standard deviations 0.15 and 0.30 in normalized pixel units, and independently missing pixels with probabilities 25\% and 50\%. Missing pixels are set to zero in the image provided to the encoder and are excluded from the fidelity term. Four chains run for 64 sweeps from temperature 2 to 1, one initialized at the observation code and three with initial bit perturbations. The lowest-energy terminal state is retained. A search over $\lambda\in\{0,100,1000\}$ on 30 validation images minimized whole-image reconstruction MSE against the clean targets, without using labels. This criterion selected $\lambda=1000$, the largest tested value, for every model variant, training seed, and corruption condition.

The matched control minimizes observed-pixel squared error using the same decoder, observations, and 8,192 bit-flip proposals per image.

\subsubsection{Cat-Image Generation and Evaluation}
We model $64\times64$ RGB cat images with binary codes of 16, 32, 64, 128, 256, or 512 bits. The preserved source archive is associated in the project records with the \emph{Cats Faces 64$\times$64} collection \cite{catdata}. A separate binary autoencoder is frozen for each dimension. HyperIsing samples are obtained by decoding codes from random initial states after a specified number of sweeps at constant $T$, with $c=0$. We compare against independent uniform bits, a latent diffusion baseline decoded with the corresponding autoencoder, and a direct pixel diffusion baseline. These are our dataset-specific diffusion baselines, following the general denoising and latent-diffusion frameworks \cite{ddpm,ldm}. Binary Latent Diffusion \cite{binarydiff} is a related approach to generative modeling over binary image codes. The main grid uses powers of two from 2 through 1,024 for temperature and sweep count; the 512-bit grid includes additional temperatures. Archived evaluation records for the displayed HyperIsing settings and latent-diffusion comparisons contain 5,000 generated samples per configuration, 9,464 training codes and 2,028 validation codes. We found exact duplicate images across the available training and validation directories, but could not quantify the overlap in the cached codes used for the reported evaluation.

We compute feature likelihood divergence (FLD) \cite{fld} in a DINOv2 feature space \cite{dinov2}, together with feature-space recall \cite{recall}. Original and autoencoder-reconstructed image references are evaluated separately. FLD evaluates validation-reference features, and recall measures coverage of training-reference features (Appendix~\ref{app:metrics}). 

We define the Ising latent moment discrepancy (ILMD) to compare the spin statistics of generated and validation codes before decoding. For an empirical distribution $P$ over $n$-spin codes, define
\begin{equation}
 m_i(P)=\mathbb E_P[s_i],\qquad
 G_{ij}(P)=\mathbb E_P[s_i s_j].
 \label{eq:ilmd-moments}
\end{equation}
The spin mean measures occupancy of $+1$ at site $i$, and the raw pair moment measures how often spins $i$ and $j$ agree. For the generated and validation distributions $P_g$ and $P_v$, respectively, define
\begin{equation}
\begin{aligned}
 D_m^2 &= \frac1n\sum_{i=1}^{n}[m_i(P_g)-m_i(P_v)]^2,\\
 D_G^2 &= \frac1{\binom n2}\sum_{i<j}[G_{ij}(P_g)-G_{ij}(P_v)]^2,\\
 \mathrm{ILMD}(P_g,P_v)&=\sqrt{(D_m^2+D_G^2)/2}.
\end{aligned}
\label{eq:ilmd}
\end{equation}
Each distinct pair is counted once. The separate normalizations give equal weight to the single-spin and pair statistics; smaller ILMD indicates closer agreement in these moments. We also report mean Jensen-Shannon divergence \cite{lin} on matched eight-spin blocks to compare joint histograms involving more than two spins. Each block has 256 possible states. Appendices~\ref{app:ilmd} and~\ref{app:js} give the marginal interpretation of ILMD and the rationale and construction of the blockwise histograms.

\section{Results and Discussion}
\label{sec:results}
In this section, we present results for MNIST and cat images. The MNIST experiments test how the same learned energy performs across classification, class-conditional generation, and masked image restoration. The cat-image experiments compare generation across six binary code lengths and a range of sampling settings. Comparisons with autoencoder reconstructions and with uniform-code and diffusion baselines help distinguish the effects of the binary representation from those of the sampling method.

\subsection{MNIST Classification, Generation, and Restoration}

We first compare how network width and latent graph connectivity affect MNIST classification and class-conditional generation. We then evaluate masked restoration and label recovery, separating the two outcomes with a matched control that optimizes observed-pixel fidelity alone.

\subsubsection{Classification and Conditional Generation Across Model Sizes}
\fig{0.55}{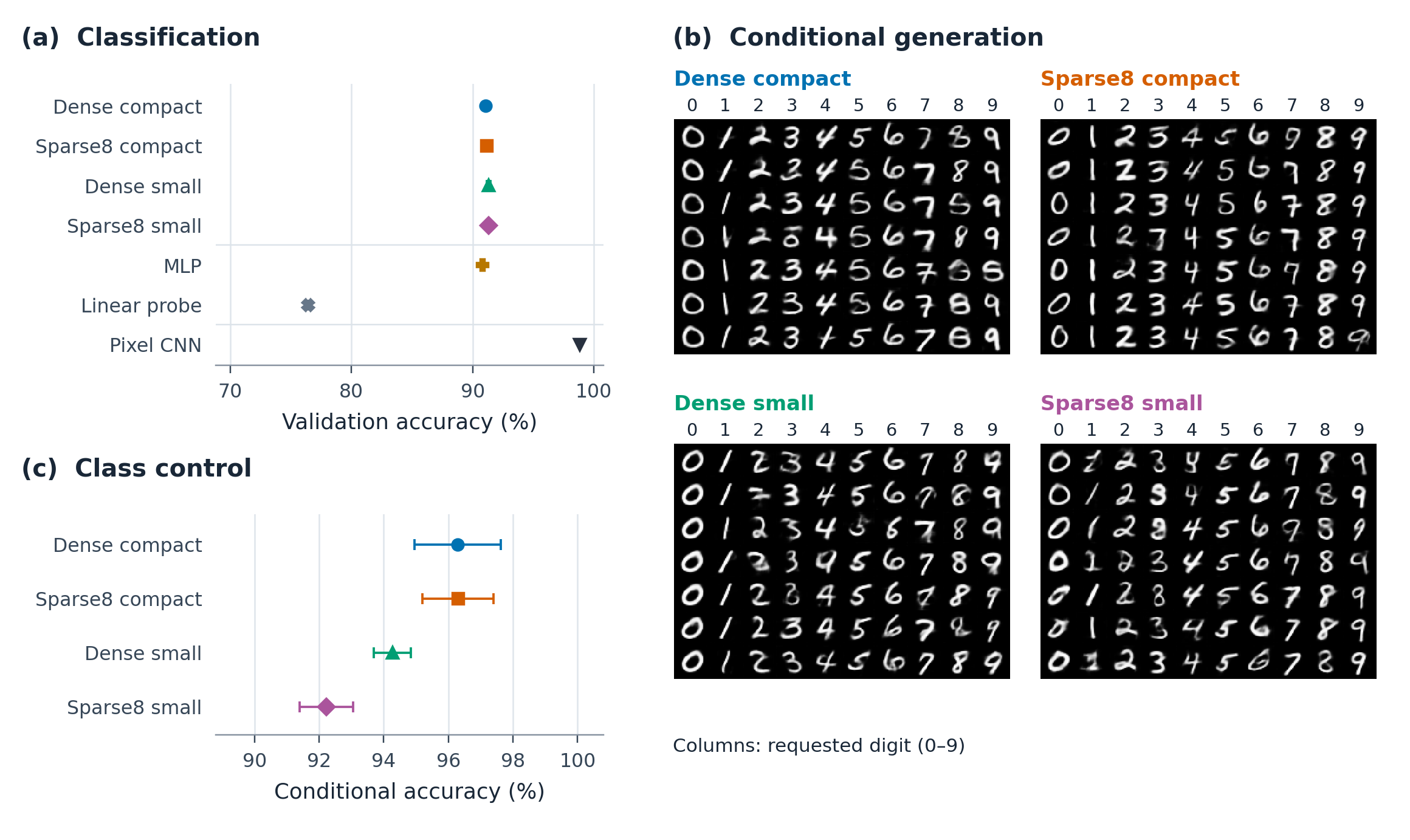}{fig:mnist}{MNIST classification and conditional generation with the same learned energy. (a) Accuracy on 10,000 validation images; the MLP and linear probe use the same binary codes, while the CNN uses pixels. (b) Seven generated examples per requested digit for each seed-zero checkpoint, shown without filtering. (c) A fixed CNN evaluates 2,000 generated images per variant and run. Error bars in (a,c) are sample standard deviations across five runs where applicable.}

\begin{table}[!tb]
\centering
\caption{MNIST energy-model sizes and class-conditional generation accuracy. Values are means and sample standard deviations over five paired training and sampling runs. Counts exclude the fixed autoencoder and CNN.}
\label{tab:mnist}
\begin{tabular}{lrrr}
\toprule
Variant & Latent edges & Trainable parameters & Conditional accuracy (\%) \\
\midrule
Dense compact   & 496 & 3,326,266 & $96.28\pm1.34$ \\
Sparse8 compact & 128 & 3,231,690 & $96.30\pm1.10$ \\
Dense small     & 496 &   894,554 & $94.27\pm0.57$ \\
Sparse8 small   & 128 &   847,082 & $92.23\pm0.83$ \\
\bottomrule
\end{tabular}
\end{table}

We use a linear probe to assess how well the digit classes can be separated in the binary representation learned by the autoencoder. With the encoder frozen, we train a classifier that maps the 32-bit image code directly to ten class scores through a single affine layer. Its validation accuracy provides a baseline for comparison with HyperIsing on the same image codes.

Across five runs, the four HyperIsing variants achieve mean validation accuracies of 91.08--91.34\% (Fig.~\ref{fig:mnist}). The MLP and linear probe trained on the same binary codes achieve $90.84\pm0.14$\% and $76.44\pm0.06$\%, respectively. The pixel CNN achieves 98.87\% on the original validation images and 92.51\% on their autoencoder reconstructions. The latter is a reference for interpreting classifier accuracy on decoded images, not a cap on conditional generation accuracy.

The compact variants yield the highest mean conditional accuracies. Dense small yields the lowest mean distance to originals ($129.08\pm19.63$), and Sparse8 small yields the lowest mean distance to reconstructions ($31.84\pm5.48$). The small variants retain the validation accuracy range above with 73--74\% fewer trainable energy parameters (Fig.~\ref{fig:tradeoffs}). At a fixed sample count of 2,000, Dense compact yields a mean of 1,257 distinct codes, whereas Sparse8 small yields 1,973. This comparison places conditional accuracy alongside exact code repetition (Appendix~\ref{app:mnist-supp}). Appendix~\ref{app:mnist-supp} reports reconstruction, diversity, and timing results.

\fig{0.57}{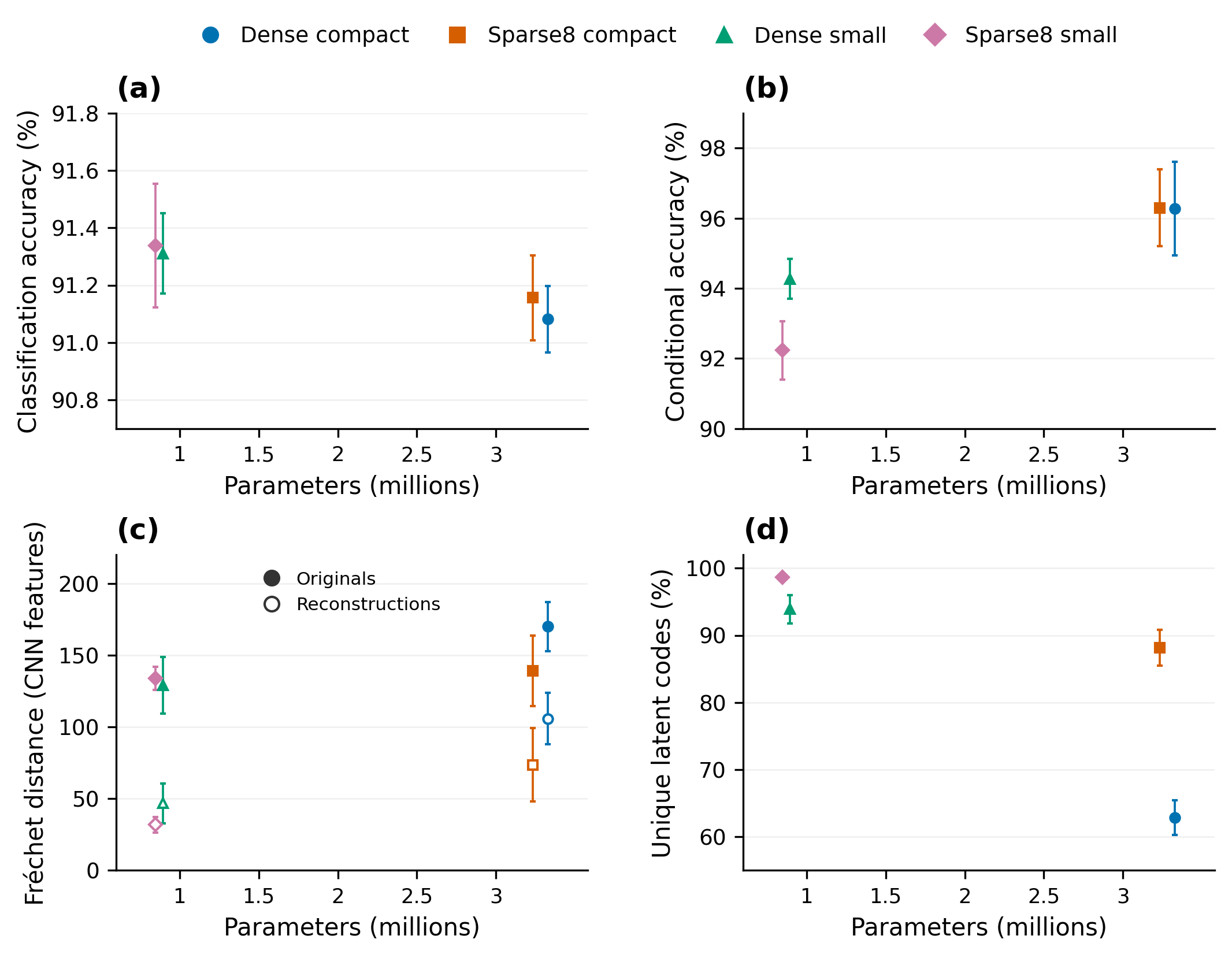}{fig:tradeoffs}{Effect of energy-model size and connectivity on MNIST. The horizontal axis counts trainable HyperIsing parameters only. Panels report validation classification, conditional generation accuracy, Fr\'{e}chet distance to original (filled) and reconstructed (open) reference images, and the unique-code fraction over 2,000 generated samples. Points and bars show means and sample standard deviations across five runs. The run-level variation includes both training and sampling.}

\subsubsection{Masked Restoration and Joint Class Inference}

At 50\% missing pixels, Dense small yields $93.2\pm2.3$\% joint label accuracy and the MLP yields $69.0\pm1.2$\% on observation codes. Sparse8 small yields the lowest whole-image mean squared error (MSE) among the four variants, $0.01700\pm0.00013$, compared with $0.04839$ for direct autoencoder reconstruction (Fig.~\ref{fig:rest}). The MLP and direct decoder operate on the observation code, whereas HyperIsing iterates over candidate codes and labels using the observed pixels. Appendix~\ref{app:mnist-supp} reports the fixed CNN accuracy on restored images.

\fig{0.73}{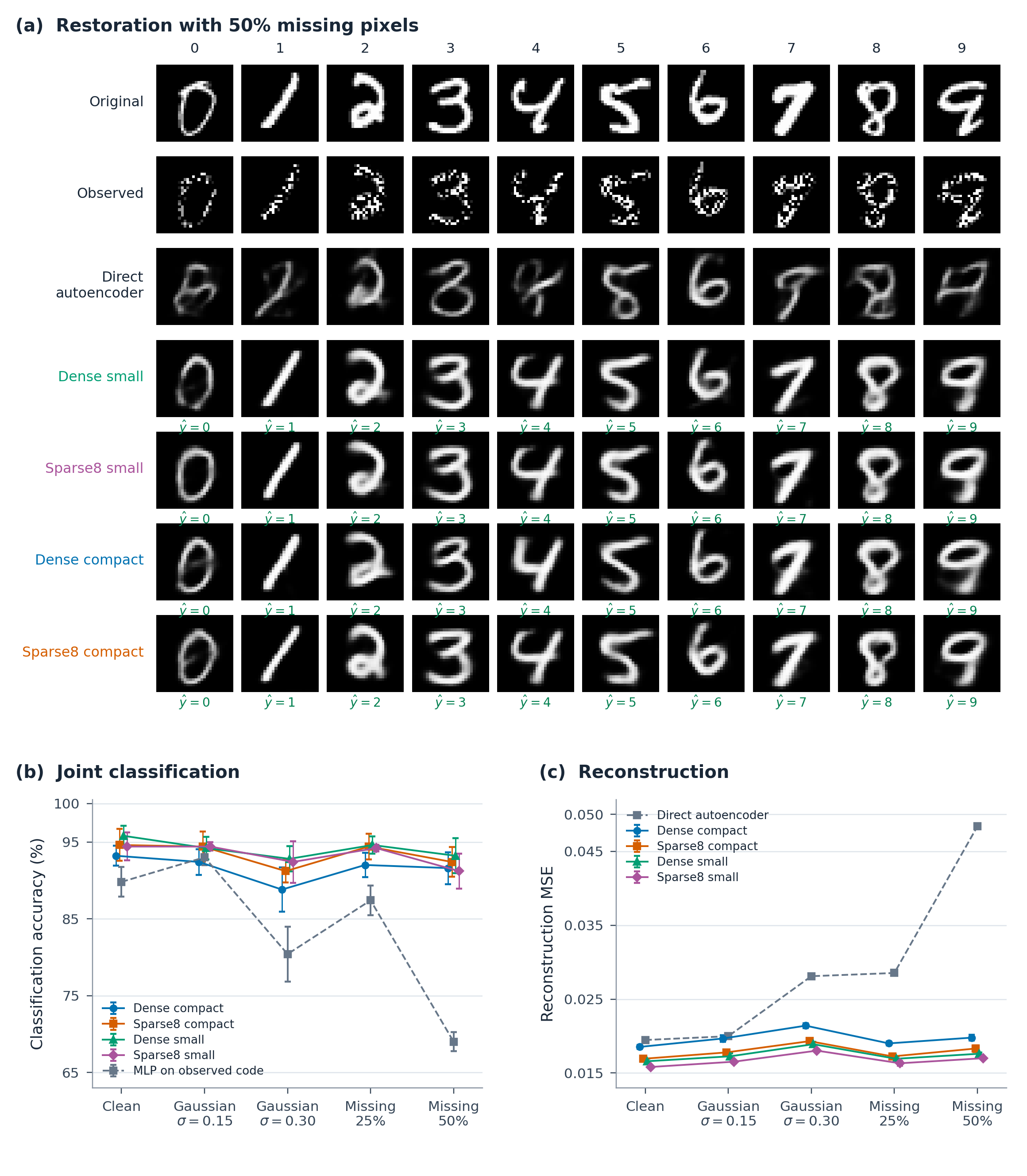}{fig:rest}{Joint restoration and classification on MNIST. (a) Rows show the clean image, observation with 50\% missing pixels, direct autoencoder reconstruction, and restorations from the four HyperIsing variants, with inferred labels. (b) Joint class-inference accuracy and the MLP on observation codes. (c) Reconstruction MSE against clean targets and the direct autoencoder reference. HyperIsing error bars are sample standard deviations across five training seeds evaluated on the same 100 images and corruptions.}

The matched control minimizes observed-pixel squared error and yields lower mean whole-image MSE across all five conditions and lower missing-pixel MSE at both mask levels. The pointwise, class-stratified bootstrap intervals for the paired missing-pixel differences lie below zero at both mask levels (Fig.~\ref{fig:fidelity}). The fidelity objective estimates the image alone; the joint objective also estimates its class. Matching the number of latent proposals controls the search budget in proposal counts. HyperIsing additionally evaluates the energy model and updates the class, so the two procedures have different computation costs. This control compares reconstruction error; the label result compares joint inference with classification on the observation code.

\fig{0.37}{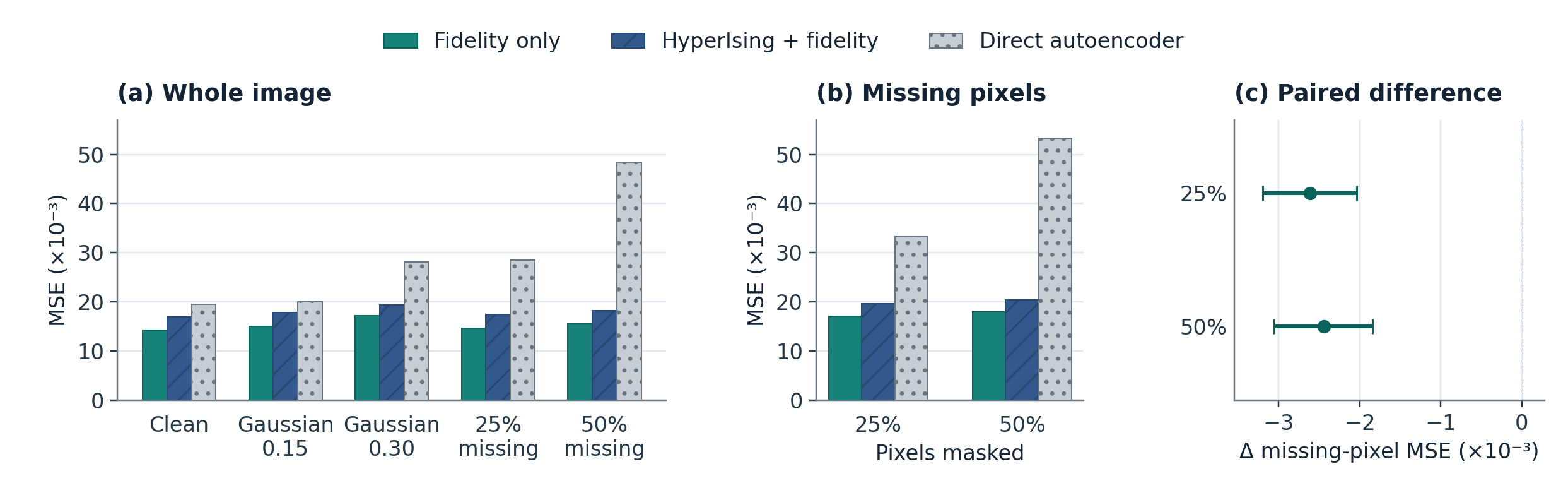}{fig:fidelity}{Matched restoration control. (a) Whole-image MSE across five conditions. (b) MSE restricted to missing pixels. (c) Paired difference in missing-pixel MSE (fidelity only minus HyperIsing with fidelity), with pointwise 95\% bootstrap intervals resampling image identities within digit class. Both iterative methods use the same frozen decoder, observations, and 8,192 bit-flip proposals per image. The iterative results are averaged across five runs, and HyperIsing additionally across its four variants. Direct autoencoding uses a single decoder pass on the observation code.}

\FloatBarrier
\subsection{Cat-Image Generation Across Latent Dimensions}
\label{sec:cats}
\fig{0.60}{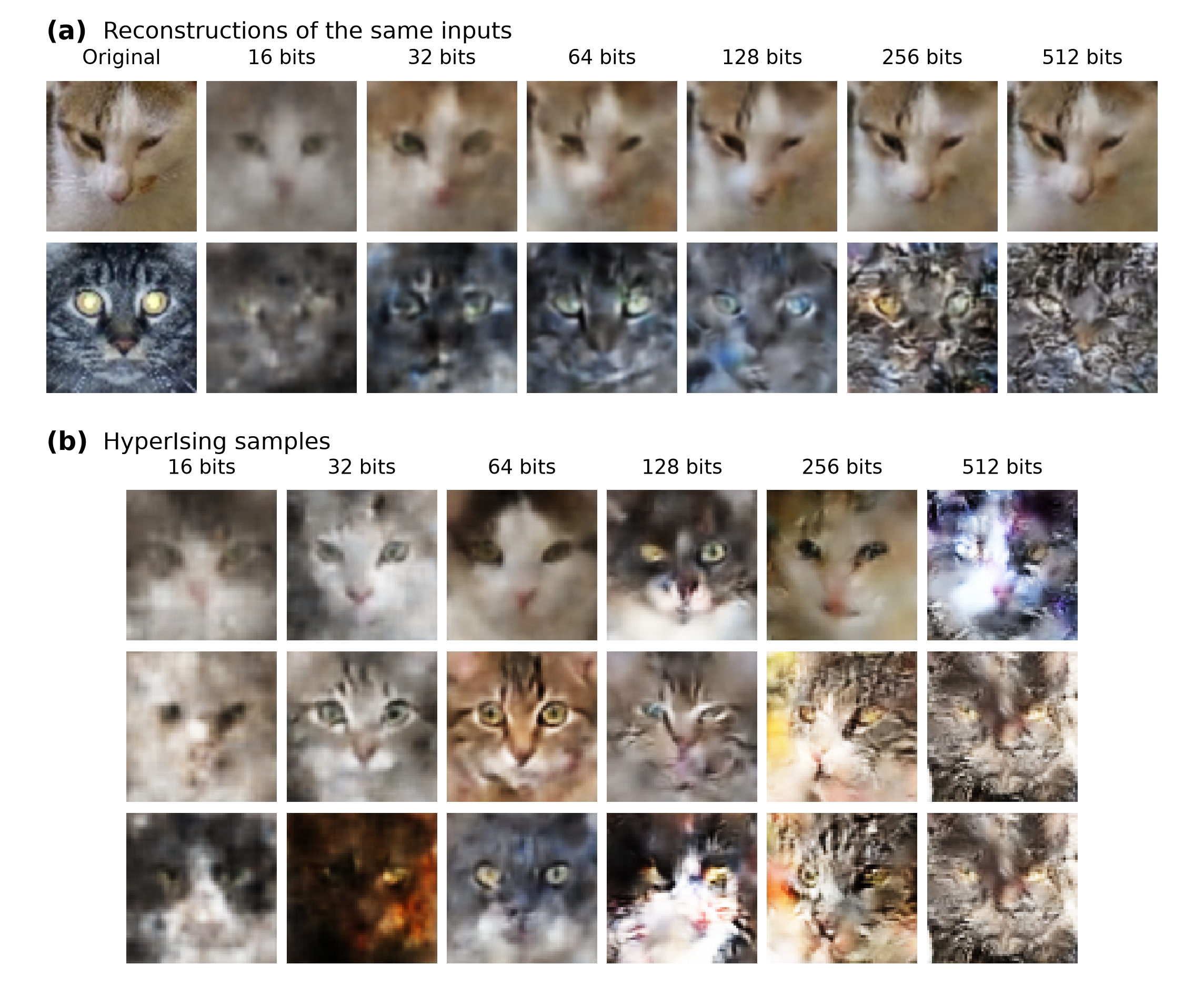}{fig:cats-reco}{Visual comparison at six latent dimensions. (a) The same two validation images and their reconstructions. (b) The first three HyperIsing samples at the evaluated setting with lowest observed validation FLD to original images for each dimension, in generation order and without visual filtering.}

\begin{figure}[h]
\centering
\includegraphics[width=\linewidth]{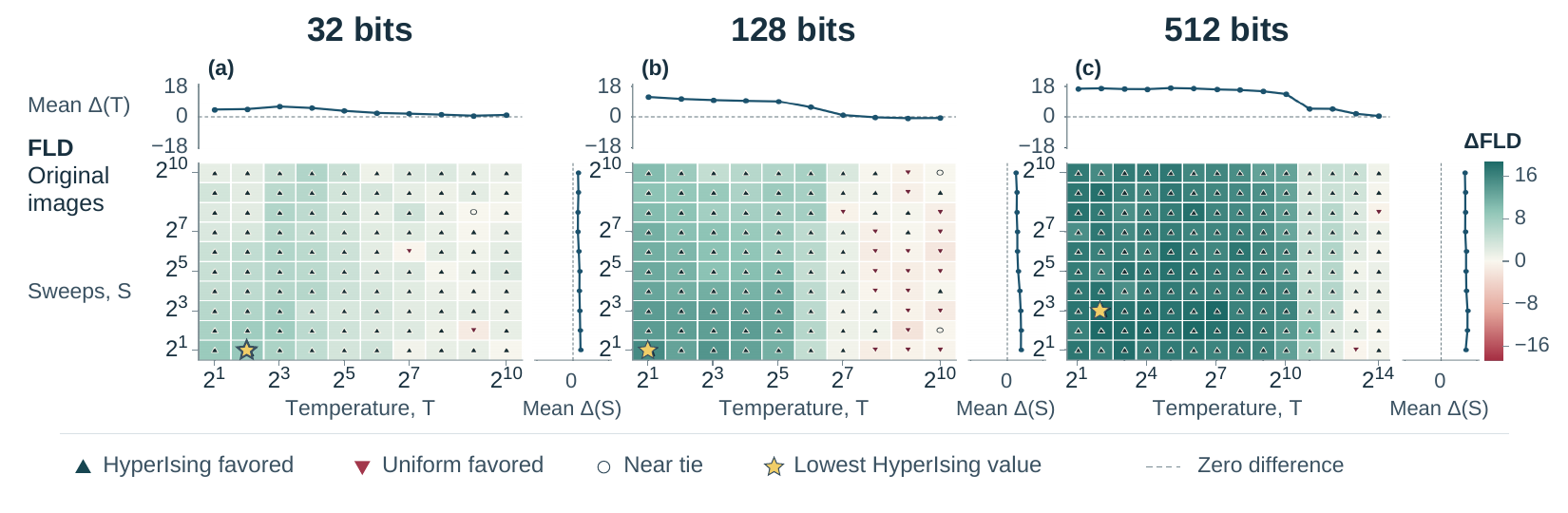}
\caption{FLD to original images for 32, 128, and 512 bits as a function of temperature $T$ and sweep count $S$. Color gives FLD for uniform latent codes minus FLD for HyperIsing; positive values favor HyperIsing. Triangles indicate the favored method, circles indicate an absolute difference of at most 0.10, and stars mark the lowest evaluated HyperIsing FLD in each panel. The top and right traces average over sweep counts and temperatures, respectively. The 512-bit grid includes additional temperatures.}
\label{fig:cats-maps}
\end{figure}

At low bit counts, the autoencoder reconstructions retain less details compared to the original images (Fig.~\ref{fig:cats-reco}). The lowest observed original-reference FLD varies nonmonotonically with code length. Figure~\ref{fig:cats-maps} compares original-reference FLD across the temperature and sweep-count grid. Figure~\ref{fig:cats-maps-extra} in Appendix~\ref{app:cats-plots} gives the corresponding maps for reconstruction-reference FLD and ILMD. At 512 bits, the lowest reconstruction-reference FLD occurs at $T=8$ and two sweeps (4.49 versus 13.86 for uniform sampling), whereas the lowest original-reference FLD occurs at $T=4$ and eight sweeps. The two minima use the same autoencoder and code length. The setting selected by FLD therefore depends on which image distribution is used as its reference.

Across the displayed dimensions, the selected HyperIsing settings yield lower original-reference FLD than uniform latent sampling and latent diffusion. Pixel diffusion yields the lowest original-reference FLD and the highest recall to originals. HyperIsing yields the lowest FLD to autoencoder reconstructions, and the highest reconstruction-reference recall varies by dimension. At the largest code sizes, the latent discrepancies of HyperIsing and uniform sampling approach one another (Fig.~\ref{fig:cats-comp}). Each point is selected separately over the temperature-sweep grid for its metric and reference. Appendix~\ref{app:cats-supp} reports the values and selected settings; Appendix~\ref{app:cats-plots} gives the remaining maps and marginal summaries.

The cat results describe the evaluated grid and baseline implementations. FLD, ILMD, and blockwise JS are scored against validation references; recall measures coverage of the training reference set. Each reported HyperIsing optimum is selected and reported using its own metric and reference. Thus the curves compare metric-specific grid optima obtained on the same data used for selection. Across code lengths, the frozen autoencoder, the number of coupling outputs, and the number of proposals per sweep also change together. The comparison therefore describes the complete configurations evaluated here.

\fig{0.73}{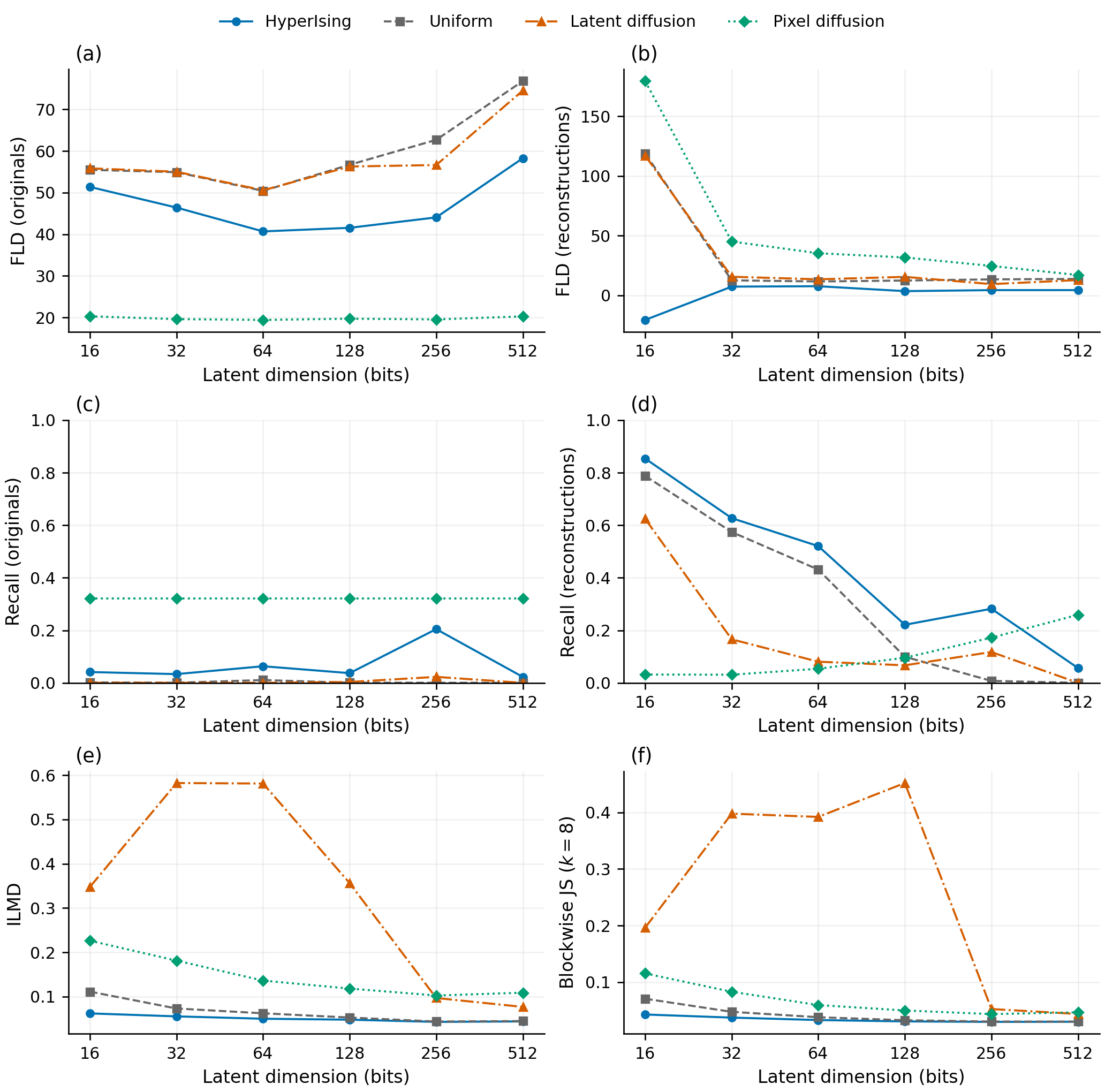}{fig:cats-comp}{Cat-image generator comparison. (a,b) FLD relative to original images and autoencoder reconstructions. (c,d) Recall using training features from those two reference types. (e,f) Ising latent moment discrepancy and mean Jensen-Shannon divergence over eight-bit blocks. Comparators are uniform latent codes, latent diffusion, and pixel diffusion. Lower is better in (a,b,e,f), higher in (c,d). Each HyperIsing point is the best observed for its metric, reference, and latent dimension on the evaluated grid. Connecting curves therefore join settings that can differ between dimensions and panels.}

\FloatBarrier
\section{Conclusions}
\label{sec:conclusions}

In this work, we presented an energy-based model for binary discrete data using hypernetworks. A hypernetwork computes the fields and pair coefficients separately for each candidate spin configuration. Training the model shapes the energy by comparing data codes with corrupted alternatives and, for labeled data, by comparing competing labels on a fixed code. Once fitted, the same energy model supports three inference problems: it ranks labels for a given code, guides sampling at a requested label, and contributes a prior over codes and labels during masked restoration. The frozen decoder maps generated and restored codes back to pixels.

We explored generative modeling for two image problems: the MNIST dataset and a dataset of cat images. The MNIST comparisons show that the three tasks emphasize different properties of the learned energy. Reducing the hypernetwork width substantially lowers the parameter count while maintaining similar mean validation classification accuracy. The compact networks give the highest mean class-conditional generation accuracy, while the small variants attain the lowest feature distances, and the sparse small version generates more distinct codes than fully connected connectivity. For restoration, the matched control gives lower reconstruction error when the search minimizes observed-pixel squared error alone. Joint inference returns both the restored image and an estimate of its class. 

For cat-image generation, the two image references distinguish how well the model samples the encoded distribution from how well the decoded images match the original data. Among the evaluated configurations, direct pixel diffusion attains the lowest FLD to original images, and HyperIsing attains the lowest FLD to autoencoder reconstructions. Even at a fixed latent dimension, the preferred temperature and sweep count can change with the reference. Together with the latent moment and blockwise comparisons, this shows why the sampling schedule and the distribution used as a target must be specified when comparing latent generators.

The configuration-dependent coefficients allow effective interactions beyond the explicit pair graph. The empirical results characterize a system that combines this parameterization with a binary representation, a contrastive training objective, and a finite sampling protocol. The two datasets establish complementary uses of that system: reuse of one learned energy across labeled inference tasks, and comparisons of unconditional generation against distinct image-space and latent-space reference distributions.

\section*{Acknowledgments}

This work was supported by Petrobr\'as/CENPES, grant No.~2025/00505-2 and by the São Paulo Research Foundation (FAPESP), under Grants 2025/27502-3, No. 2023/15739-3, No. 2022/00209-6, and No.  2025/15490-0. C. J. V.-B. acknowledges the support from the National Council for Scientific and Technological Development (CNPq), grants No. 422316/2023-7 and No. 302234/2026-8.

\section*{Data availability}

The numerical results underlying the MNIST and cat-image comparisons, including the selected cat sampling settings, are provided in the supplementary tables and figures accompanying this manuscript. The source collection associated with the preserved cat-image archive is linked in Ref.~\cite{catdata}. The supplementary material does not include a file-level manifest of the historical split or the run-specific checkpoints. The source code and instructions for running the MNIST and cat-image experiments are provided in the HamilGen repository at \url{https://github.com/NicoCabrera3225/HamilGen}.

\FloatBarrier
\bibliographystyle{unsrt}
\bibliography{references}
\clearpage
\appendix
\section{Binary Encoding and HyperIsing Training}
\label{app:learning}
\subsection{Binary Autoencoder and Straight-Through Estimator}\label{app:binary-ae}
Write $a_\phi(x)$ for the encoder logits of an image and $b^\star=\mathbf 1\{a_\phi(x)\geq0\}$ for the code, component by component. The forward pass is binary. With $\operatorname{sg}$ denoting stop-gradient, the quantity fed to the decoder during autoencoder training is
\[
b_{\mathrm{STE}}=a_\phi(x)+\operatorname{sg}\!\left[b^\star-a_\phi(x)\right].
\]
The forward pass applies hard thresholding, and the backward pass uses the identity as the surrogate Jacobian with respect to $a_\phi$ \cite{ste}. The archived MNIST and Cats training implementations use the same autoencoder objective:
\[
\mathcal L_{\mathrm{AE}} =
\operatorname{MSE}\bigl(D_\psi(b_{\mathrm{STE}}),x\bigr)
+0.25\,\operatorname{MSE}\bigl(\sigma(a_\phi),b^\star\bigr)
+10^{-4}\operatorname{mean}(a_\phi^2),
\]
where $b^\star$ is a fixed target for the commitment term and $\sigma$ denotes the logistic sigmoid. The MNIST autoencoder was trained with Adam \cite{adam} for 30 epochs at batch size 64 and learning rate $10^{-3}$, selecting by validation reconstruction MSE. The cat autoencoder also uses a deterministic binary latent representation and an STE; its encoder reduces $64\times64$ spatial inputs to $8\times8$ feature maps and projects them to $n$ logits. This deterministic threshold differs from VQ-VAE, where encoder outputs are assigned to learned codebook entries \cite{vqvae}. Both pipelines freeze encoder and decoder before fitting the energy; ranking updates operate on the cached binary codes.

\subsection{Hypernetwork Parameterization and Energy Gradients}
\label{app:hypernetwork}
Let $z$ denote the full network input in spin coordinates and $c$ the corruption condition. A learned affine layer maps $c$ to 16 features, which are concatenated with $z$. At width $w$, the residual unit is
\[
R_w(v)=v+W_2\,\operatorname{SiLU}\!\left(W_1\operatorname{LN}(v)+b_1\right)+b_2,
\quad W_1\in\mathbb R^{4w\times w},\quad W_2\in\mathbb R^{w\times4w},
\]
Here the residual connection adds $v$ to the learned branch \cite{resnet}. Layer normalization $\operatorname{LN}$ precedes the first affine map \cite{layernorm}, and $\operatorname{SiLU}(u)=u\,\sigma(u)$ \cite{silu}, where $\sigma$ is the logistic sigmoid. Let $w_0=\dim d_0$ and $w_k=\dim d_k$ for $k=1,2,3$, with the values given in Table~\ref{tab:architecture}. The down stages are
\[
d_k=R^{\mathrm{down}}_k(W^{\mathrm{down}}_k d_{k-1}+b^{\mathrm{down}}_k).
\]
The matrix $W^{\mathrm{down}}_k$ has shape $w_k\times w_{k-1}$, $b^{\mathrm{down}}_k$ is a bias vector of length $w_k$, and $R^{\mathrm{down}}_k$ is the composition of two independently parameterized residual units at width $w_k$. Two further units at width $w_3$ produce $u_4$ from $d_3$.

Following the skip-connection design of U-Net~\cite{unet}, each expansion stage concatenates its projected features with those from the corresponding contraction stage. The update is
\begin{equation}
u_k=R_k\!\left(M_k\!\left([P_k(u_{k+1})\,;\,d_k]\right)\right),
\qquad k=3,2,1.
\label{eq:hypernet-skip}
\end{equation}
Here $P_k$ projects $u_{k+1}$ to width $w_k$, $M_k$ is an affine map from the concatenated features of width $2w_k$ back to width $w_k$, and $R_k$ applies two residual units at that width.
In Eq.~\eqref{eq:hypernet-skip}, the linear parts of $P_3,P_2,P_1$ have shapes $w_3\times w_3$, $w_2\times w_3$, and $w_1\times w_2$, respectively; the linear part of each $M_k$ has shape $w_k\times 2w_k$.

The final feature vector $u_1$ is shared by two affine output heads:
\begin{equation}
h_\theta(z,c)=W_hu_1+b_h,\qquad
\mathbf j_\theta(z,c)=W_ju_1+b_j.
\label{eq:hypernet-heads}
\end{equation}
Here $W_h$ and $W_j$ are learned weight matrices, $b_h$ and $b_j$ are their corresponding biases, and $\mathbf j_\theta$ contains the raw coupling outputs.
The two heads in Eq.~\eqref{eq:hypernet-heads} therefore have $w_1$ input features and the output sizes listed in Table~\ref{tab:architecture}.

\begin{table}[!htbp]
\centering
\caption{Shared hypernetwork design and dataset-specific inputs and outputs. $n$ is the number of image-code spins. The 16 condition features are included in the input widths shown below.}
\label{tab:architecture}
\setlength{\tabcolsep}{5pt}
\begin{tabular}{@{}p{0.18\linewidth}p{0.19\linewidth}p{0.22\linewidth}p{0.34\linewidth}@{}}
\toprule
Model & Input width & Down widths & Output heads \\
\midrule
MNIST compact & $32+10+16$ & 256, 128, 64 & 42 fields; 816 Dense or 448 Sparse8 edge coefficients \\
MNIST small & $32+10+16$ & 128, 64, 32 & Same heads as compact \\
Cats & $n+16$ & 1,024, 512, 256 & $n$ fields; $n^2$ entries symmetrized with zero diagonal \\
\bottomrule
\end{tabular}
\end{table}

MNIST uses $z=(s,\ell(y))$, where $s=2b^\star-1$ and $\ell(y)$ is the ten-spin representation of the scalar class label $y$ defined in Sec.~\ref{sec:method}. With $n=32$ there are 42 fields and one output per retained unordered edge: 816 for Dense, 448 for Sparse8 (including 320 latent--label edges in each). The class input affects the trunk as well as the explicit label-spin products. For cats, $z=s$ and the heads return $n$ fields and a raw matrix $A\in\mathbb R^{n\times n}$. The latter is converted to
\[
J=\tfrac12(A+A^\top)-\operatorname{diag}(A_{11},\ldots,A_{nn}),
\qquad \E_\theta(z;c)=-h^\top z-\tfrac12 z^\top Jz.
\]
The factor of one half prevents double-counting symmetric interactions. In MNIST the edge sum in Eq.~\eqref{eq:energy} counts each edge once. For a fixed input $z$, the gradients with respect to the evaluated heads are $\partial \E/\partial h_i=-z_i$ and $\partial \E/\partial J_{ij}=-z_i z_j$ for each independently stored edge coefficient. At fixed $z$ and $c$ the chain rule is
\[
\nabla_\theta\E=-\sum_i z_i\nabla_\theta h_i
-\sum_{(i,j)\in\mathcal G,\ i<j}z_i z_j\nabla_\theta J_{ij}.
\]
For cats the raw off-diagonal head entries have $\partial \E/\partial A_{ij}=-z_i z_j/2$ for $i\ne j$ because both $A_{ij}$ and $A_{ji}$ contribute to the same independent symmetric edge. The raw diagonal has zero derivative. Both heads and the shared blocks receive these gradients; the binary encoder is frozen. Because $h$ and $J$ are functions of $z$, the full scalar map need not be quadratic and can contain effective interactions of order higher than two. For example, a coefficient $J_{12}(z)=a z_3$ produces the term $-a z_1 z_2 z_3$. Sparsity constrains the explicit pair terms; the coefficient networks still depend on the full spin configuration.

\subsection{Corruption Kernel and Pairwise Ranking}
An encoded example supplies a positive spin vector $s$ and, on MNIST, its true class $y$. Draw $\rho\sim\mathcal U[0,\tfrac12]$. With probability $1-u$ form $s^-$ by flipping each spin independently with probability $\rho$; with probability $u=0.05$ replace the whole negative by independent uniform spins. The MNIST label is unchanged. Both positive and negative are scored at the same condition. On MNIST, $c=0$ with probability $0.25$, independently of the draws used for spin corruption; otherwise $c=\rho$. The MNIST branch therefore includes corrupted negatives at $c=0$. The cat ranking implementation sets $c=\rho$. The independently specified Metropolis temperature $T$ controls inference.

Define $\E_i^+=\E_\theta(s_i,y_i;c_i)$, $\E_i^-=\E_\theta(s_i^-,y_i;c_i)$, and omit $y_i$ for cats. With $\operatorname{sp}(t)=\log(1+e^t)$ and ranking scale $\tau$,
\begin{align}
g_i&=\frac{\E_i^+-\E_i^-}{\tau},&
\ell_{\mathrm{rank},i}&=\operatorname{sp}(g_i),\\
\nabla_\theta\ell_{\mathrm{rank},i}
&=\frac{\sigma(g_i)}{\tau}
\left(\nabla_\theta\E_i^+-\nabla_\theta\E_i^-\right).
\label{eq:rank-grad}
\end{align}
A pair with an incorrectly high positive energy contributes a larger multiplier. In scalar energy coordinates, $\partial \ell/\partial \E_i^+=\sigma(g_i)/\tau$ and $\partial\ell/\partial\E_i^-=-\sigma(g_i)/\tau$; thus the pair term pushes the positive score down and the negative score up in energy. The optimizer aggregates these contributions through shared network parameters. The explicit corruption kernel generates the negatives, and the loss uses only the pair energies. We use the pair-ordering form associated with conditional noise-contrastive estimation \cite{cnce}; Eq.~\eqref{eq:mnistloss} gives the complete supervised and regularized MNIST objective.

\subsection{Dataset-Specific Objectives and Optimization}
\label{app:objectives}
Equation~\eqref{eq:mnistloss} gives the MNIST minibatch loss. A pair with an unchanged code has ranking loss $\log 2$ and zero pair gradient; excluding it changes the denominator of the ranking average. For the class loss derived from Eq.~\eqref{eq:class-prob-main}, its derivative with respect to the energy of class $j$ is $(\mathbf 1\{j=y_i\}-\pi_\theta(j\mid s_i))/\tau$. In energy coordinates, gradient descent therefore lowers the true-class energy and raises the other energies in proportion to their current probabilities.
The last term in Eq.~\eqref{eq:mnistloss} penalizes squared energies for every pair, including unchanged pairs. Before weighting, it contributes $(\mathcal E_i^{\pm}/B)\nabla_\theta\mathcal E_i^{\pm}$ from each branch. The run settings are $\tau=1$, $\lambda_{\mathrm{CE}}=1$, and $\lambda_{\mathrm E}=10^{-4}$. Each of 5,000 Adam updates uses 64 examples sampled with replacement at learning rate $10^{-4}$; gradient norm is clipped at 5. The validation criterion is ranking plus weighted class cross-entropy with fixed validation corruption draws, omitting the energy penalty. The archived MNIST energy-model checkpoint records these loss settings.

For cats, the training implementation uses
\[
\widehat{\mathcal L}_{\mathrm{cat}}=\frac1B\sum_{i=1}^B
\operatorname{sp}\bigl(\E_\theta(s_i;c_i)-\E_\theta(s_i^-;c_i)\bigr).
\]
The mean includes unchanged pairs, and the ranking scale is one. The cat training implementation uses Adam with learning rate $10^{-4}$ and batch size 128, tracks validation ranking, and saves at the end of training. These settings describe the training implementation; checkpoint-specific results are indexed by latent dimension and sampling configuration.

\subsection{Population Identifiability under a Symmetric Corruption Kernel}
Fix a class $y$ and a value of $c$. At fixed $\rho$ the negative transition on $n$ spins is
\[
q_\rho(t\mid s)=(1-u)\rho^{d_H(s,t)}(1-\rho)^{n-d_H(s,t)}
+u\,2^{-n}.
\]
It is symmetric in $s$ and $t$. Conditional on $c=0$ in the MNIST zero-conditioning branch, the effective transition is a mixture of these symmetric kernels with weights independent of $s$, so it remains symmetric. More generally the following argument needs symmetry of $q_c$ and full connection of the positive support; the uniform branch connects all binary states. Restricting to distinct pairs multiplies both orderings by the same selection indicator and leaves their odds unchanged.

In an idealized population ranking-only problem, let $p_y$ be the clean code probability for class $y$. For an unordered pair with positive probability in both orders, the two ordering weights are $p_y(s)q_c(t\mid s)$ and $p_y(t)q_c(s\mid t)$. Their log-ratio is
\[
\log\frac{p_y(s)q_c(t\mid s)}{p_y(t)q_c(s\mid t)}
=\log p_y(s)-\log p_y(t).
\]
A logistic log loss is minimized by the true ordering probability: if that probability is $r$, differentiation of $-r\log \hat r-(1-r)\log(1-\hat r)$ gives its minimum at $\hat r=r$. Predicting the order with logit $(\E_\theta(t,y;c)-\E_\theta(s,y;c))/\tau$ therefore admits $\E(s,y;c)=-\tau\log p_y(s)+C(y,c)$ on the positive support in the unrestricted ranking-only model \cite{cnce}. If a state has zero density but can be paired with a positive state, exact separation on such pairs instead requires an infinite energy gap in the ideal limit. For cats replace $p_y$ by $p$ and the additive constant by $C(c)$. The continuous corruption draw assigns zero probability mass to the endpoint $c=0$ used for cat-image generation. The population result holds almost everywhere in the training condition; identifying the cat endpoint requires regularity in $c$. The MNIST zero-conditioning branch gives that endpoint positive mass. The calculation applies to unrestricted energies under the ranking-only population objective. Standard noise-contrastive estimation uses independent noise \cite{nce}, whereas contrastive divergence uses transitions of the model chain \cite{cd}.

\section{Monte Carlo Sampling and Restoration}
\label{app:inference}
\subsection{Metropolis Updates for State-Dependent Interactions}
At fixed class (if present), $c=0$ and positive sampling temperature $T$, define the finite-state target $\mu_T(z)\propto e^{-\E_\theta(z;0)/T}$. A symmetric proposed flip has Metropolis acceptance $a(z,z')=\min\{1,e^{-\Delta\E/T}\}$, where $\Delta\E=\E_\theta(z';0)-\E_\theta(z;0)$ \cite{metropolis,hastings}. The fixed-site update satisfies detailed balance because both oriented probability flows equal $\min\{\mu_T(z),\mu_T(z')\}$ times the common proposal factor; the probability of a rejected move remains on the diagonal. Every fixed-site kernel preserves $\mu_T$, hence a sweep, as a composition in any prescribed order, preserves it too. The argument uses stationarity of each component kernel. A state-independent random permutation mixes these compositions and preserves $\mu_T$ at fixed $T$.

For $z'$ obtained by flipping site $k$, write $h=h(z)$, $J=J(z)$, $\delta h=h(z')-h(z)$, and $\delta J=J(z')-J(z)$ at fixed condition. Adding and subtracting the energy at $z'$ computed with the old coefficients gives
\[
\Delta\E =
2z_k\!\left(h_k+\sum_{j\ne k}J_{kj}z_j\right)
-\delta h^\top z'-\tfrac12 z'^\top\delta J z'.
\]
The first term is the constant-coefficient Ising update. The remaining terms account for the changes in the predicted fields and couplings. We evaluate a fresh energy for each proposal. The current energy can be cached: the generation code evaluates the initial state for each sweep and then one proposed state per site, retaining the accepted energy. The invariance result holds at fixed $T$. We evaluate finite sweeps at constant temperature on cats and finite cooling schedules on MNIST.

\subsection{Joint Restoration and Matched Fidelity Control}
Let $f_m(b)=\sum_i m_i[D_i(b)-\tilde x_i]^2/\sum_i m_i$ for a nonempty mask. In the implementation the learned energy is first divided by its checkpoint energy scale $\tau_{\mathrm{energy}}$, separately from the sampler temperature, so the target at fixed $T$ is proportional to
\[
\exp\!\left[-\frac{\E_\theta(2b-1,y;0)/\tau_{\mathrm{energy}}+\lambda f_m(b)}{T}\right].
\]
The MNIST result uses $\tau_{\mathrm{energy}}=1$; this reduces to Eq.~\eqref{eq:rest}. The quadratic data term sums over observed pixels. The coefficient $\lambda$ sets its weight relative to the spin energy. The prior-plus-data-penalty formulation follows the general logic of Gibbs image restoration \cite{geman}, with a learned energy on latent codes and a class variable in this implementation.

At fixed label, a latent-bit proposal compares the total objective. At fixed code $b$, the fidelity term is independent of $y$ and cancels from the Gibbs conditional over the labels:
\[
p_T(y\mid b,\tilde x,m)
=\frac{\exp[-\E_\theta(2b-1,y;0)/(\tau_{\mathrm{energy}}T)]}
{\sum_j\exp[-\E_\theta(2b-1,j;0)/(\tau_{\mathrm{energy}}T)]}.
\]
Both updates preserve the joint target at fixed $T$. The actual algorithm uses a finite cooling schedule and returns the lowest-objective terminal state among four chains. The returned state is the best terminal candidate under this finite search. The fidelity-only control removes the learned term and the label variable, retaining the same observation, decoder and count of latent proposals $4\cdot64\cdot32=8192$. The control estimates the image; classification accuracy is evaluated for the label-producing methods. In Figure~\ref{fig:fidelity}, uncertainty for a paired MSE contrast comes from resampling test image identities within class while pairing the two outcomes for every resampled image \cite{efron}; the intervals quantify variability across test images conditional on the evaluated runs.

\section{Evaluation Metrics and Statistical Protocols}
\label{app:metrics}
\subsection{MNIST Classification, Generation, and Restoration Metrics}
For observed images, classification accuracy counts agreement with the true labels. For generated images, conditional accuracy counts agreement between the requested label $y_k$ and the prediction $C(\hat x_k)$ of the frozen pixel CNN:
\[
A_{\mathrm{cond}}=\frac1{N_g}\sum_{k=1}^{N_g}\mathbf 1\{C(\hat x_k)=y_k\}.
\]
Each run is balanced over the ten requested classes. This statistic measures class control according to a fixed evaluator. The unique-code fraction $|\{b_1,\ldots,b_{N_g}\}|/N_g$ quantifies exact code repetition at fixed sample count.

Let $(\widehat\mu_g,\widehat\Sigma_g)$ and $(\widehat\mu_r,\widehat\Sigma_r)$ be the empirical mean and sample covariance of generated and reference images in the frozen CNN's 64-dimensional feature space. We report the squared Gaussian Fr\'echet expression \cite{dowson,heusel},
\[
F_{\mathrm{CNN64}}=\|\widehat\mu_g-\widehat\mu_r\|_2^2+
\operatorname{tr}\!\left[
\widehat\Sigma_g+\widehat\Sigma_r
-2\left(\widehat\Sigma_g^{1/2}\widehat\Sigma_r\widehat\Sigma_g^{1/2}\right)^{1/2}
\right].
\]
The implementation uses symmetric eigendecompositions, clipping negative roundoff in the spectra before square roots and clipping the final value below at zero. It evaluates 2,000 images per generation run against the same 2,000 validation originals or their autoencoder reconstructions. The CNN64 Fr\'echet metric compares Gaussian approximations specified by the first and second moments of the extracted features.

Restoration MSE is the pixel mean of $(\hat x_i-x_i)^2$ against the clean image $x$, including every pixel for the whole-image score and only $m_i=0$ for the missing-pixel score. These scores are distinct from the observed-pixel penalty optimized by the inference procedure. The joint model supplies the inferred label. The matched restoration contrast subtracts the two outcomes for the same image and corruption before averaging; negative fidelity-only-minus-HyperIsing differences favor the control.

\subsection{Feature Likelihood Divergence and Generative Recall}
We embed images with DINOv2 \cite{dinov2}. For original-image comparisons, the reference feature sets come from the original training and validation images; for reconstruction comparisons, each reference image is encoded and decoded using the autoencoder for the relevant latent width before extracting features. These reference images ask different questions: how closely the generated images resemble the data, and how closely they resemble what this frozen representation can reconstruct. The reconstruction reference evaluates agreement with the distribution produced by the frozen autoencoder.

Feature likelihood divergence (FLD) \cite{fld,fldcode} estimates a feature-space density using an equal-weight Gaussian mixture centered at generated features $g_1,\ldots,g_{N_g}$:
\[
\widehat p_g(v)=\frac{1}{N_g}\sum_{j=1}^{N_g}
\mathcal N(v;g_j,\sigma_j^2 I_d).
\]
The implementation of the reference images are standardized for the training, validation, and generated features coordinate by coordinate using the validation mean and standard deviation. The bandwidths $\sigma_j^2$ are fitted to the training reference; the bandwidth-fitting routine includes an auxiliary density term, while the final evaluation uses the fitted mixture. The fitted mixture is scored on validation features $V$. Schematically the reported score is
\[
\mathrm{FLD}_{\rm test}=100\left[
-\frac{1}{d\,|V|}\sum_{v\in V}\log\widehat p_g(v)
-\mathrm{NLL}_{\rm ref}\right],
\]
where $\mathrm{NLL}_{\rm ref}$ is also scored on $V$. The reference mixture centers are $\min(N_g,\lfloor N_{\mathrm{train}}/2\rfloor)$ shuffled training features; its bandwidths are fitted to the remaining training features.

Generative recall estimates coverage with the neighborhood construction of Kynk\"{a}\"{a}nniemi et al. \cite{recall}, as implemented in the FLD library \cite{fldcode}. For each generated feature $g_j$, let $r_j$ be its neighborhood radius from a nearest-neighbor order statistic within the generated set. The generated set defines a union of balls; recall is
\[
\operatorname{Recall}(R,G)=
\frac{1}{|R|}\sum_{v\in R}
\mathbf 1\{\exists\,g_j:\|v-g_j\|_2<r_j\}.
\]
The cat evaluator passes 5 neighbors to the reference implementation: its distance array includes the zero distance to the point itself, so in the usual non-tied case the radius is the fourth nonself distance. The evaluated $R$ consists of training-reference features, for either original or reconstructed images.

\subsection{Ising Latent Moment Discrepancy (ILMD)}
\label{app:ilmd}
Equations~\eqref{eq:ilmd-moments} and~\eqref{eq:ilmd} define the empirical moments and ILMD.
The first term reports single-spin occupancy because $m_i=2P(s_i=+1)-1$. The second reports co-agreement because $G_{ij}=2P(s_i=s_j)-1$. Together they determine the binary pair marginals: for $a,b\in\{-1,1\}$,
\[
P(s_i=a,s_j=b)
=\frac{1+a m_i(P)+b m_j(P)+ab G_{ij}(P)}4.
\]
This follows by multiplying the indicator identities $\mathbf 1\{s_i=a\}=(1+as_i)/2$ and taking the expectation. Diagonals are excluded because $s_i^2=1$, and each distinct pair is counted once. Normalizing the two sums separately gives equal weight to the one-spin group and the group of $\binom n2$ distinct pairs. We use raw moments in the main metric; connected correlations $G_{ij}-m_im_j$ provide a separate diagnostic. The train-validation comparison provides an empirical reference at finite sample size.

ILMD measures the distance between the stated moment vectors; distinct distributions can share those moments. For example, take three spins with $s_1$ and $s_2$ independent and fair, and set $s_3=s_1s_2$. This distribution is uniform over the four states with $s_1s_2s_3=1$. It has zero means and zero distinct-spin pair moments, exactly like independent fair spins on all eight states. The two distributions thus have population ILMD zero, although their three-spin moments are different. Blockwise Jensen--Shannon divergence complements these moments by comparing joint histograms on selected coordinates.

\subsection{Blockwise Jensen-Shannon Divergence}
\label{app:js}
A histogram on the complete code has $2^n$ possible bins. With finite sample sets at large $n$ it can report almost no empirical overlap even when two underlying distributions are identical. For two independent sample sets of sizes $N_g,N_v$ from the same uniform $n$-bit distribution, the expected number of matching cross-set sample pairs is $N_gN_v/2^n$. Summing the pairwise match probabilities bounds the probability of seeing any match by the same quantity. When the two empirical supports are disjoint, their Jensen-Shannon divergence is $\ln 2$, regardless of whether the underlying sources differ.

We therefore restrict each code to the same subset $S$ of eight coordinates in both sample sets. Each projected distribution has at most $2^8=256$ possible strings. The subsets are coordinate selections independent of spatial adjacency. Write $P_S,Q_S$ for the unsmoothed empirical histograms of generated and validation codes and $M_S=(P_S+Q_S)/2$. For each subset we calculate \cite{lin}
\[
\mathrm{JS}(P_S,Q_S)=
\tfrac12\sum_a P_S(a)\log\frac{P_S(a)}{M_S(a)}
+\tfrac12\sum_a Q_S(a)\log\frac{Q_S(a)}{M_S(a)} ,
\]
with terms of zero mass omitted. The implementation draws up to 500 distinct subsets with a fixed seed (12345) and reuses their indices for comparisons at the same dimensionality. The reported score averages one divergence per subset. Limiting each histogram to 256 states reduces empirical support separation and captures joint structure within each subset, including interactions beyond pairs. It summarizes the selected marginals at the stated sample size.

\subsection{Temperature-Sweep Grid and Metric Selection}
For each metric where smaller is better, the heat-map contrast at temperature $T$ and sweep count $S$ is $\Delta(T,S)=M_{\rm uniform}-M_{\rm HyperIsing}(T,S)$. Positive cells indicate lower HyperIsing values. The right and top curves average over grid cells. The figure showing best observed scores reports $\min_{T,S}M_{\rm HyperIsing}(T,S)$ for FLD, ILMD and blockwise JS, and $\max_{T,S}\operatorname{Recall}_{\rm HyperIsing}(T,S)$ for coverage, each separately for its reference and latent dimension. Each metric and reference defines its own grid optimum. The diffusion values are computed for the evaluated pixel and latent implementations.

\FloatBarrier
\setcounter{figure}{0}
\renewcommand{\thefigure}{S\arabic{figure}}
\renewcommand{\theHfigure}{supp.\arabic{figure}}
\setcounter{table}{0}
\renewcommand{\thetable}{S\arabic{table}}
\renewcommand{\theHtable}{supp.\arabic{table}}

\section{MNIST Reconstruction, Latent Diversity, and Runtime}
\label{app:mnist-supp}
\begin{figure}[!tbp]\centering
\includegraphics[width=.62\linewidth]{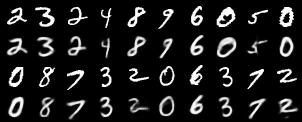}
\caption{MNIST reconstruction examples for the frozen 32-bit autoencoder. The first and third rows show validation originals; the second and fourth rows show the corresponding reconstructions.}
\label{fig:supp-ae}\end{figure}

The frozen 32-bit autoencoder obtains a validation reconstruction MSE of 0.01889. The frozen pixel CNN recognizes 98.87\% of the originals and 92.51\% of their reconstructions under that evaluation. Figure~\ref{fig:supp-ae} shows validation images and their corresponding autoencoder reconstructions.

The compact variants achieve higher conditional accuracy, but their generated batches contain fewer distinct codes. Dense compact gives $1257.0\pm52.3$ distinct codes out of 2,000, compared with $1973.2\pm9.5$ for Sparse8 small. Figure~\ref{fig:supp-diversity} also reports whether a generated code is equal to any training code and the minimum training Hamming distance. Exact matches count all generated samples, including repeats, whose codes occur in the training set. The nearest-code distance quantifies separation from the training codes. The frozen CNN measures recognition accuracy from the restored pixels (Fig.~\ref{fig:supp-cnn}). Figures~\ref{fig:rest} and~\ref{fig:fidelity} report joint class inference and reconstruction error, respectively. Table~\ref{tab:supp-cost} reports timings by workflow stage. The MLP and linear probe serve as classification baselines.

\fig{0.35}{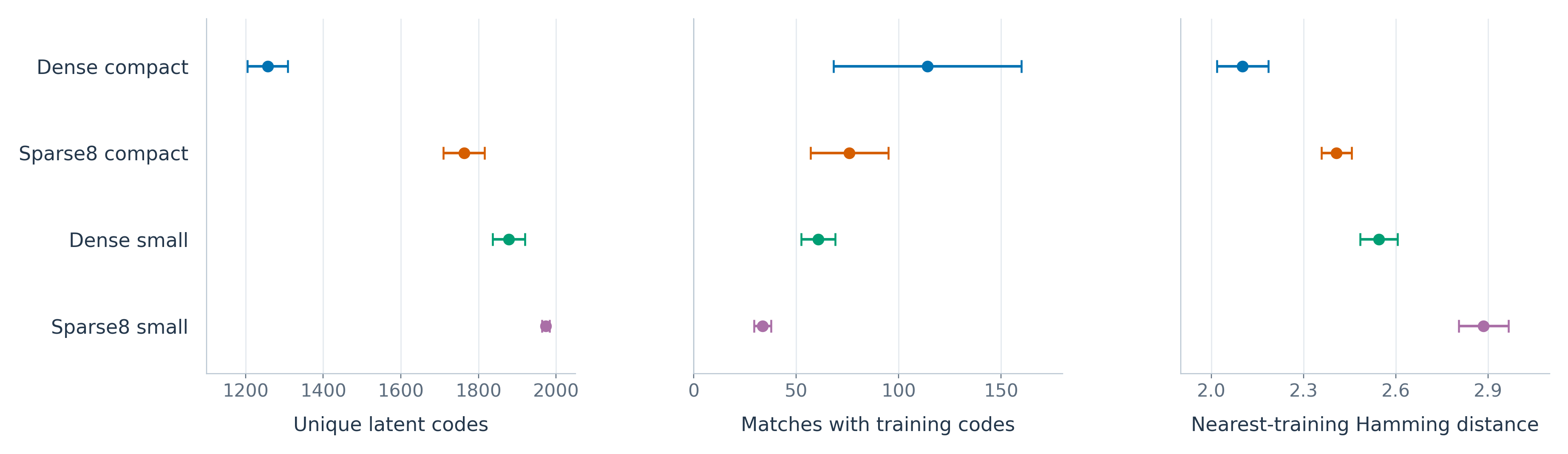}{fig:supp-diversity}{MNIST latent diversity diagnostics. Left: distinct configurations among 2,000 samples per run. Middle: generated samples whose latent code appears exactly in the training set. Right: mean unnormalized Hamming distance from each generated code to its closest training code. Points and bars show mean and sample standard deviation across the five runs.} 

\fig{0.55}{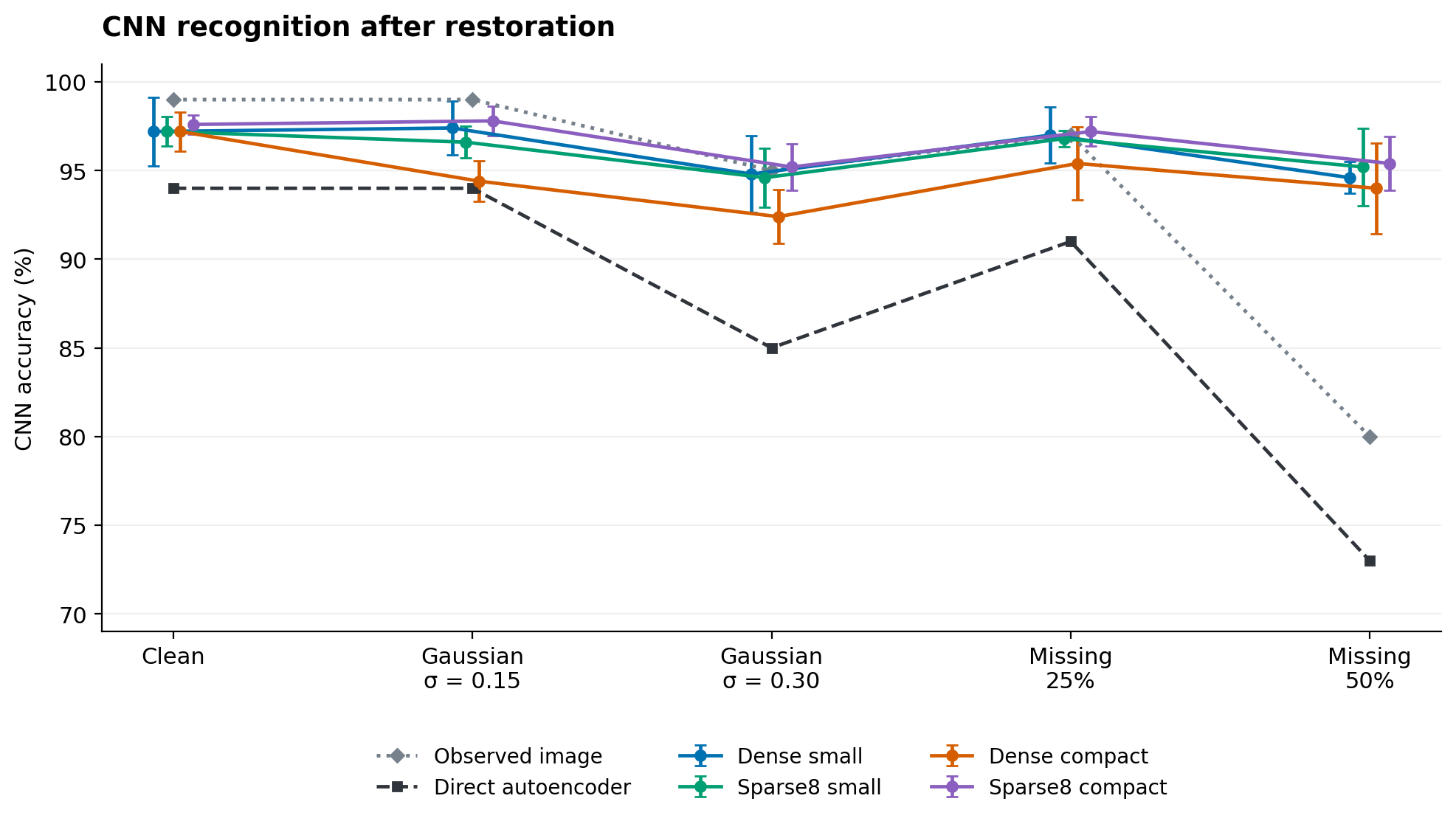}{fig:supp-cnn}{Recognition after MNIST restoration. Every curve uses the same frozen CNN to predict from observed pixels, direct autoencoder reconstructions, or the pixels restored by HyperIsing. Points and bars for the four HyperIsing variants show means and sample standard deviations across five seeds on the same test inputs.}

\begin{table}[h]
\centering
\caption{Observed MNIST compute times, mean and sample standard deviation over five runs. Training time is the accumulated update time at the selected checkpoint. Classification is timed on 10,000 cached codes. Latent sampling time covers 2,000 output codes prior to decoding and evaluation.}
\label{tab:supp-cost}
\setlength{\tabcolsep}{5pt}
\begin{tabular}{@{}lrrr@{}}
\toprule
Model & Training (s) & Classification (ms) & Latent sampling (s)\\
\midrule
Dense compact   & $217.4\pm25.6$ & $503.31\pm7.26$ & $5.22\pm0.09$\\
Sparse8 compact & $221.9\pm20.0$ & $389.67\pm16.48$ & $5.14\pm0.14$\\
Dense small     & $224.5\pm10.8$ & $380.99\pm10.85$ & $5.12\pm0.13$\\
Sparse8 small   & $208.9\pm22.3$ & $383.27\pm13.12$ & $5.09\pm0.22$\\
MLP             & $12.1\pm1.1$   & $9.64\pm0.09$   & ---\\
Linear probe    & $10.4\pm0.3$   & $4.36\pm0.04$   & ---\\
\bottomrule
\end{tabular}
\end{table}

\clearpage
\section{Cat-Image Metrics and Selected Sampling Settings}
\label{app:cats-supp}
Tables~\ref{tab:cats-fld}-\ref{tab:cats-latent} report the values underlying Fig.~\ref{fig:cats-comp}. Each row lists the best observed HyperIsing setting for that metric and reference. Baseline values follow their respective sampling protocols. Recall is reported as a proportion. The train--validation contrasts provide empirical references for the latent metrics (Appendix~\ref{app:ilmd}).

\begin{table}[!htbp]
\centering\small
\caption{Cat-image validation FLD against the original and reconstructed image reference sets. Lower is better.}
\label{tab:cats-fld}
\setlength{\tabcolsep}{5pt}
\begin{tabular}{@{}llrrrrrr@{}}
\toprule
Reference & Bits & $T$ & Sweeps & HyperIsing & Uniform & Latent diff. & Pixel diff. \\
\midrule
Originals & 16 & 2 & 512 & 51.38 & 55.48 & 55.85 & 20.28 \\
 & 32 & 4 & 2 & 46.41 & 54.92 & 55.03 & 19.61 \\
 & 64 & 8 & 2 & 40.68 & 50.43 & 50.56 & 19.41 \\
 & 128 & 2 & 2 & 41.55 & 56.70 & 56.28 & 19.71 \\
 & 256 & 2 & 2 & 44.05 & 62.69 & 56.64 & 19.53 \\
 & 512 & 4 & 8 & 58.27 & 76.81 & 74.54 & 20.26 \\
Reconstr. & 16 & 2 & 4 & -20.49 & 118.87 & 117.13 & 179.49 \\
 & 32 & 2 & 4 & 7.40 & 12.66 & 15.68 & 45.14 \\
 & 64 & 8 & 2 & 7.74 & 11.74 & 13.53 & 35.38 \\
 & 128 & 32 & 2 & 3.63 & 12.50 & 15.58 & 31.81 \\
 & 256 & 4 & 2 & 4.48 & 13.48 & 9.51 & 24.73 \\
 & 512 & 8 & 2 & 4.49 & 13.86 & 12.89 & 17.11 \\
\bottomrule
\end{tabular}
\end{table}

\begin{table}[!htbp]
\centering\small
\caption{Generative recall against training images or their reconstructions in the DINOv2 feature space. Higher is better.}
\label{tab:cats-recall}
\setlength{\tabcolsep}{5pt}
\begin{tabular}{@{}llrrrrrr@{}}
\toprule
Reference & Bits & $T$ & Sweeps & HyperIsing & Uniform & Latent diff. & Pixel diff. \\
\midrule
Originals & 16 & 1024 & 256 & 0.041 & 0.002 & 0.000 & 0.322 \\
 & 32 & 16 & 1024 & 0.033 & 0.000 & 0.000 & 0.322 \\
 & 64 & 64 & 64 & 0.063 & 0.011 & 0.000 & 0.322 \\
 & 128 & 16 & 256 & 0.037 & 0.001 & 0.003 & 0.322 \\
 & 256 & 8 & 128 & 0.205 & 0.000 & 0.023 & 0.322 \\
 & 512 & 2 & 1024 & 0.023 & 0.000 & 0.000 & 0.322 \\
Reconstr. & 16 & 4 & 128 & 0.854 & 0.788 & 0.624 & 0.032 \\
 & 32 & 16 & 128 & 0.627 & 0.574 & 0.166 & 0.031 \\
 & 64 & 32 & 4 & 0.520 & 0.431 & 0.081 & 0.054 \\
 & 128 & 2 & 2 & 0.221 & 0.099 & 0.067 & 0.096 \\
 & 256 & 8 & 1024 & 0.282 & 0.008 & 0.117 & 0.172 \\
 & 512 & 2 & 1024 & 0.056 & 0.000 & 0.000 & 0.259 \\
\bottomrule
\end{tabular}
\end{table}

\begin{table}[!htbp]
\centering\small
\caption{Latent discrepancies to validation codes. ILMD and blockwise Jensen--Shannon use unrelated scales; lower is better for each.}
\label{tab:cats-latent}
\setlength{\tabcolsep}{5pt}
\begin{tabular}{@{}llrrrrrr@{}}
\toprule
Metric & Bits & $T$ & Sweeps & HyperIsing & Uniform & Latent diff. & Pixel diff. \\
\midrule
ILMD & 16 & 2 & 2 & 0.0621 & 0.1113 & 0.3477 & 0.2263 \\
 & 32 & 32 & 128 & 0.0554 & 0.0732 & 0.5823 & 0.1813 \\
 & 64 & 64 & 512 & 0.0501 & 0.0623 & 0.5812 & 0.1362 \\
 & 128 & 128 & 64 & 0.0479 & 0.0528 & 0.3564 & 0.1182 \\
 & 256 & 1024 & 1024 & 0.0430 & 0.0437 & 0.0969 & 0.1026 \\
 & 512 & 16384 & 256 & 0.0440 & 0.0448 & 0.0768 & 0.1088 \\
JS (8 bits) & 16 & 4 & 64 & 0.0430 & 0.0709 & 0.1964 & 0.1156 \\
 & 32 & 32 & 8 & 0.0376 & 0.0478 & 0.3977 & 0.0832 \\
 & 64 & 64 & 1024 & 0.0333 & 0.0382 & 0.3920 & 0.0597 \\
 & 128 & 128 & 64 & 0.0309 & 0.0329 & 0.4519 & 0.0502 \\
 & 256 & 512 & 32 & 0.0300 & 0.0306 & 0.0527 & 0.0437 \\
 & 512 & 16384 & 256 & 0.0302 & 0.0305 & 0.0440 & 0.0469 \\
\bottomrule
\end{tabular}
\end{table}

\FloatBarrier
\section{Cat-Image Sampling Maps and Distributional Diagnostics}
\label{app:cats-plots}
\begin{figure}[!htbp]
\centering
\includegraphics[width=\linewidth]{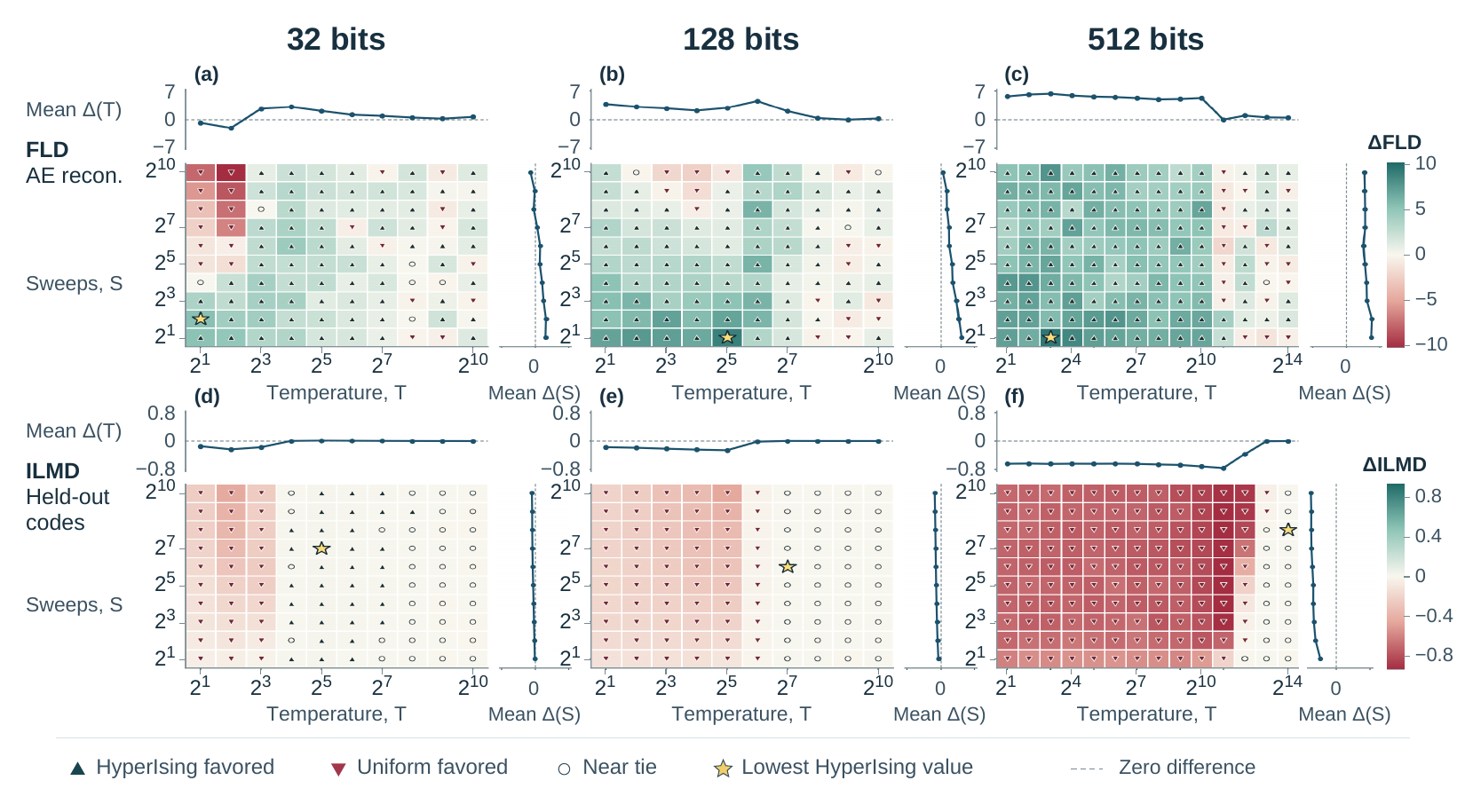}
\caption{Additional cat sampling maps for 32, 128, and 512 bits. The top row shows FLD to autoencoder reconstructions; the bottom row shows ILMD to held-out codes. Positive differences favor HyperIsing over uniform latent sampling. Triangles indicate the favored method; circles indicate near ties (absolute difference at most 0.10 for FLD and 0.005 for ILMD); stars mark the lowest evaluated HyperIsing metric in each panel. The top and right traces average over sweep counts and temperatures, respectively.}
\label{fig:cats-maps-extra}
\end{figure}

\begin{figure}[!tbp]\centering
\begin{minipage}[t]{.48\linewidth}\centering \textbf{Originals}\par\smallskip \includegraphics[width=\linewidth]{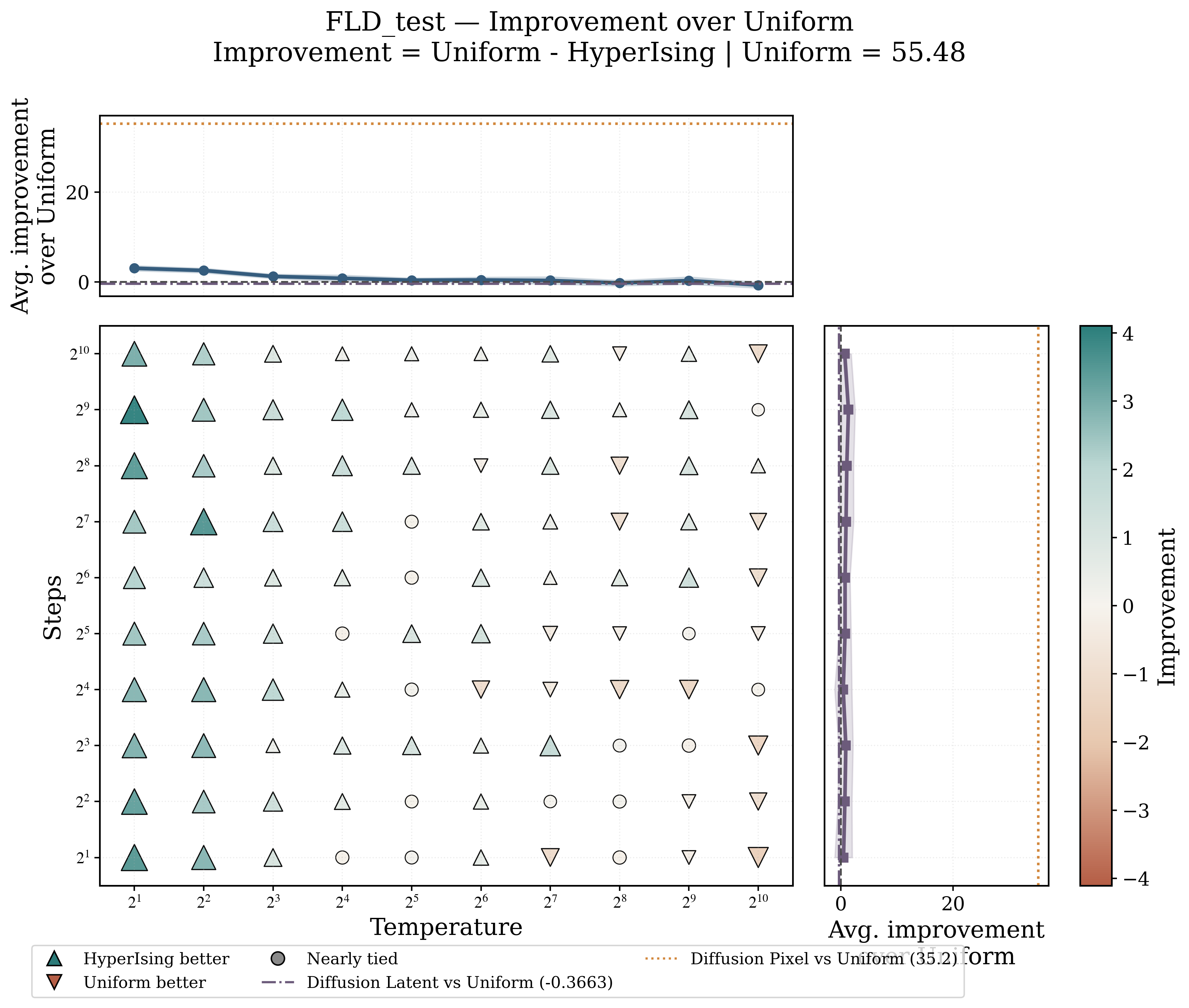}\end{minipage}\hfill \begin{minipage}[t]{.48\linewidth}\centering \textbf{Reconstructions}\par\smallskip \includegraphics[width=\linewidth]{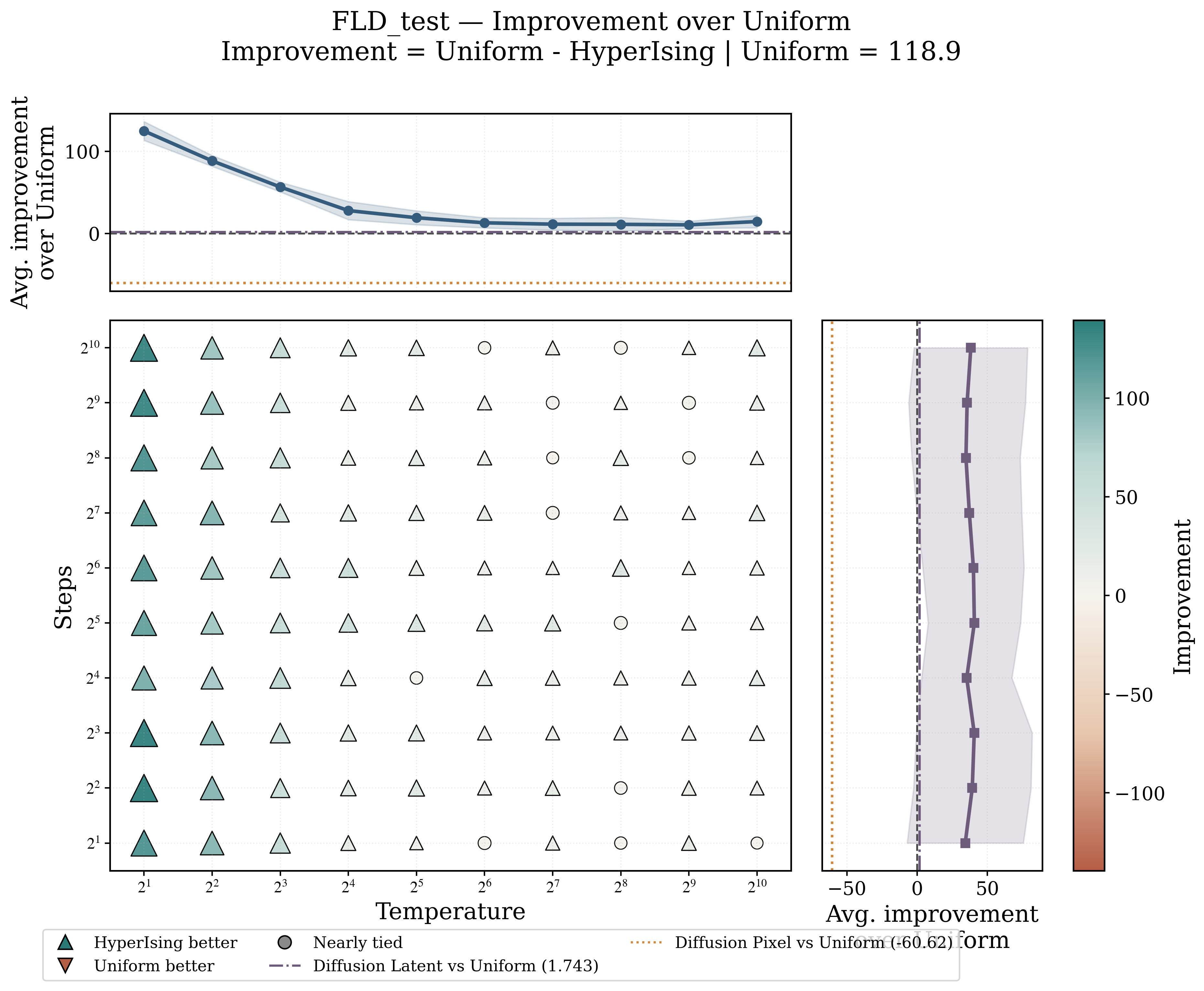}\end{minipage}\par\medskip
\begin{minipage}[t]{.48\linewidth}\centering \textbf{Validation codes}\par\smallskip \includegraphics[width=\linewidth]{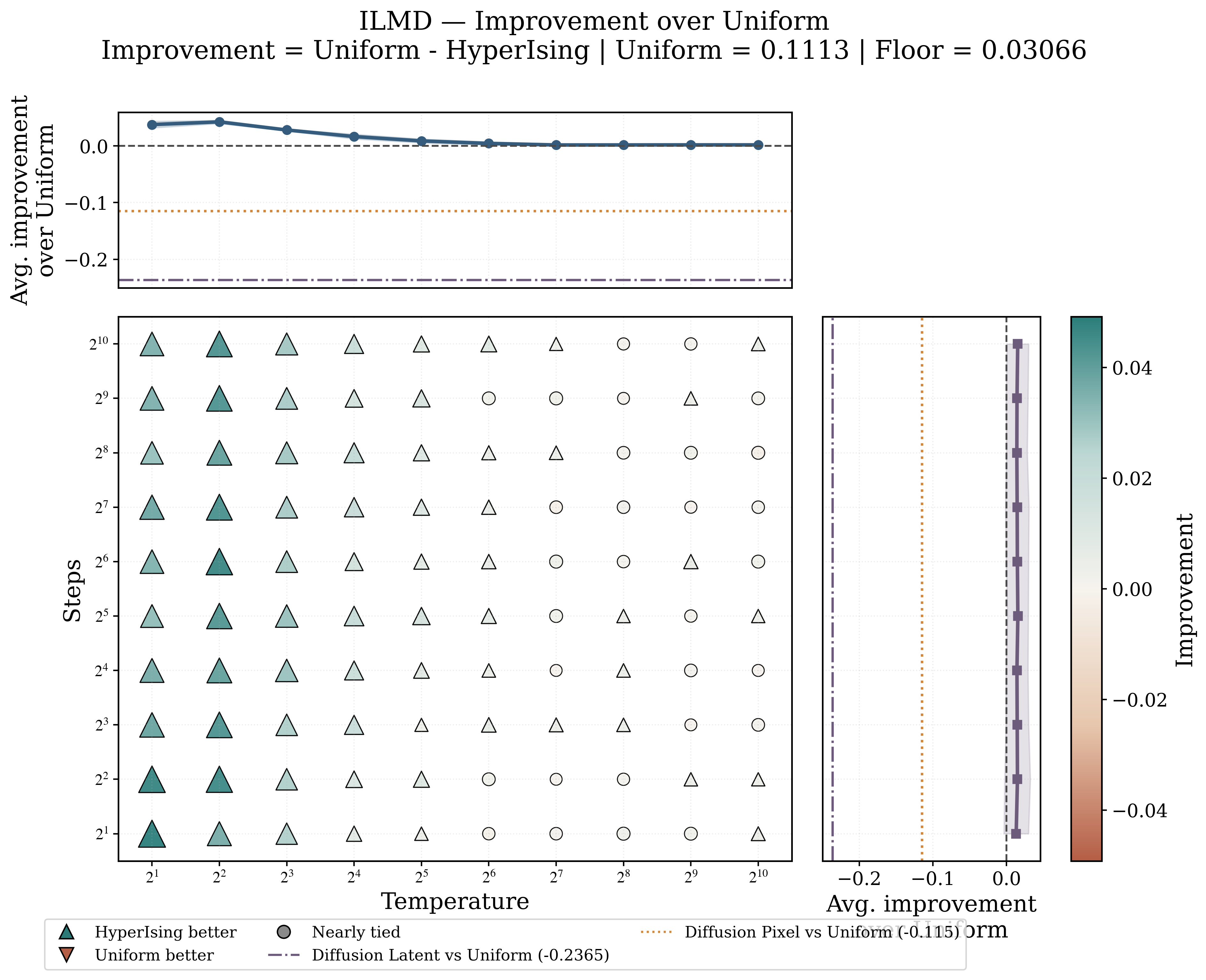}\end{minipage}
\caption{16-bit cat sampling landscape. Original-reference FLD, reconstruction-reference FLD, and ILMD to validation codes. Positive improvements indicate lower values than uniform. The diffusion reference lines show the corresponding metric contrasts with uniform sampling. Each panel has its own scale.}\label{fig:cats16-supp}\end{figure}

\begin{figure}[!tbp]\centering
\begin{minipage}[t]{.48\linewidth}\centering \textbf{Originals}\par\smallskip \includegraphics[width=\linewidth]{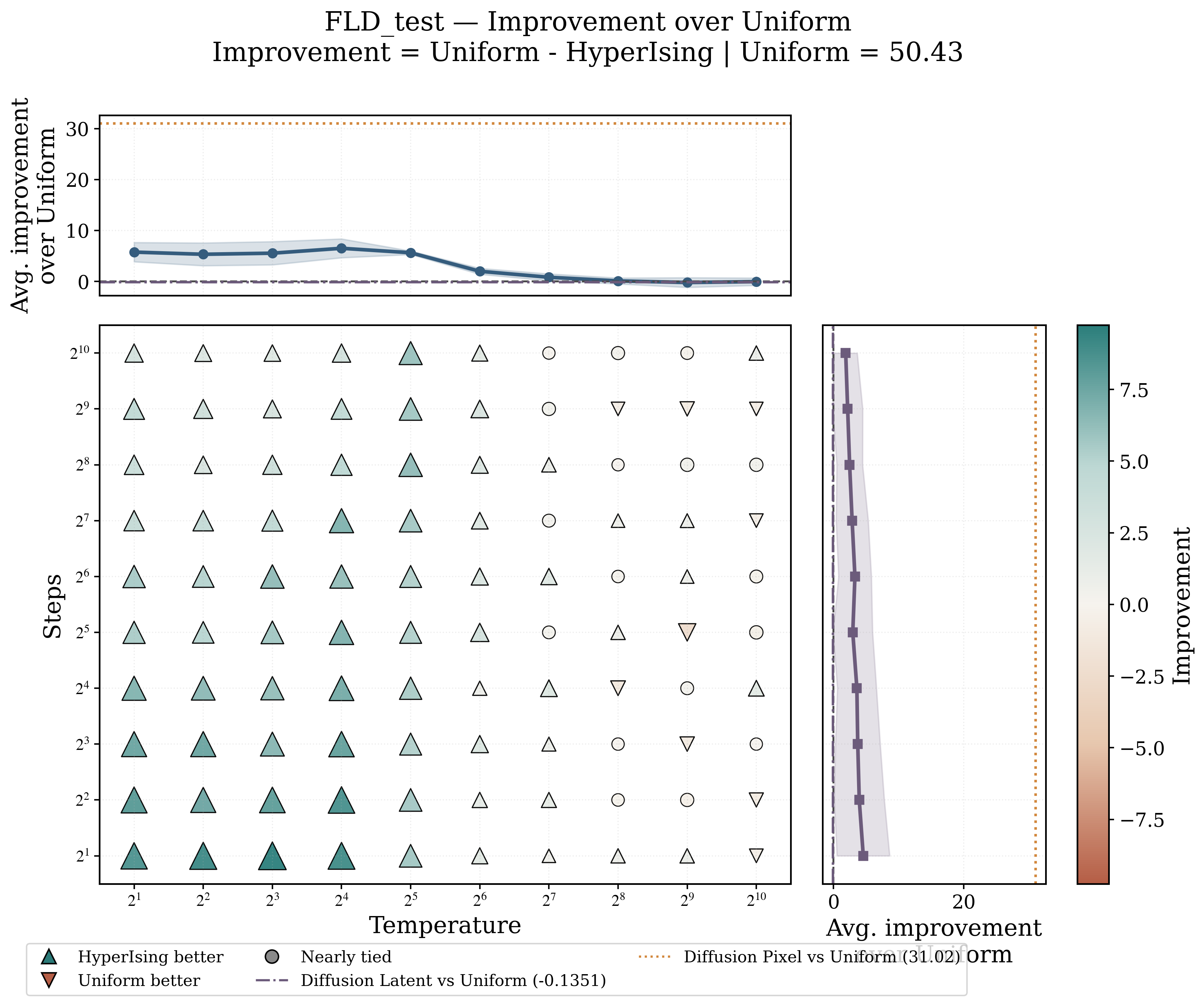}\end{minipage}\hfill \begin{minipage}[t]{.48\linewidth}\centering \textbf{Reconstructions}\par\smallskip \includegraphics[width=\linewidth]{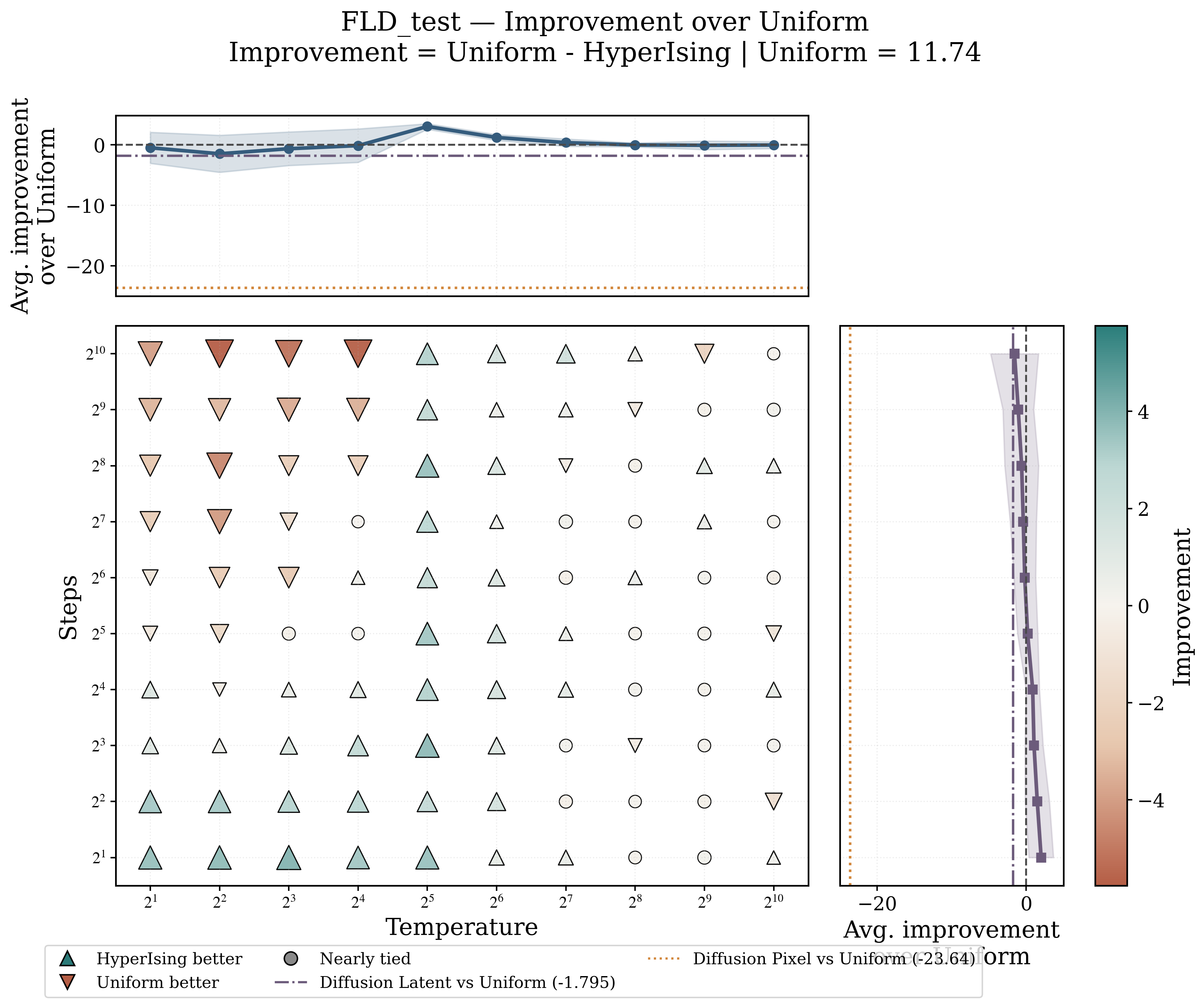}\end{minipage}\par\medskip
\begin{minipage}[t]{.48\linewidth}\centering \textbf{Validation codes}\par\smallskip \includegraphics[width=\linewidth]{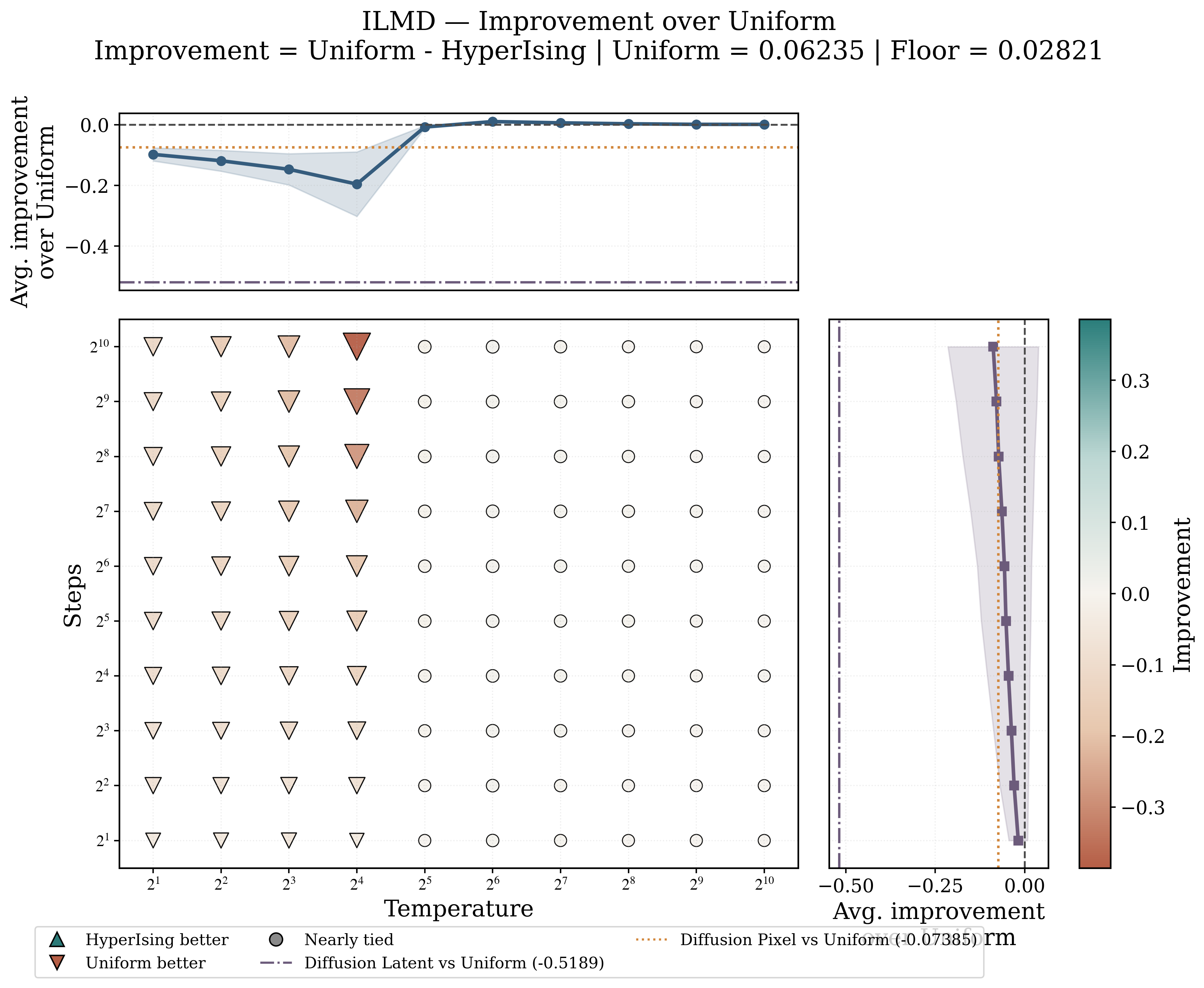}\end{minipage}
\caption{64-bit cat sampling landscape. Original-reference FLD, reconstruction-reference FLD, and ILMD to validation codes. Positive improvements indicate lower values than uniform. The diffusion reference lines show the corresponding metric contrasts with uniform sampling. Each panel has its own scale.}\label{fig:cats64-supp}\end{figure}

\begin{figure}[!tbp]\centering
\begin{minipage}[t]{.48\linewidth}\centering \textbf{Originals}\par\smallskip \includegraphics[width=\linewidth]{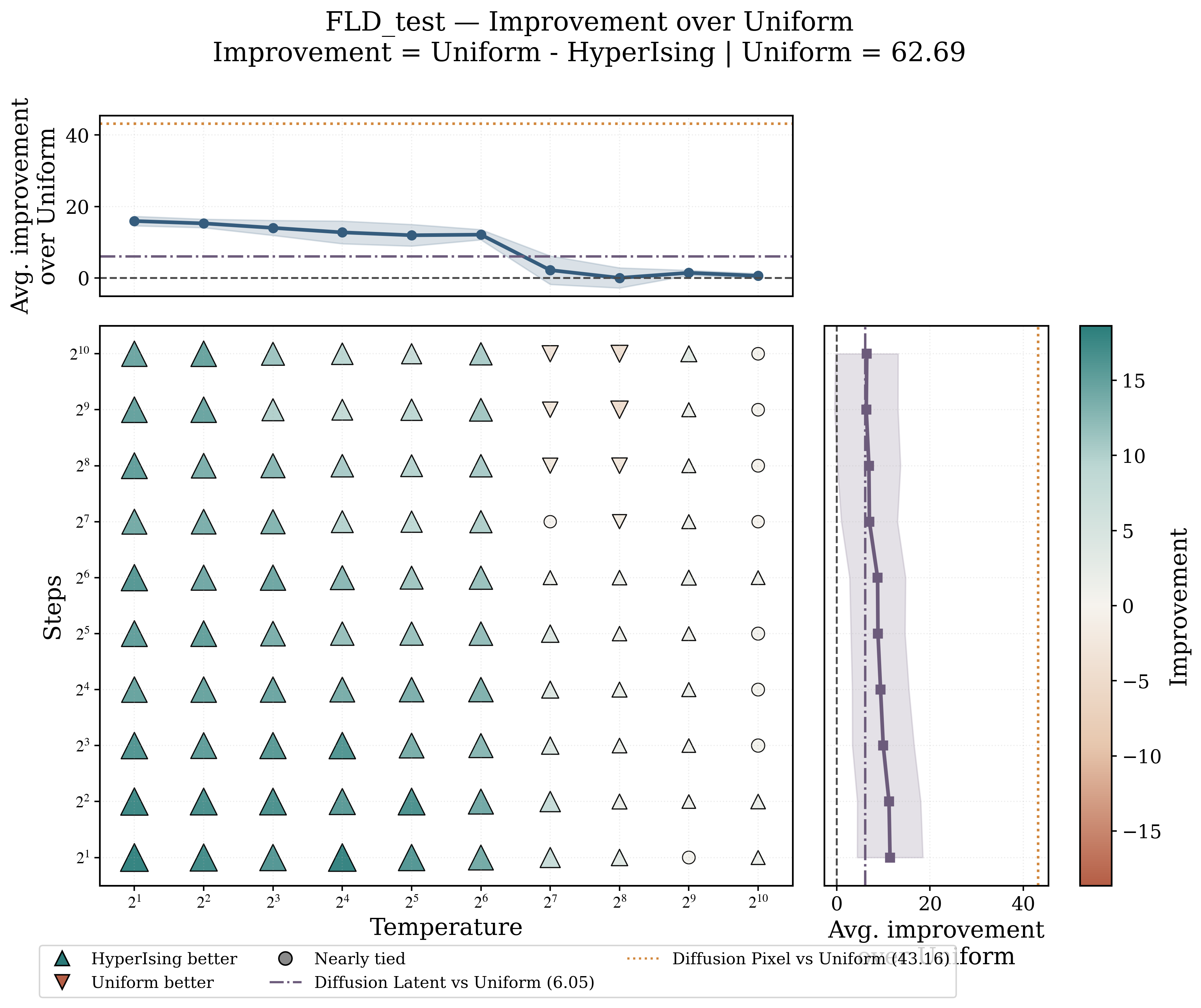}\end{minipage}\hfill \begin{minipage}[t]{.48\linewidth}\centering \textbf{Reconstructions}\par\smallskip \includegraphics[width=\linewidth]{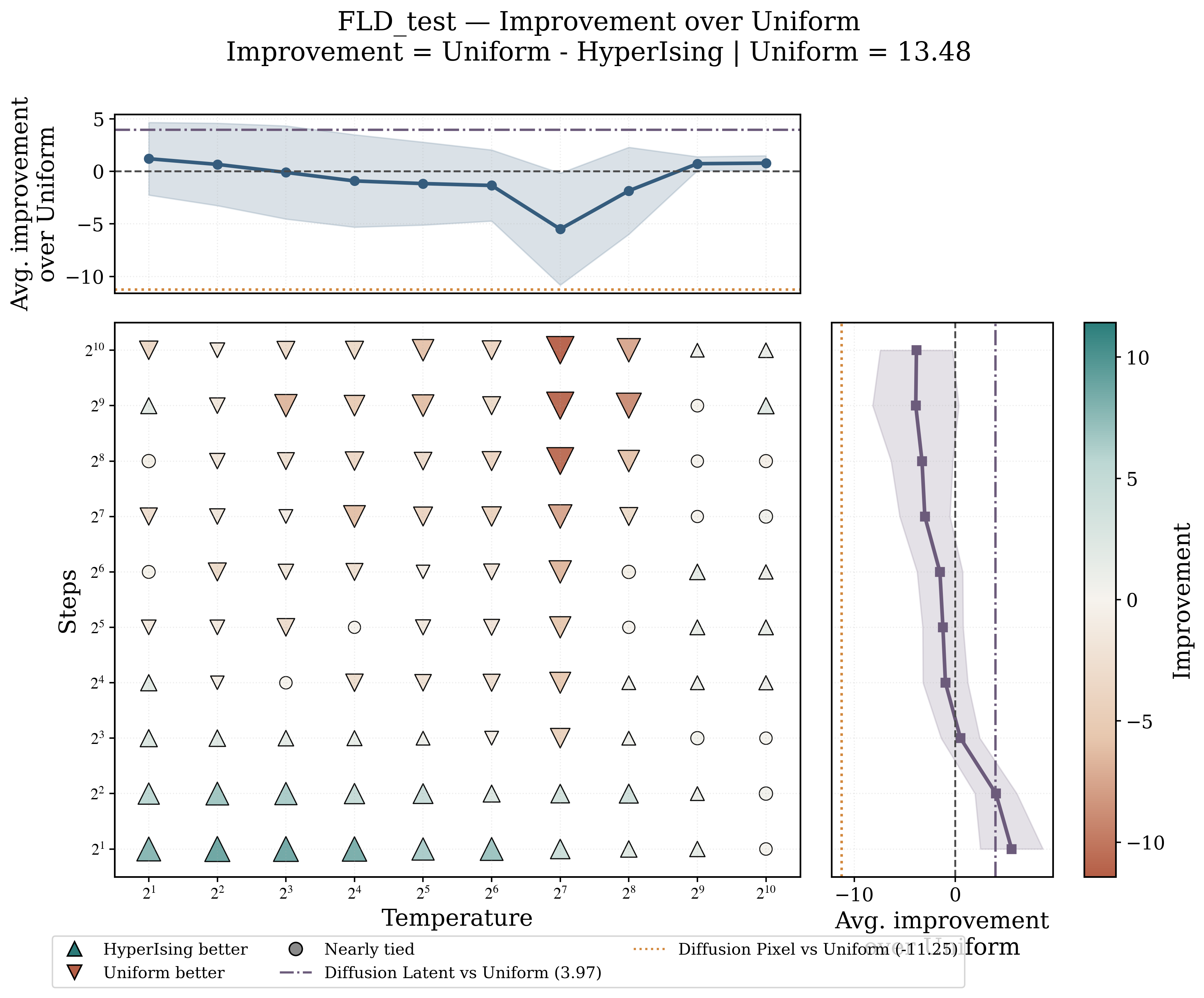}\end{minipage}\par\medskip
\begin{minipage}[t]{.48\linewidth}\centering \textbf{Validation codes}\par\smallskip \includegraphics[width=\linewidth]{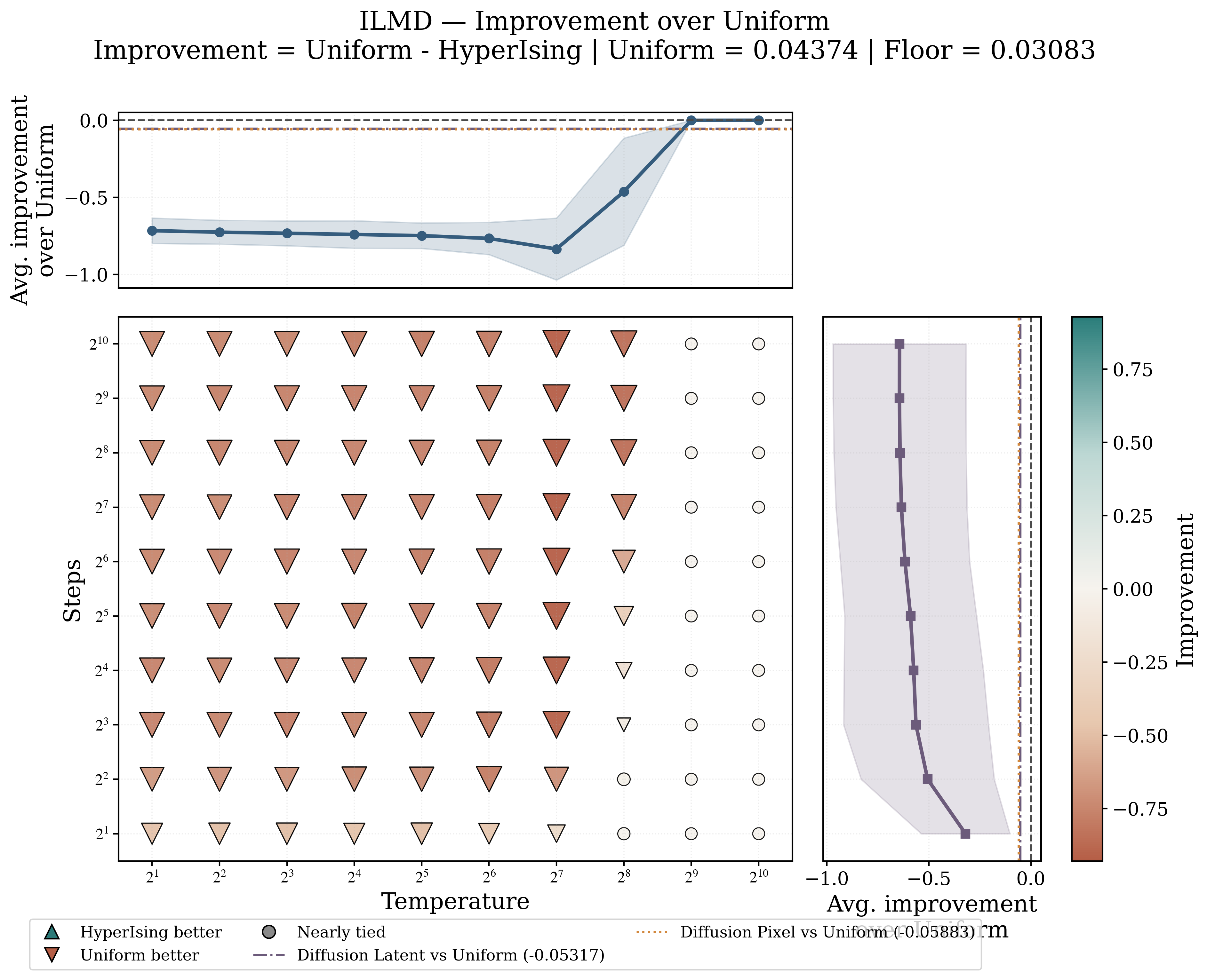}\end{minipage}
\caption{256-bit cat sampling landscape. Original-reference FLD, reconstruction-reference FLD, and ILMD to validation codes. Positive improvements indicate lower values than uniform. The diffusion reference lines show the corresponding metric contrasts with uniform sampling. Each panel has its own scale.}\label{fig:cats256-supp}\end{figure}

\begin{figure}[!tbp]\centering
\begin{minipage}[t]{.48\linewidth}\centering \textbf{16 bits}\par\smallskip \includegraphics[width=\linewidth]{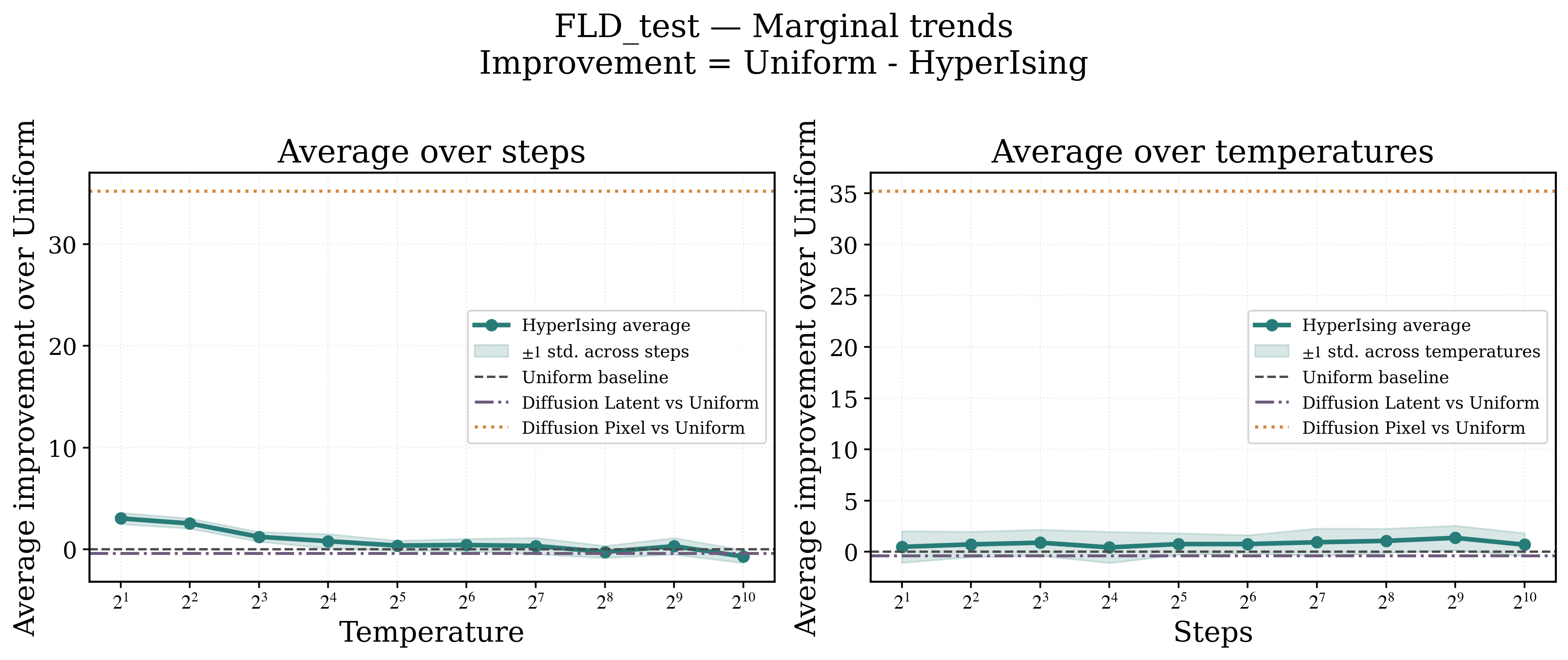}\end{minipage}\hfill \begin{minipage}[t]{.48\linewidth}\centering \textbf{32 bits}\par\smallskip \includegraphics[width=\linewidth]{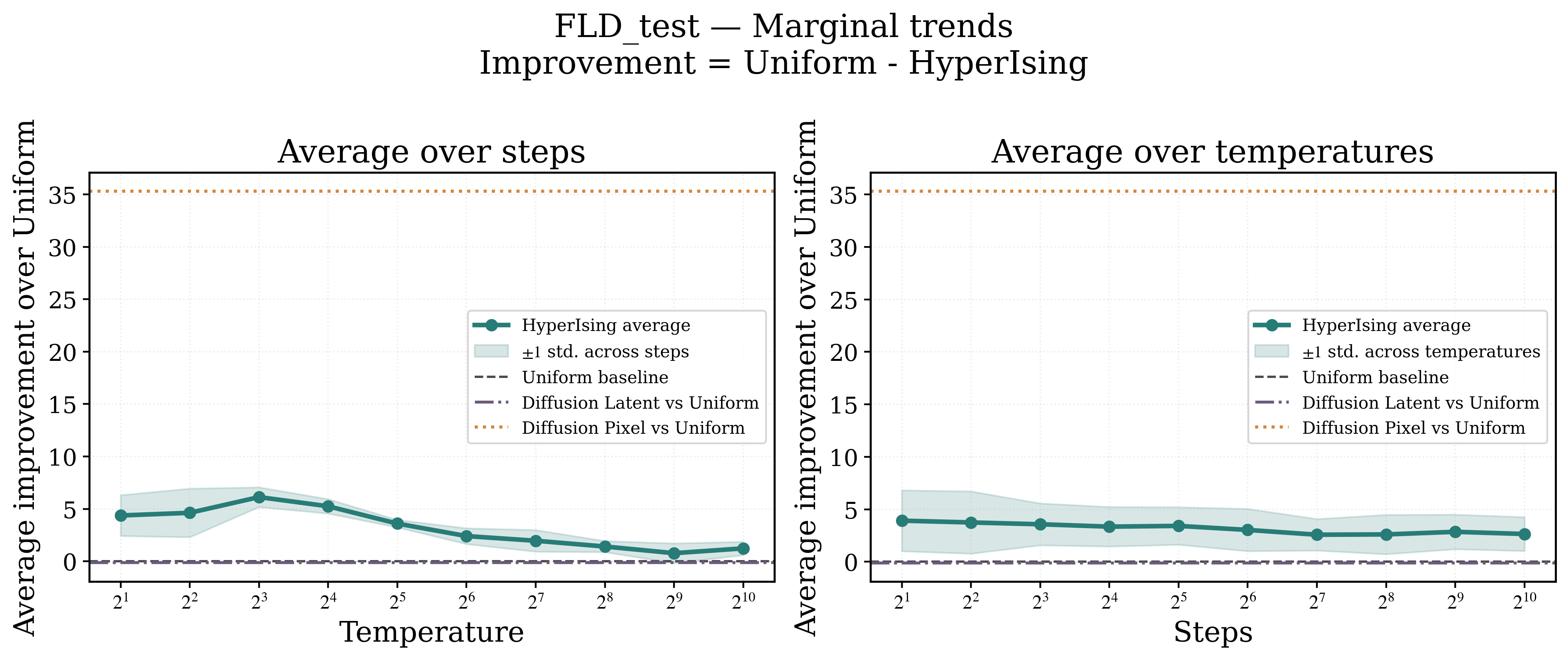}\end{minipage}\par\medskip
\begin{minipage}[t]{.48\linewidth}\centering \textbf{64 bits}\par\smallskip \includegraphics[width=\linewidth]{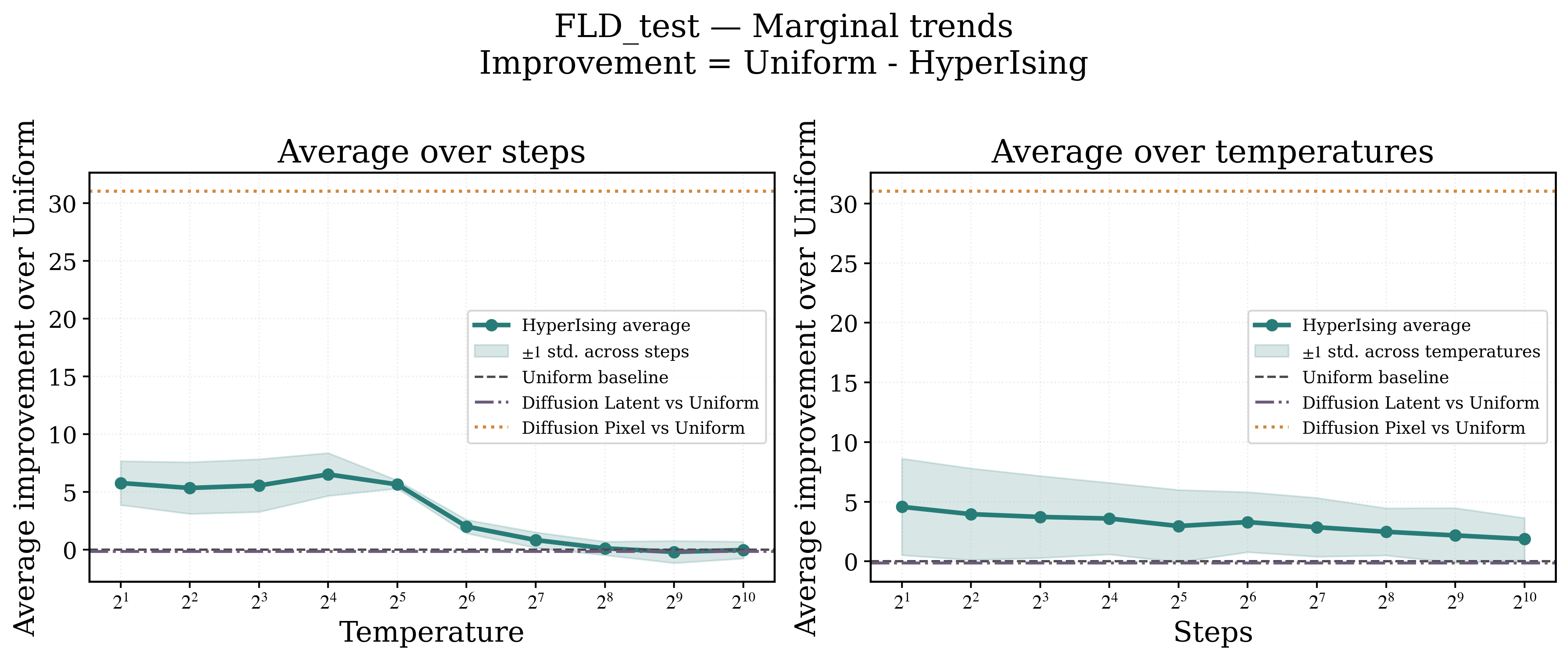}\end{minipage}\hfill \begin{minipage}[t]{.48\linewidth}\centering \textbf{128 bits}\par\smallskip \includegraphics[width=\linewidth]{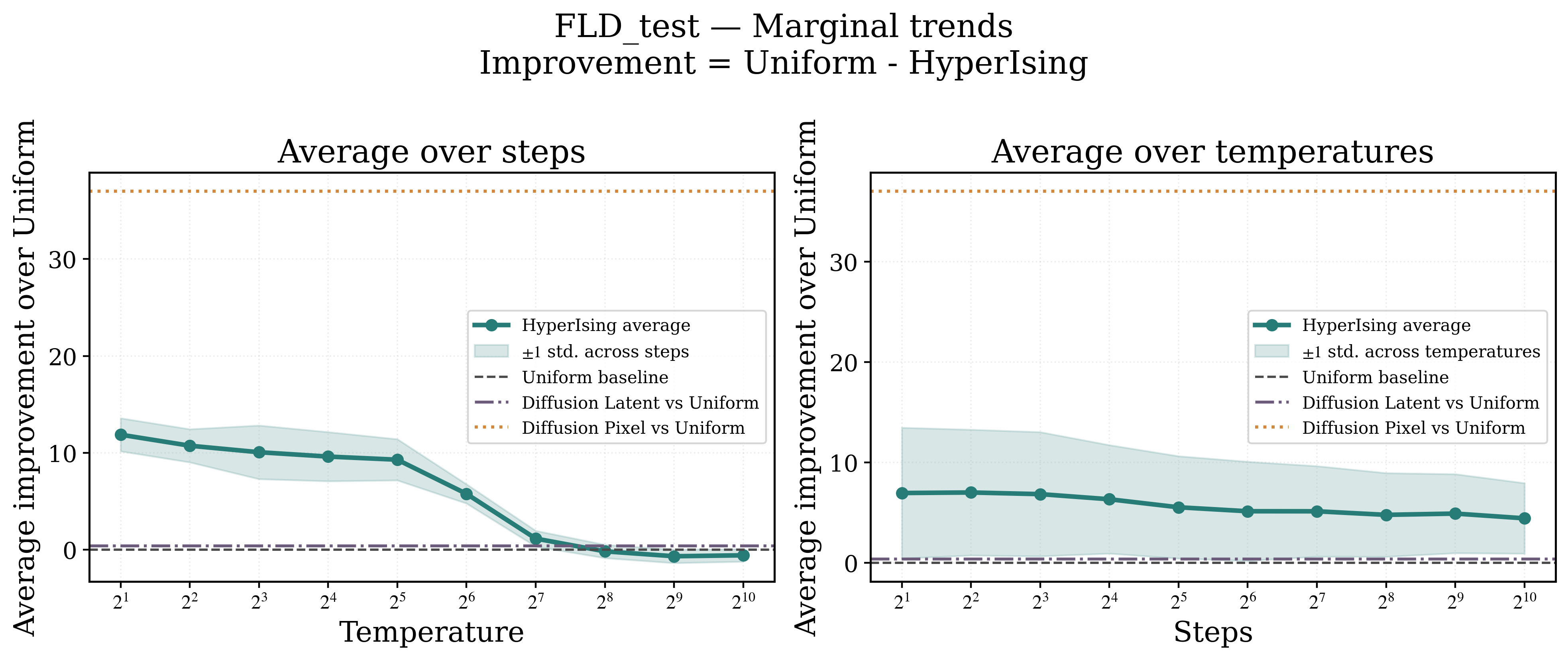}\end{minipage}\par\medskip
\begin{minipage}[t]{.48\linewidth}\centering \textbf{256 bits}\par\smallskip \includegraphics[width=\linewidth]{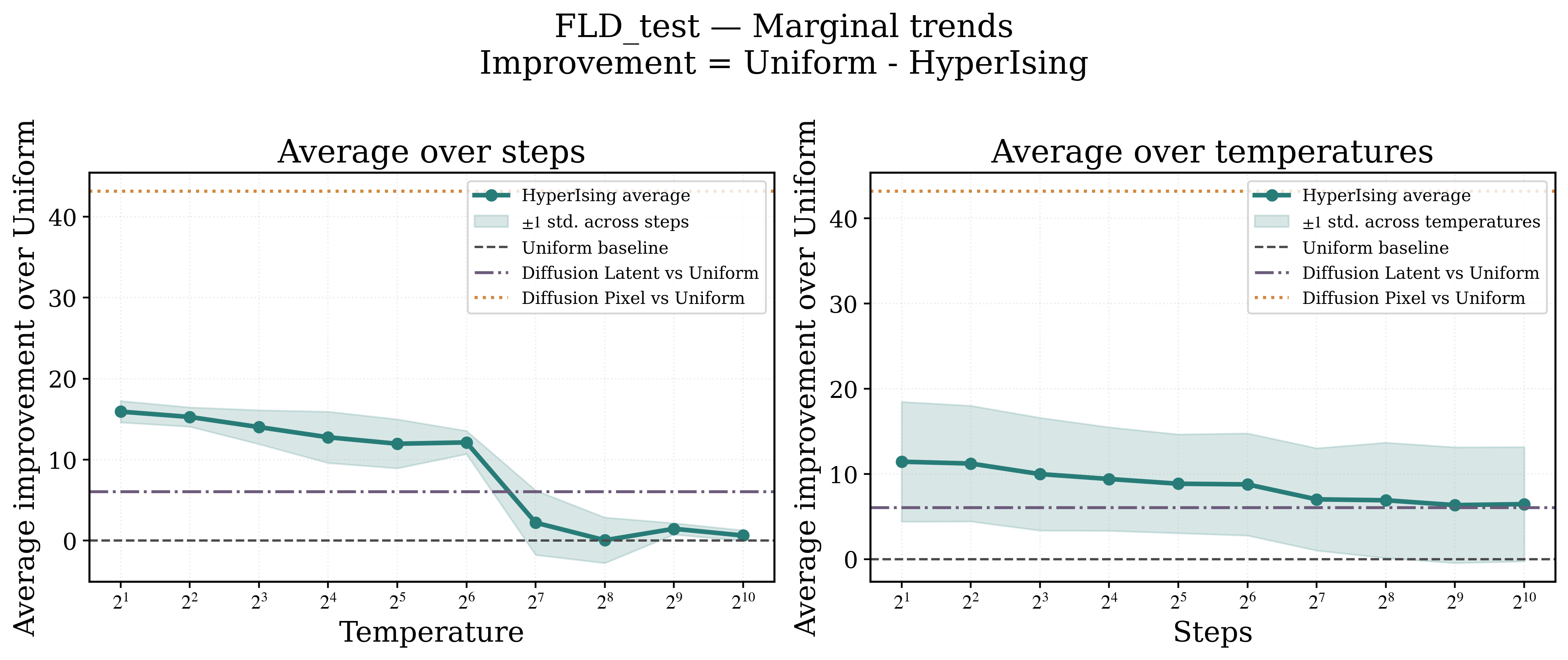}\end{minipage}\hfill \begin{minipage}[t]{.48\linewidth}\centering \textbf{512 bits}\par\smallskip \includegraphics[width=\linewidth]{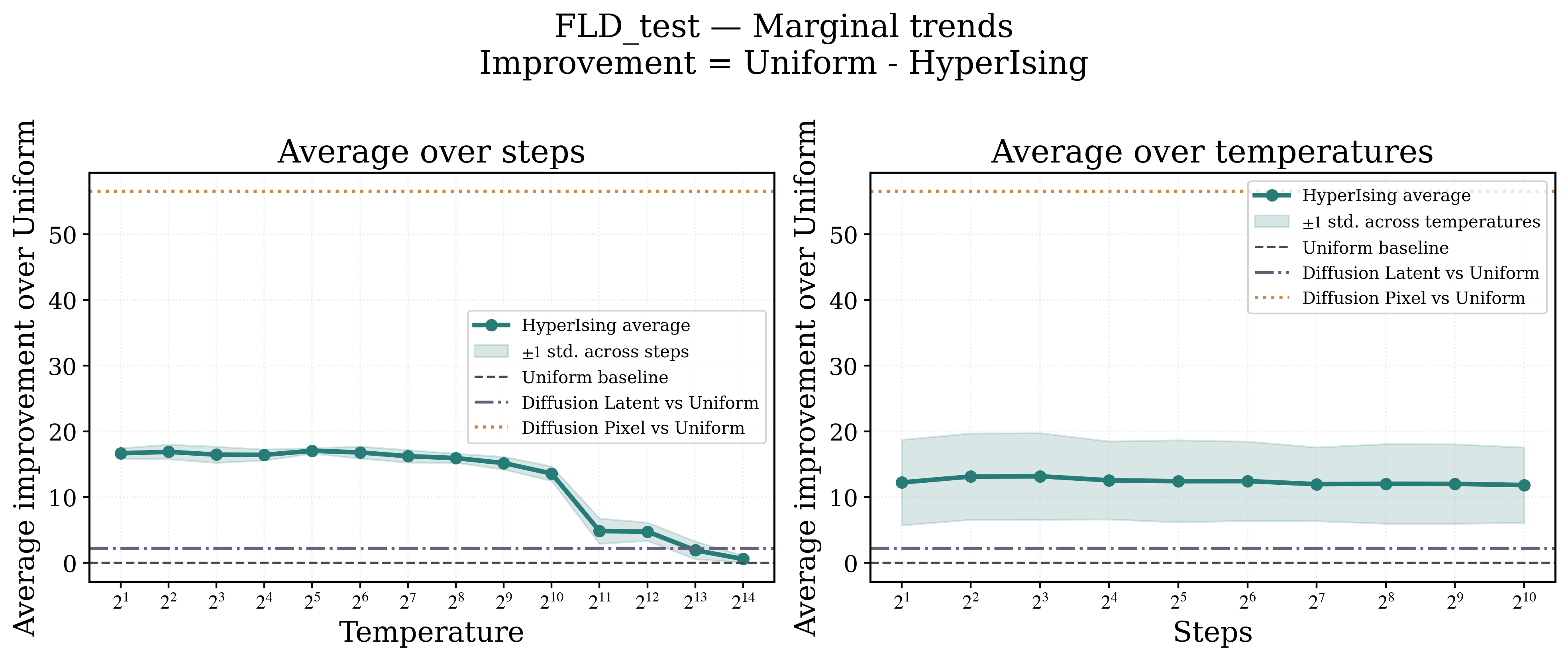}\end{minipage}
\caption{Marginal trends of FLD against original images. For each dimension, the left half averages improvement over sweep counts at each temperature and the right half over temperatures at each sweep count. Bands are one standard deviation across those settings. Dashed and dotted lines identify fixed comparators. The 512-bit range includes additional temperatures.}\label{fig:cats-marginal-real}\end{figure}

\begin{figure}[!tbp]\centering
\begin{minipage}[t]{.48\linewidth}\centering \textbf{16 bits}\par\smallskip \includegraphics[width=\linewidth]{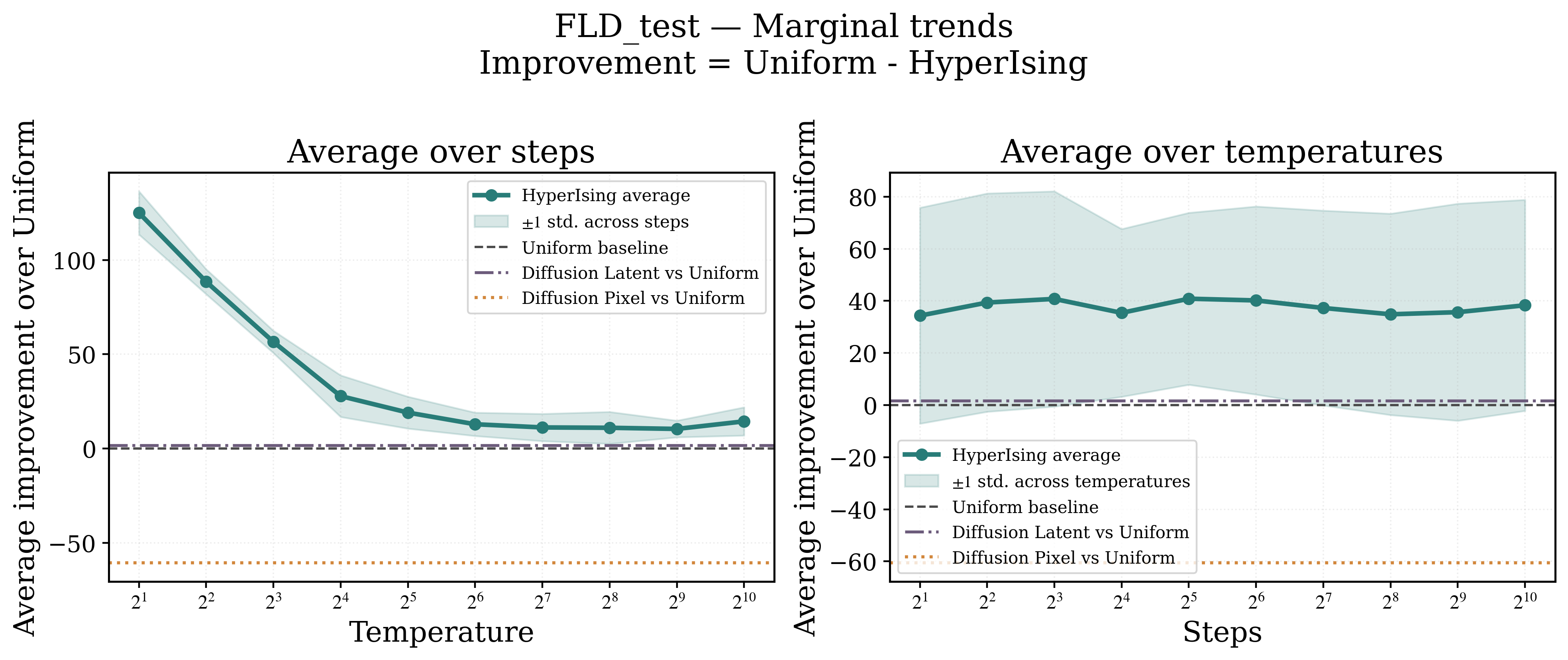}\end{minipage}\hfill \begin{minipage}[t]{.48\linewidth}\centering \textbf{32 bits}\par\smallskip \includegraphics[width=\linewidth]{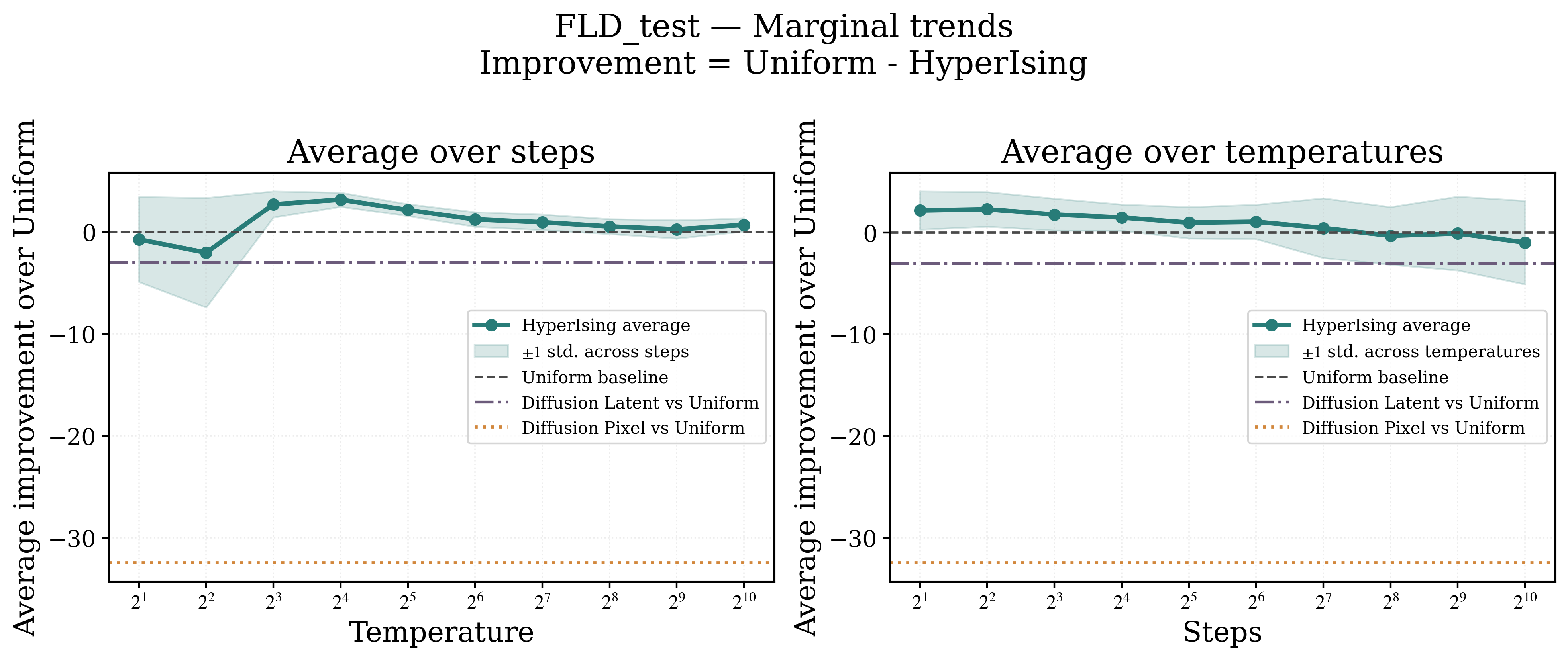}\end{minipage}\par\medskip
\begin{minipage}[t]{.48\linewidth}\centering \textbf{64 bits}\par\smallskip \includegraphics[width=\linewidth]{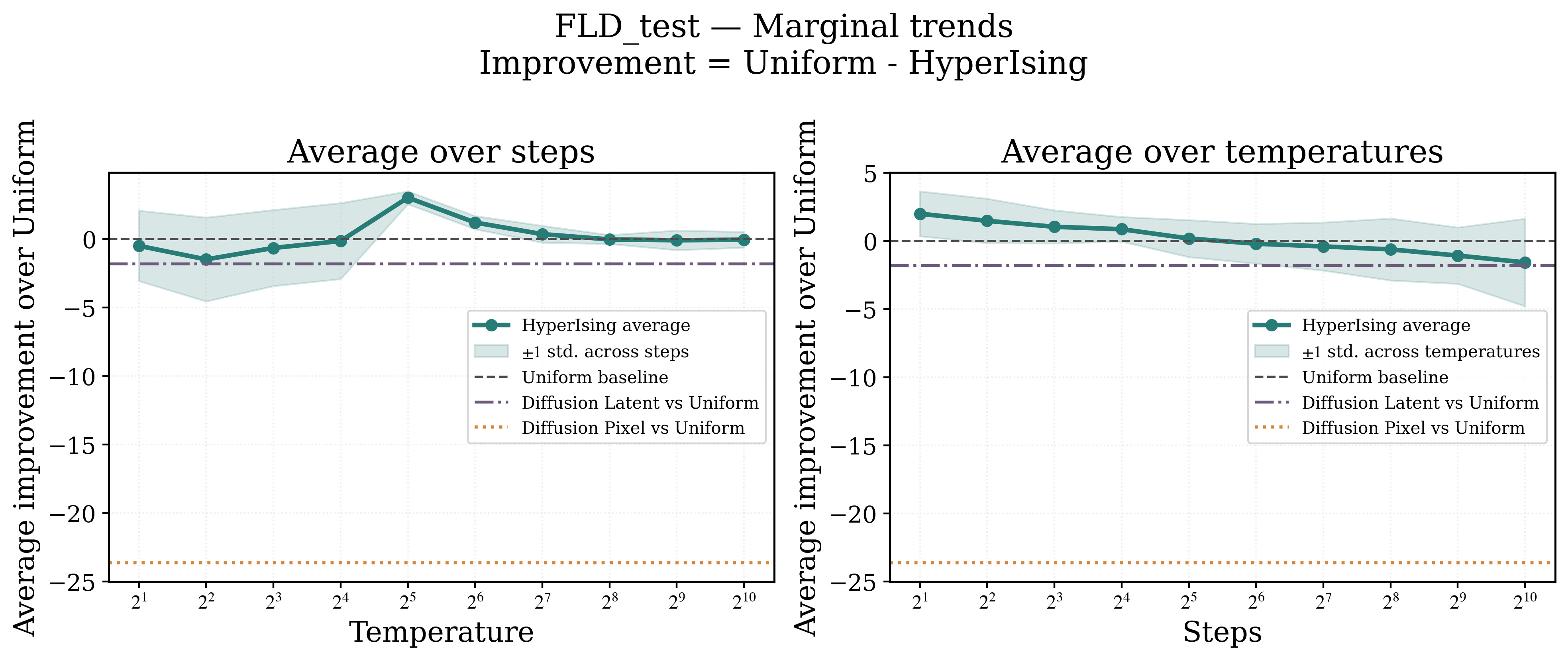}\end{minipage}\hfill \begin{minipage}[t]{.48\linewidth}\centering \textbf{128 bits}\par\smallskip \includegraphics[width=\linewidth]{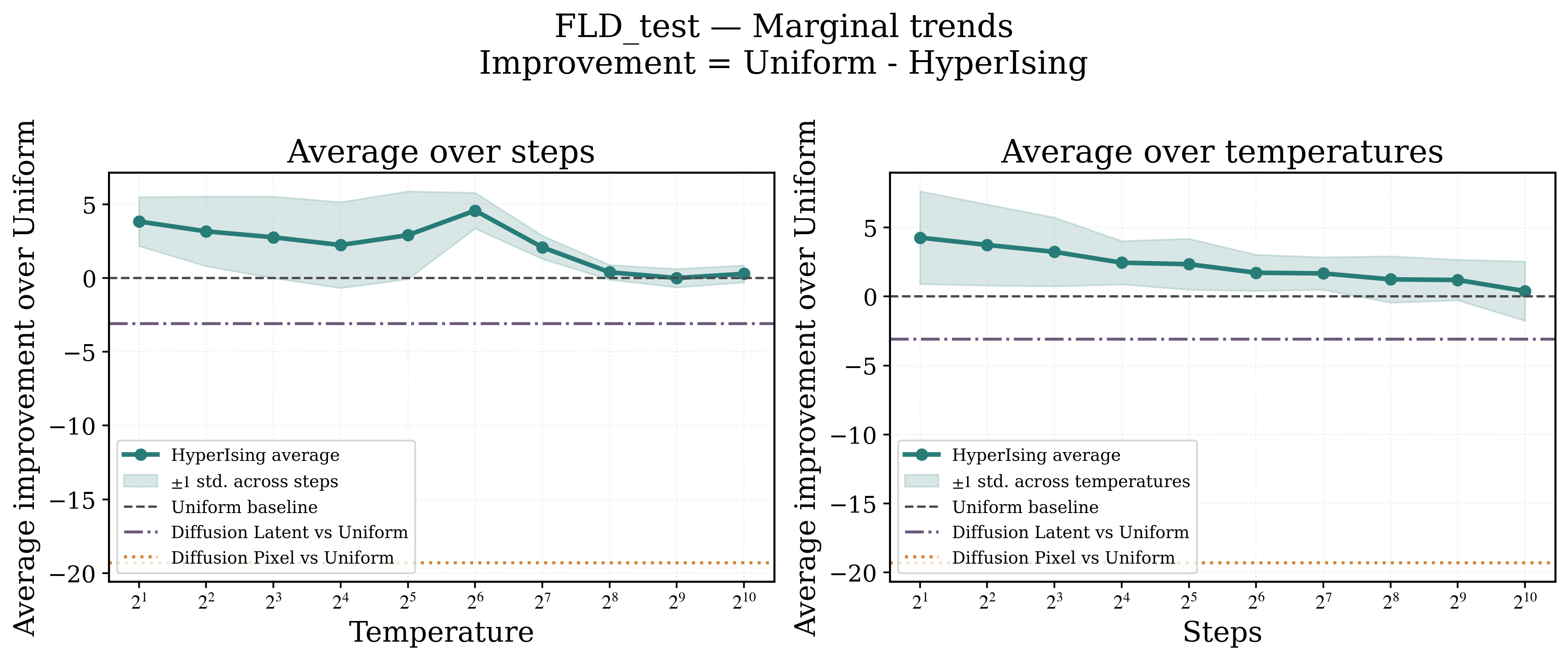}\end{minipage}\par\medskip
\begin{minipage}[t]{.48\linewidth}\centering \textbf{256 bits}\par\smallskip \includegraphics[width=\linewidth]{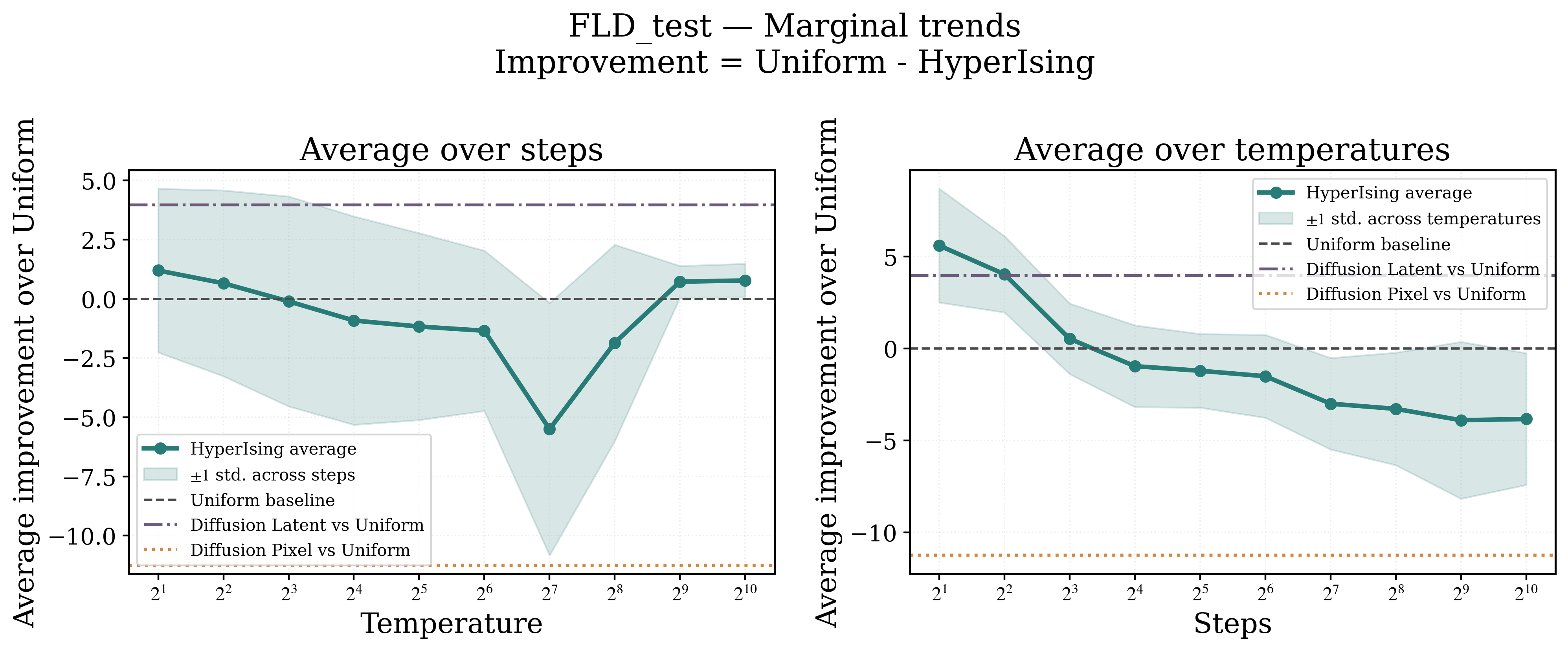}\end{minipage}\hfill \begin{minipage}[t]{.48\linewidth}\centering \textbf{512 bits}\par\smallskip \includegraphics[width=\linewidth]{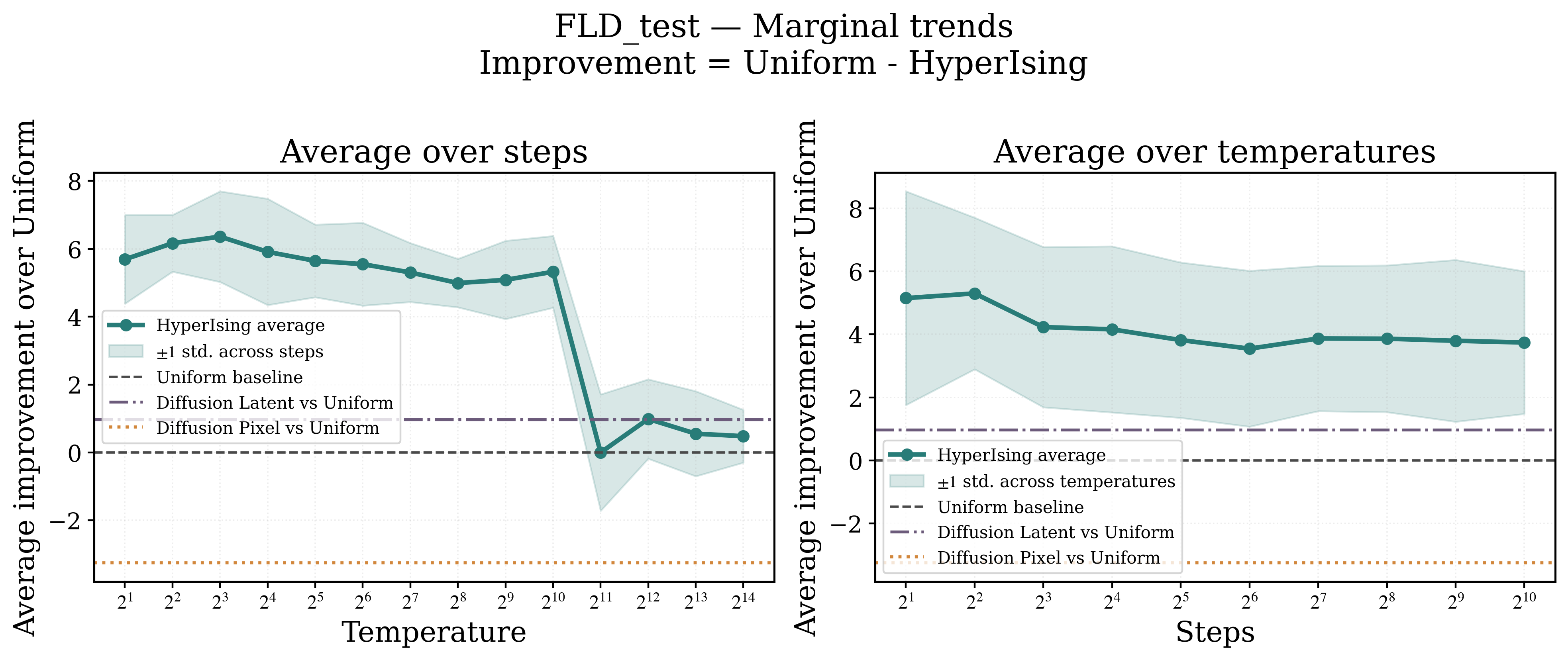}\end{minipage}
\caption{Marginal trends of FLD against autoencoder reconstructions. For each dimension, the left half averages improvement over sweep counts at each temperature and the right half over temperatures at each sweep count. Bands are one standard deviation across those settings. Dashed and dotted lines identify fixed comparators. The 512-bit range includes additional temperatures.}\label{fig:cats-marginal-reco}\end{figure}

\begin{figure}[!tbp]\centering
\begin{minipage}[t]{.48\linewidth}\centering \textbf{16 bits}\par\smallskip \includegraphics[width=\linewidth]{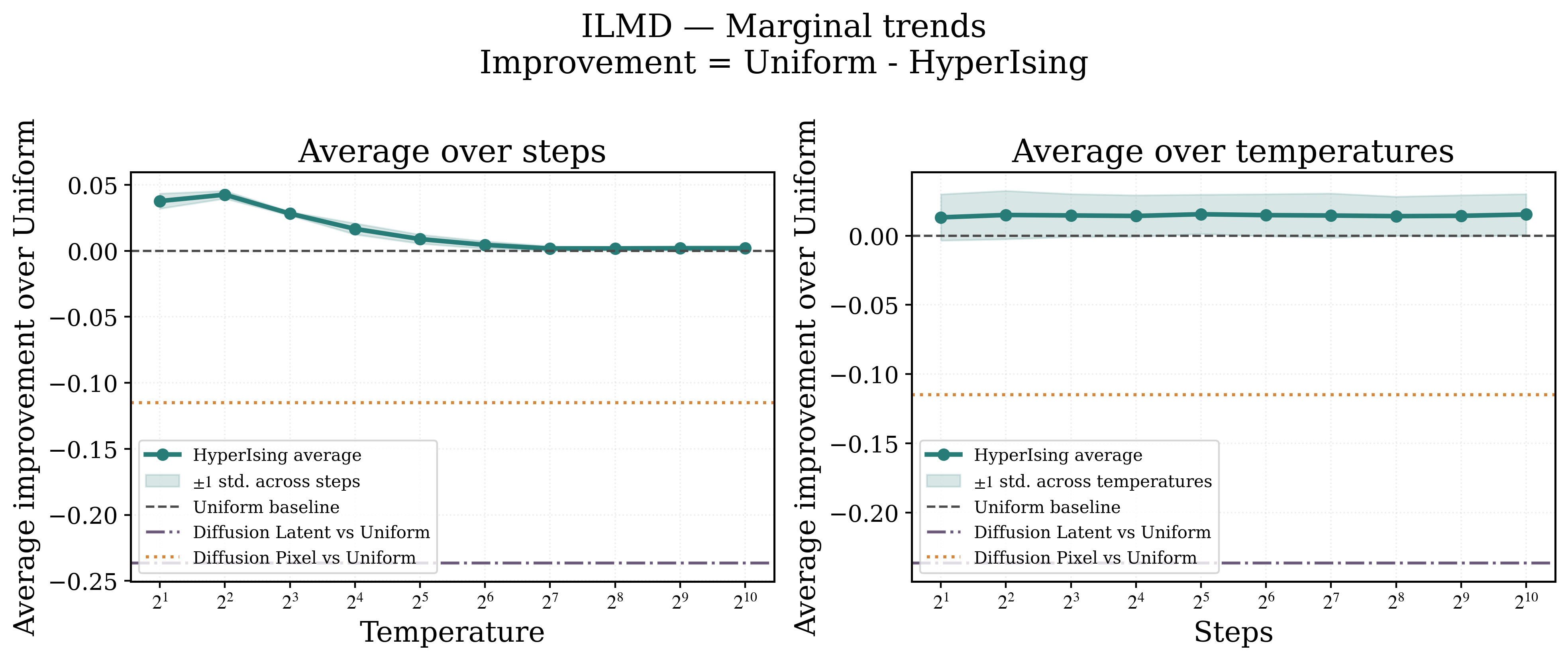}\end{minipage}\hfill \begin{minipage}[t]{.48\linewidth}\centering \textbf{32 bits}\par\smallskip \includegraphics[width=\linewidth]{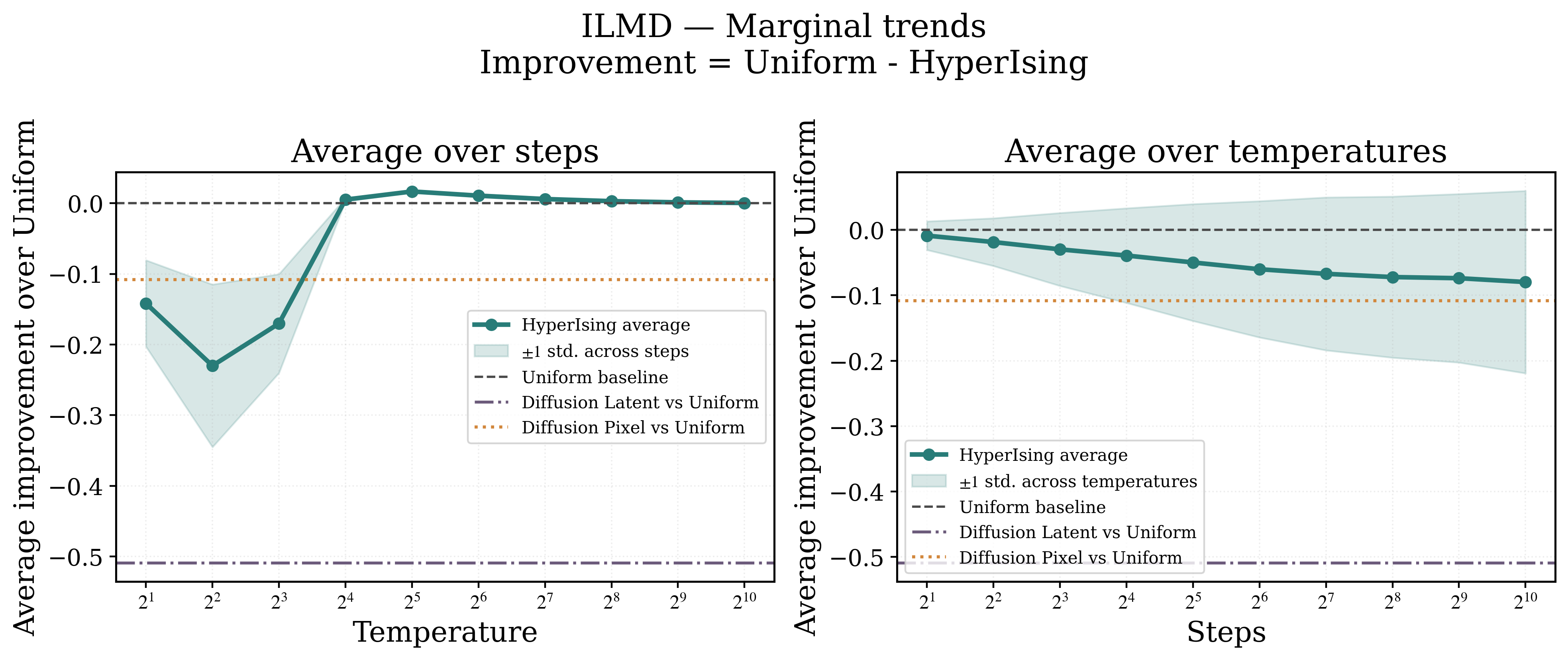}\end{minipage}\par\medskip
\begin{minipage}[t]{.48\linewidth}\centering \textbf{64 bits}\par\smallskip \includegraphics[width=\linewidth]{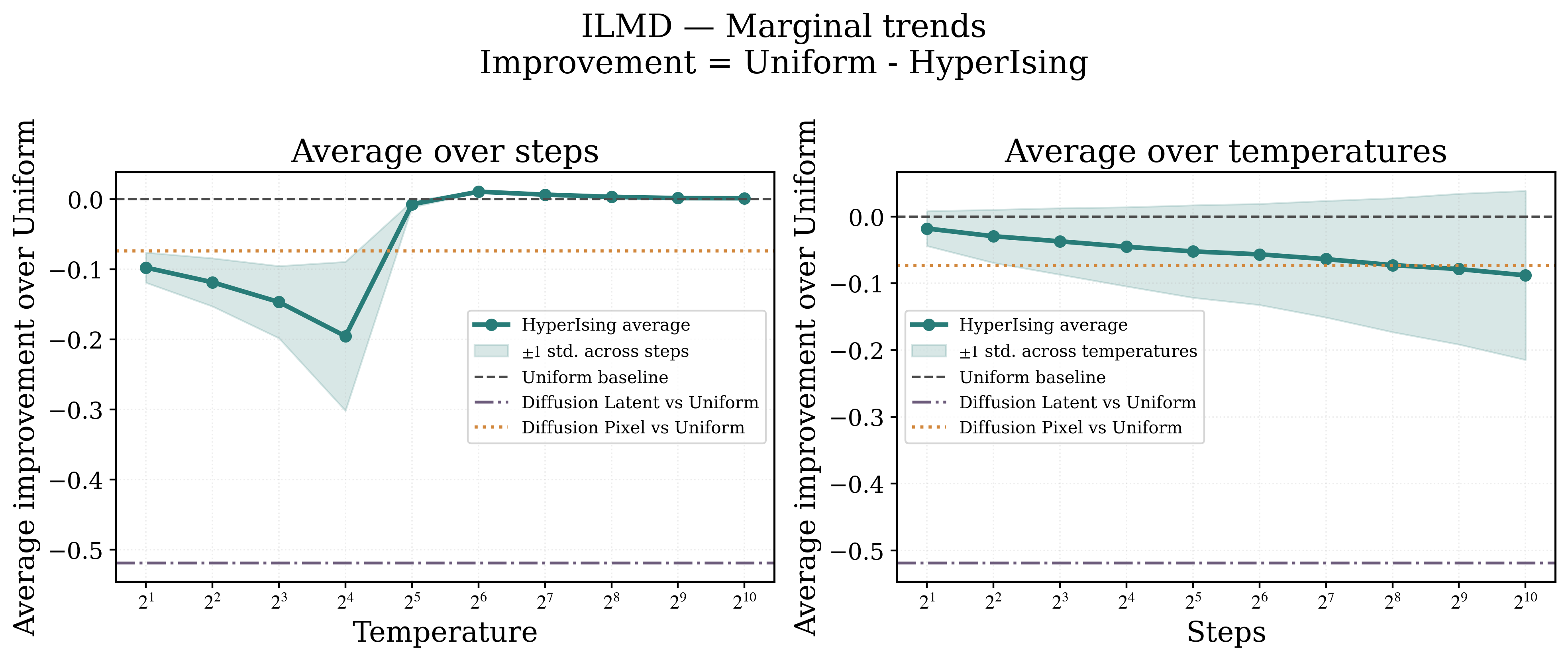}\end{minipage}\hfill \begin{minipage}[t]{.48\linewidth}\centering \textbf{128 bits}\par\smallskip \includegraphics[width=\linewidth]{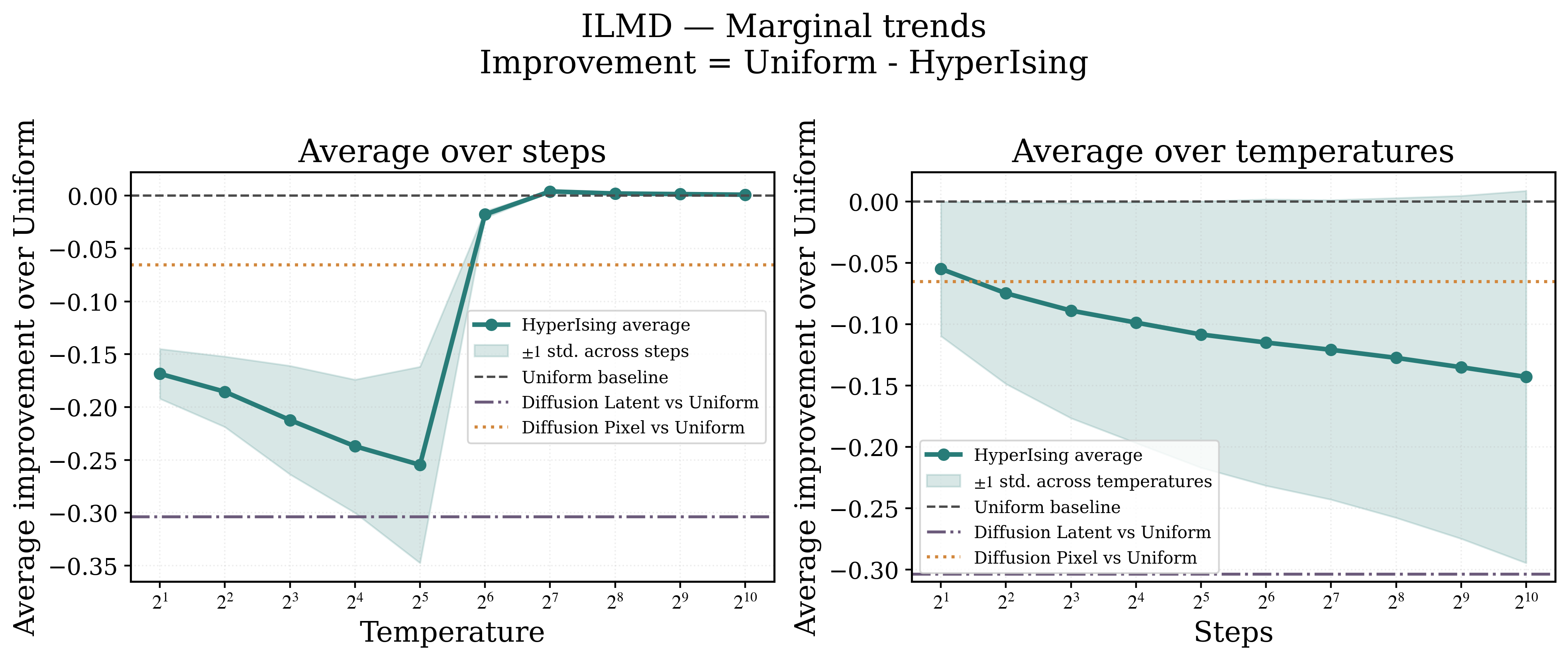}\end{minipage}\par\medskip
\begin{minipage}[t]{.48\linewidth}\centering \textbf{256 bits}\par\smallskip \includegraphics[width=\linewidth]{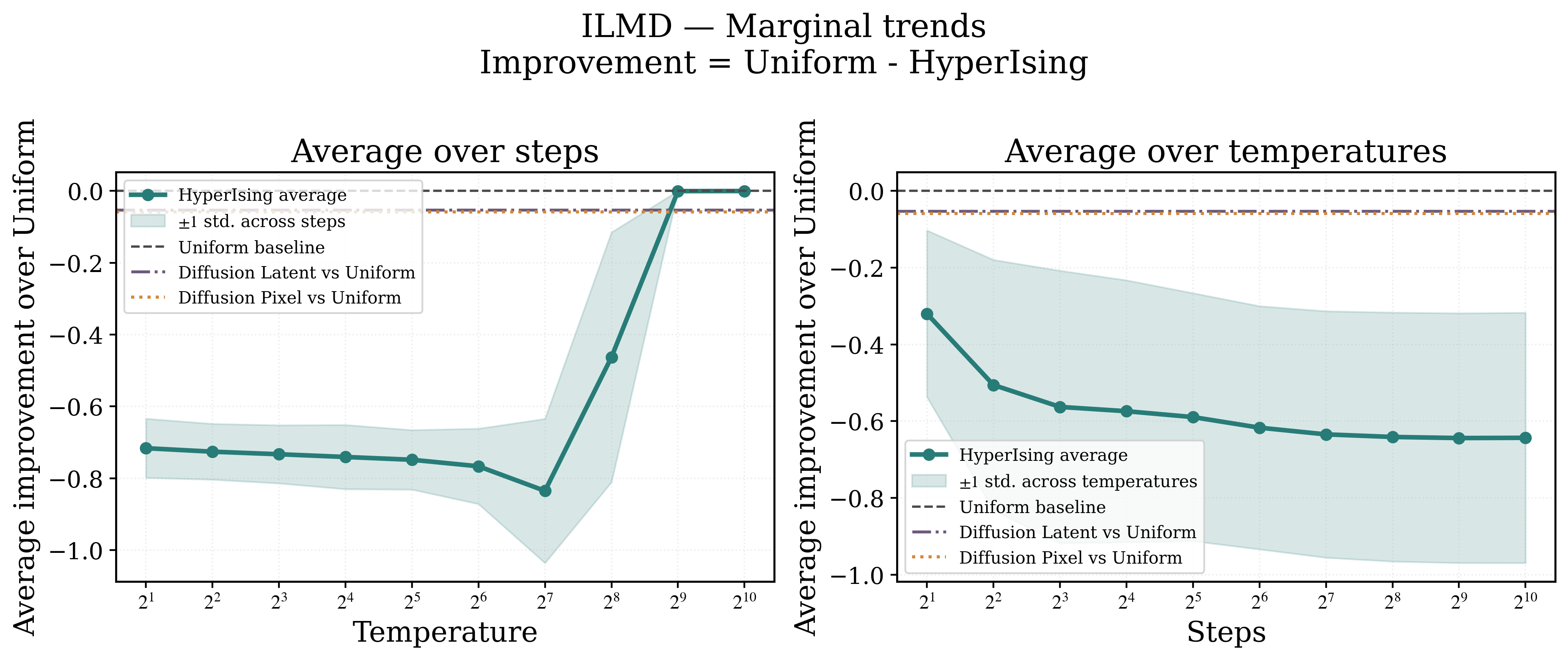}\end{minipage}\hfill \begin{minipage}[t]{.48\linewidth}\centering \textbf{512 bits}\par\smallskip \includegraphics[width=\linewidth]{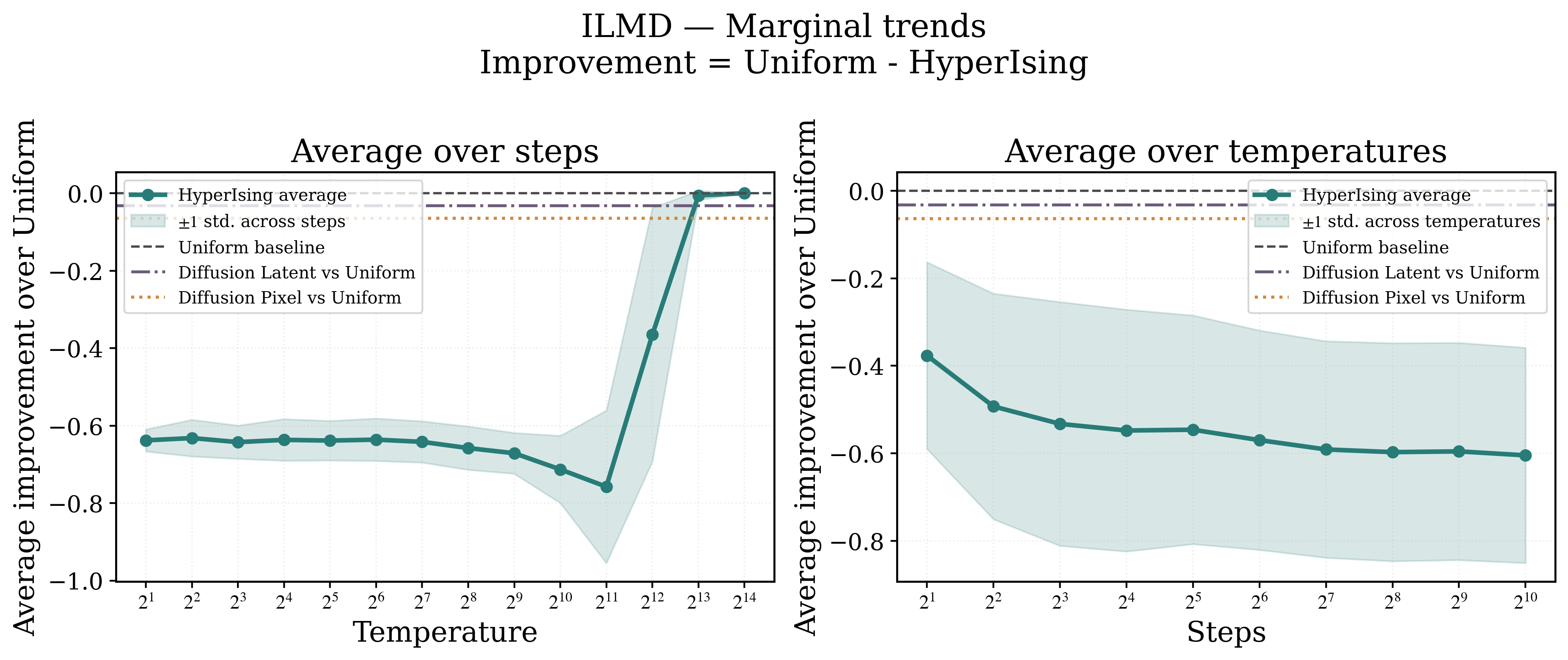}\end{minipage}
\caption{Marginal trends of ILMD to validation codes. For each dimension, the left half averages improvement over sweep counts at each temperature and the right half over temperatures at each sweep count. Bands are one standard deviation across those settings. Dashed and dotted lines identify fixed comparators. The 512-bit range includes additional temperatures.}\label{fig:cats-marginal-ilmd}\end{figure}

Figure~\ref{fig:cats-maps-extra} complements Fig.~\ref{fig:cats-maps} with reconstruction-reference FLD and ILMD at 32, 128, and 512 bits. Figures~\ref{fig:cats16-supp}--\ref{fig:cats256-supp} report all three metrics for the other dimensions. The expanded views in Figs.~\ref{fig:cats-marginal-real}--\ref{fig:cats-marginal-ilmd} show temperature and sweep-count marginals separately for all six dimensions.

\FloatBarrier
\subsection{Reference Reconstructions and Latent Distributions}
\begin{figure}[!tbp]\centering
\begin{minipage}[t]{.48\linewidth}\centering \textbf{FID to originals}\par\smallskip \includegraphics[width=\linewidth]{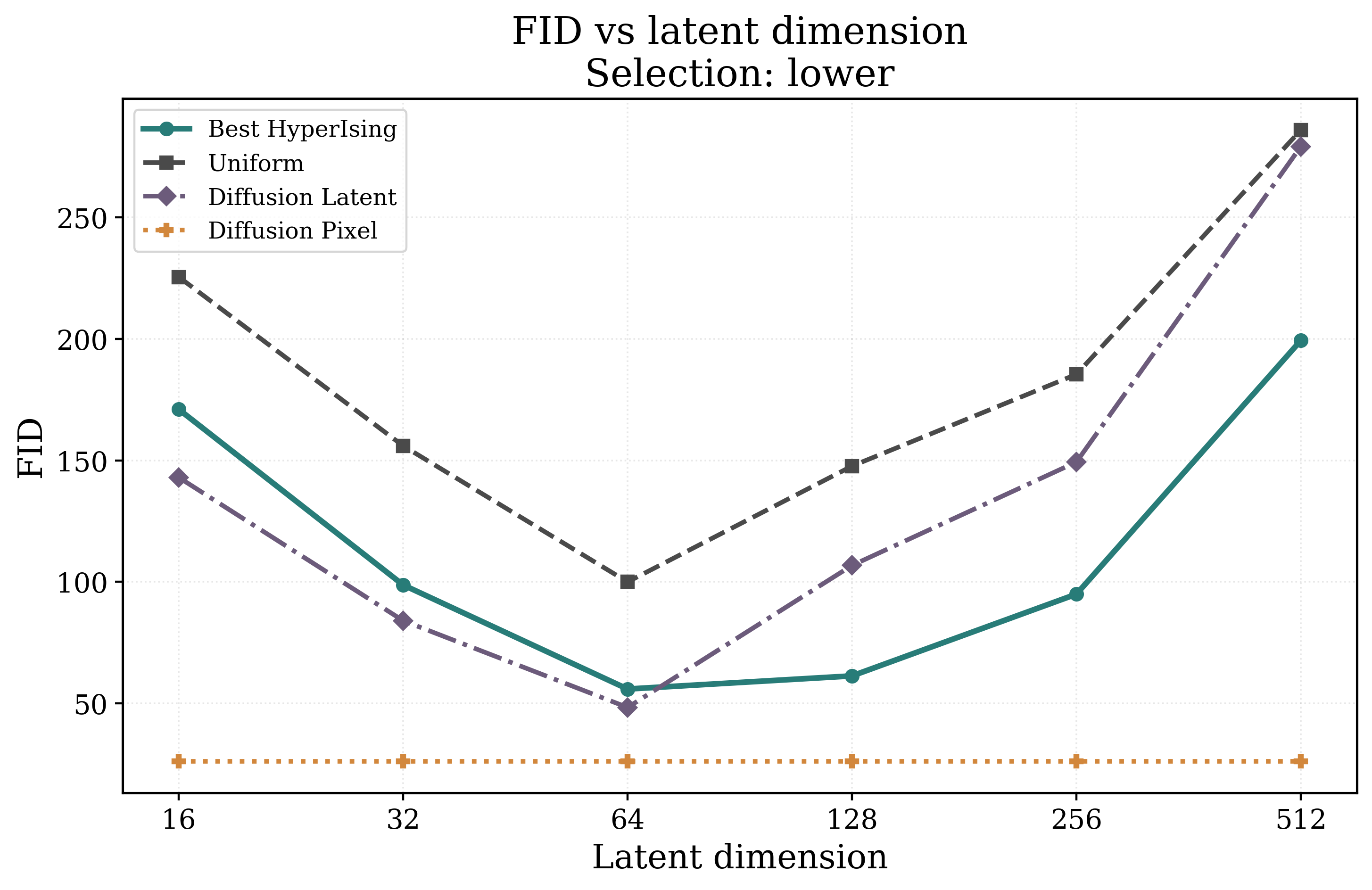}\end{minipage}\hfill \begin{minipage}[t]{.48\linewidth}\centering \textbf{FID to reconstructions}\par\smallskip \includegraphics[width=\linewidth]{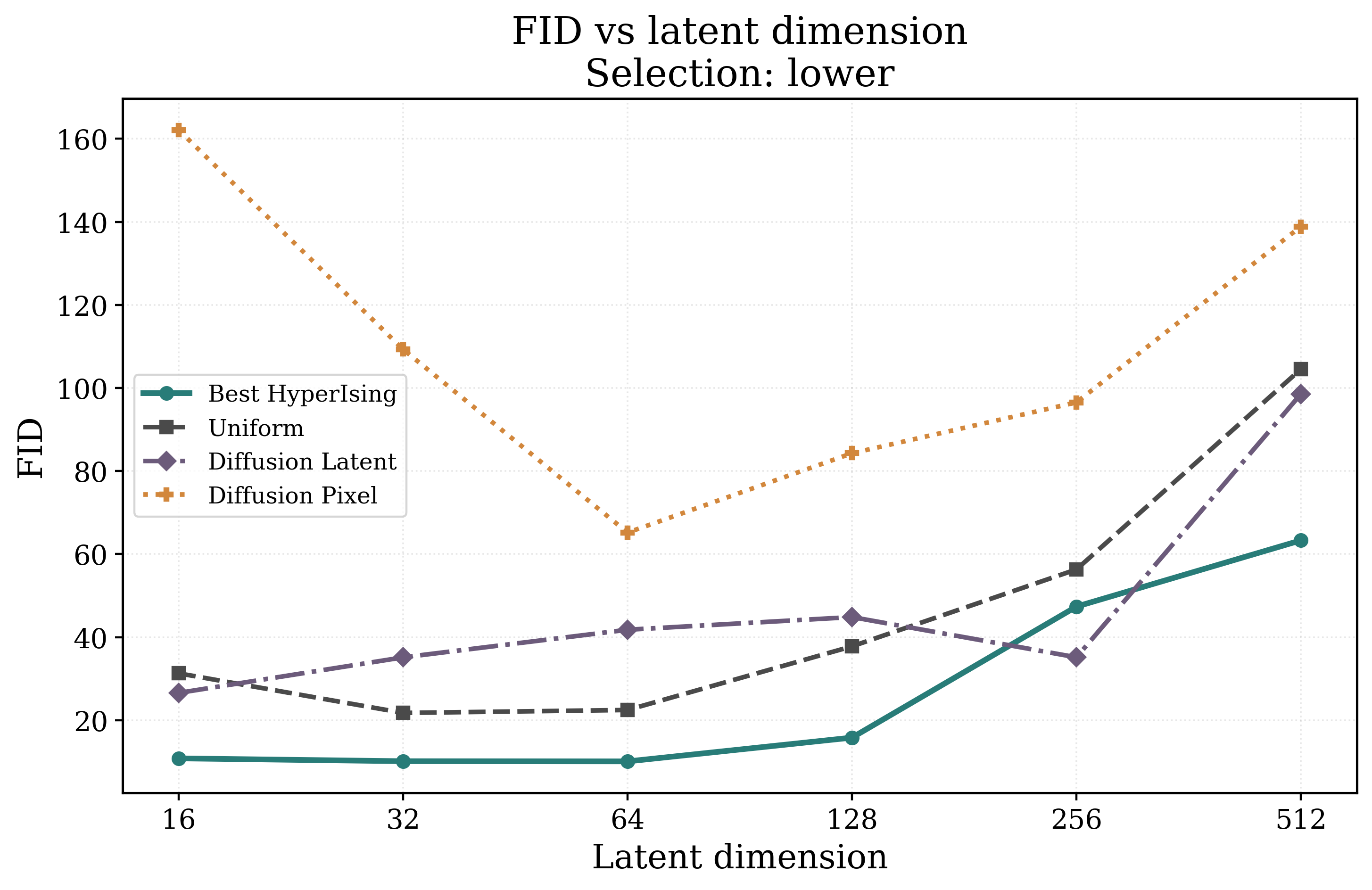}\end{minipage}\par\medskip
\begin{minipage}[t]{.48\linewidth}\centering \textbf{Generative precision to originals}\par\smallskip \includegraphics[width=\linewidth]{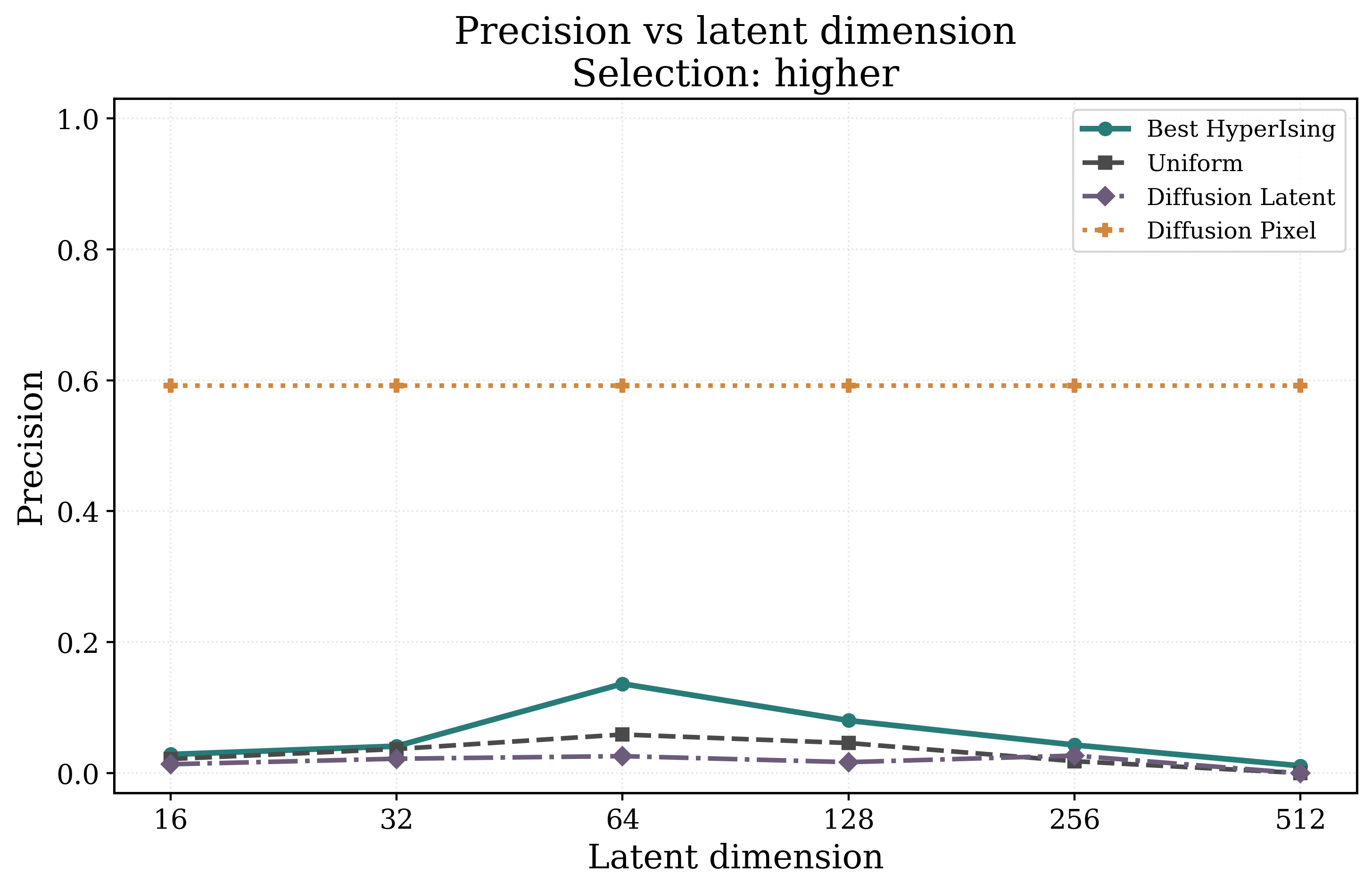}\end{minipage}\hfill \begin{minipage}[t]{.48\linewidth}\centering \textbf{Generative precision to reconstructions}\par\smallskip \includegraphics[width=\linewidth]{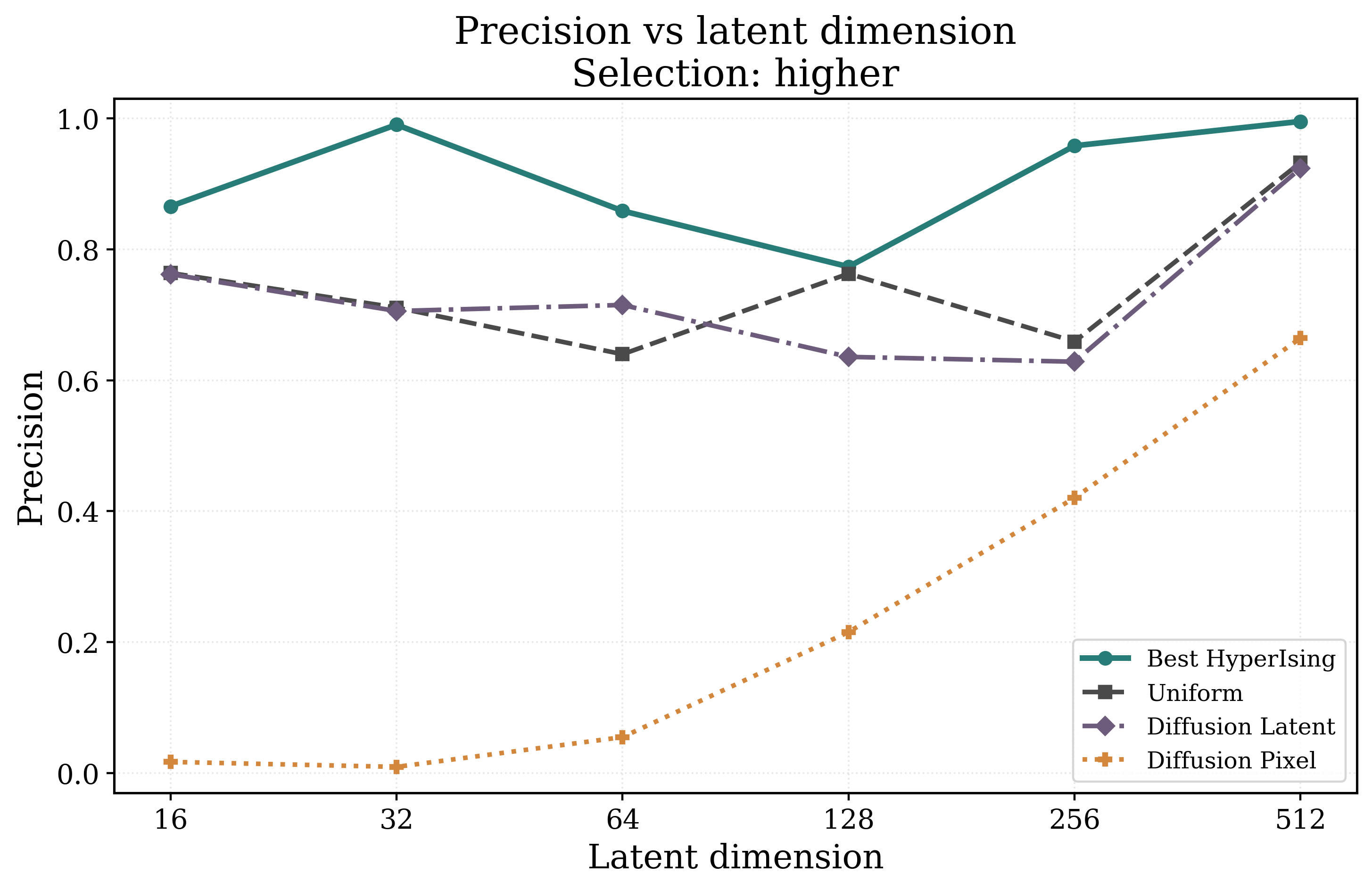}\end{minipage}
\caption{Additional cat-image comparators. Each HyperIsing series is selected independently by the quantity shown. Inception-based FID is lower-is-better; generative precision is higher-is-better. Original and reconstruction references are reported separately.}\label{fig:cats-extra-visual}\end{figure}

The FID implementation used in this study uses Inception features \cite{heusel}; it is distinct from the MNIST CNN64 statistic and the cat DINOv2 FLD. The cat generative precision reverses the coverage question in Appendix~\ref{app:metrics}: it asks what fraction of generated features lie within neighborhoods around training-reference features \cite{recall,fldcode}. Comparing both original and reconstructed references makes explicit how a different target changes the apparent quality of a generated set (Fig.~\ref{fig:cats-extra-visual}).

\begin{figure}[!tbp]\centering
\begin{minipage}[t]{.48\linewidth}\centering \textbf{One-spin means}\par\smallskip \includegraphics[width=\linewidth]{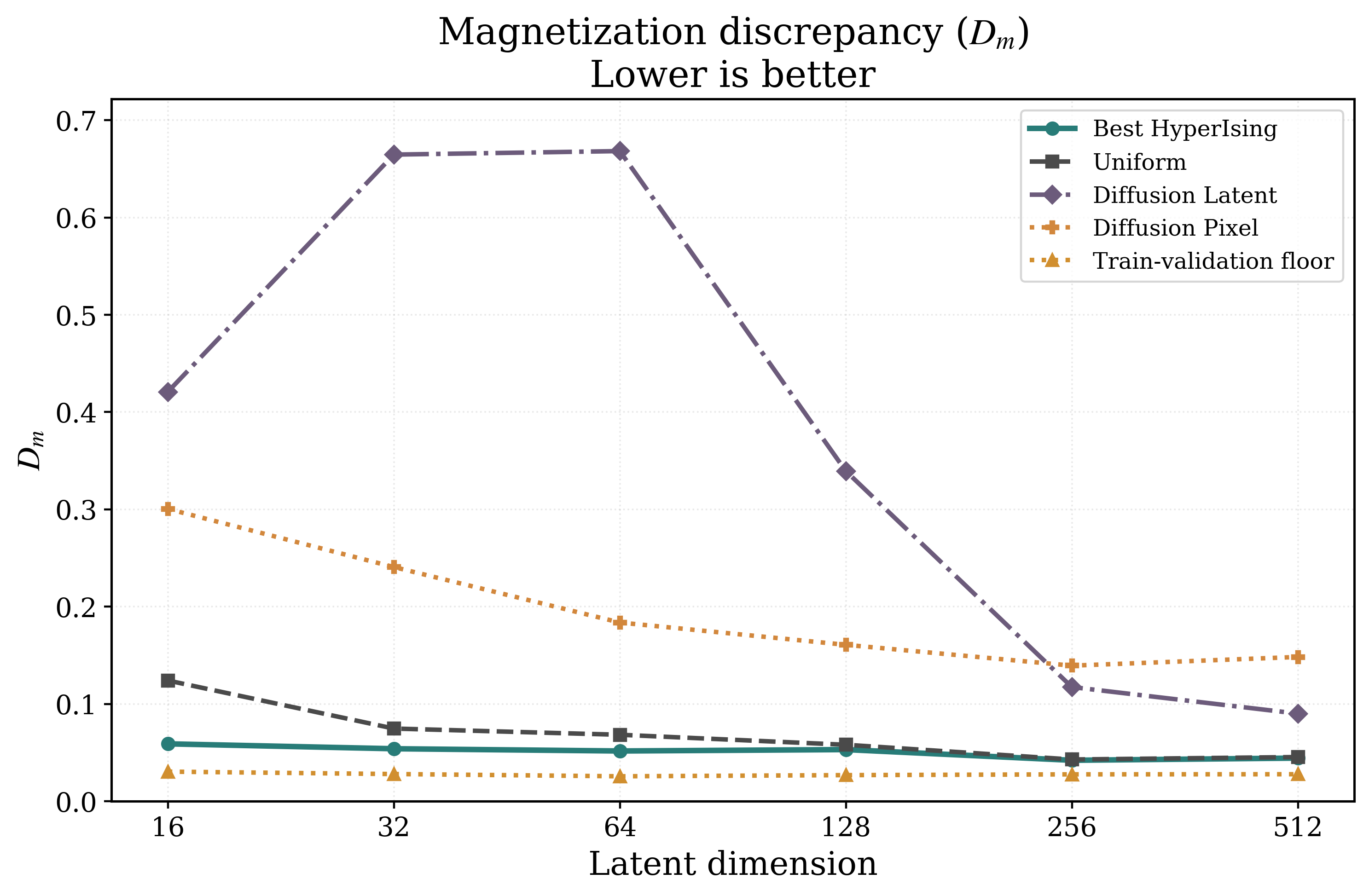}\end{minipage}\hfill \begin{minipage}[t]{.48\linewidth}\centering \textbf{Raw pair moments}\par\smallskip \includegraphics[width=\linewidth]{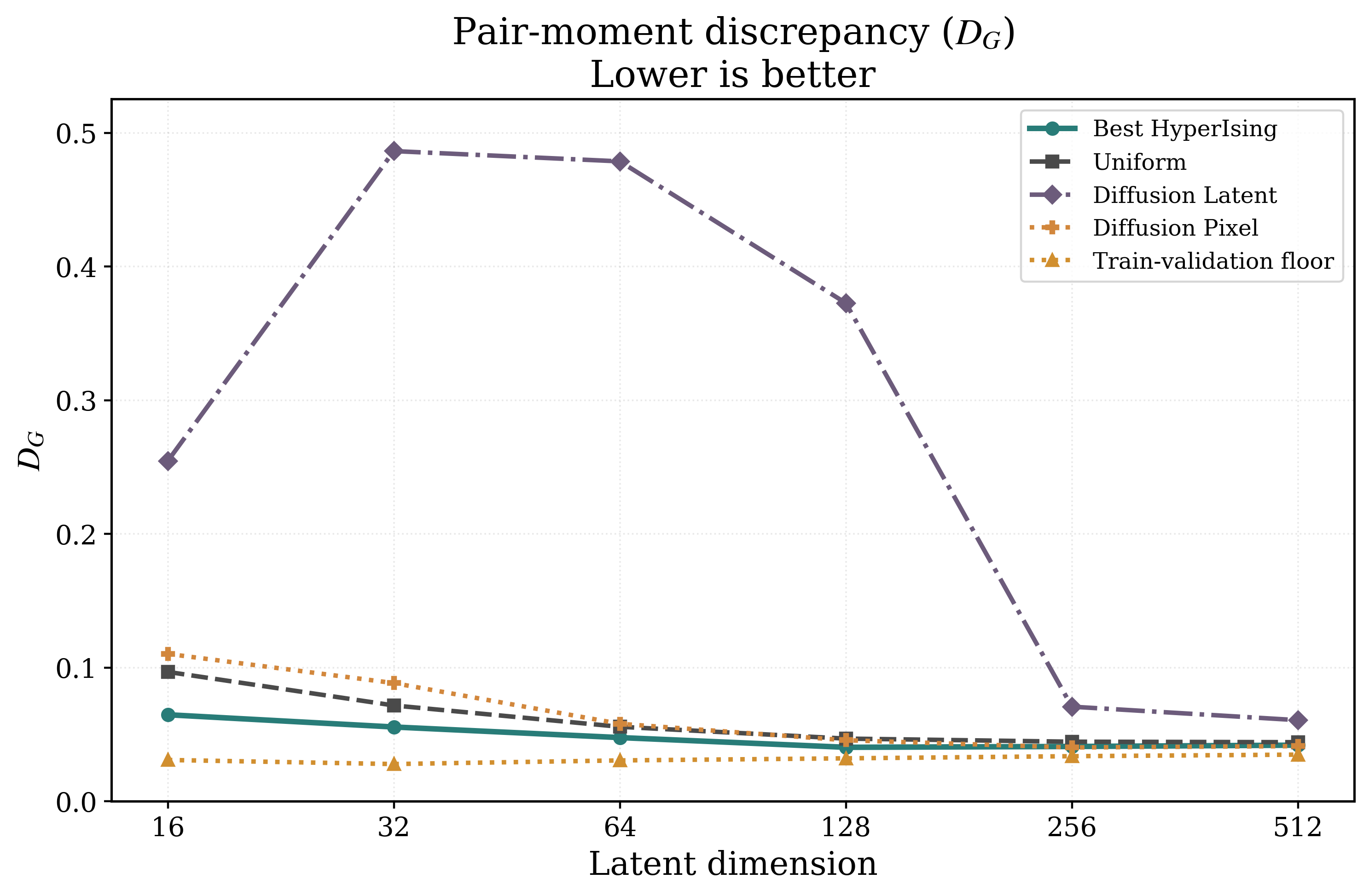}\end{minipage}\par\medskip
\begin{minipage}[t]{.48\linewidth}\centering \textbf{Connected correlations}\par\smallskip \includegraphics[width=\linewidth]{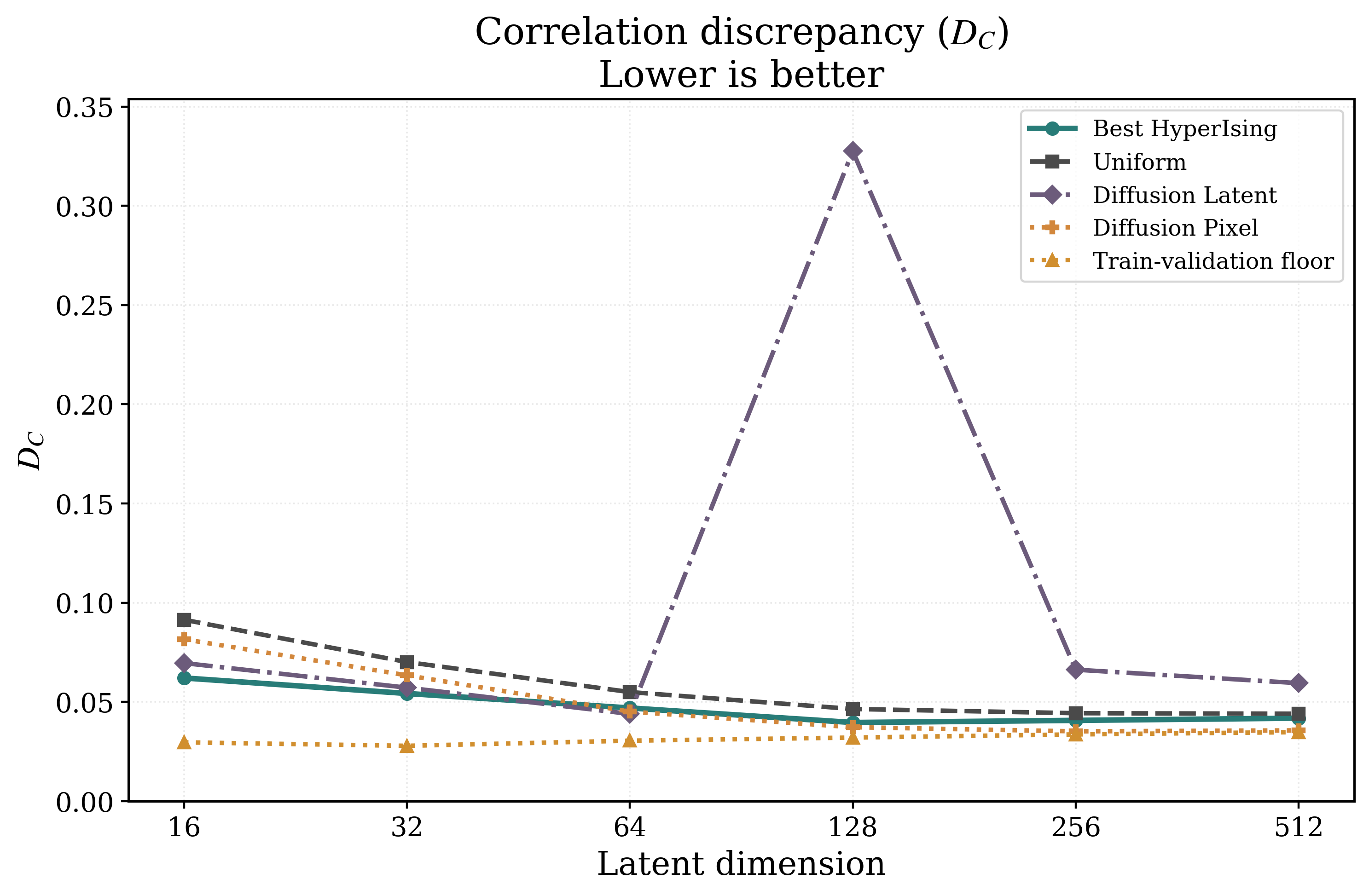}\end{minipage}
\caption{Cat latent-moment components relative to validation codes. The first two quantities enter ILMD; connected correlations are reported separately. Each HyperIsing series uses the best available setting for its own metric. The train--validation contrast is a finite-sample diagnostic.}\label{fig:cats-extra-moments}\end{figure}

\begin{figure}[!tbp]\centering
\begin{minipage}[t]{.48\linewidth}\centering \textbf{4 bits per block}\par\smallskip \includegraphics[width=\linewidth]{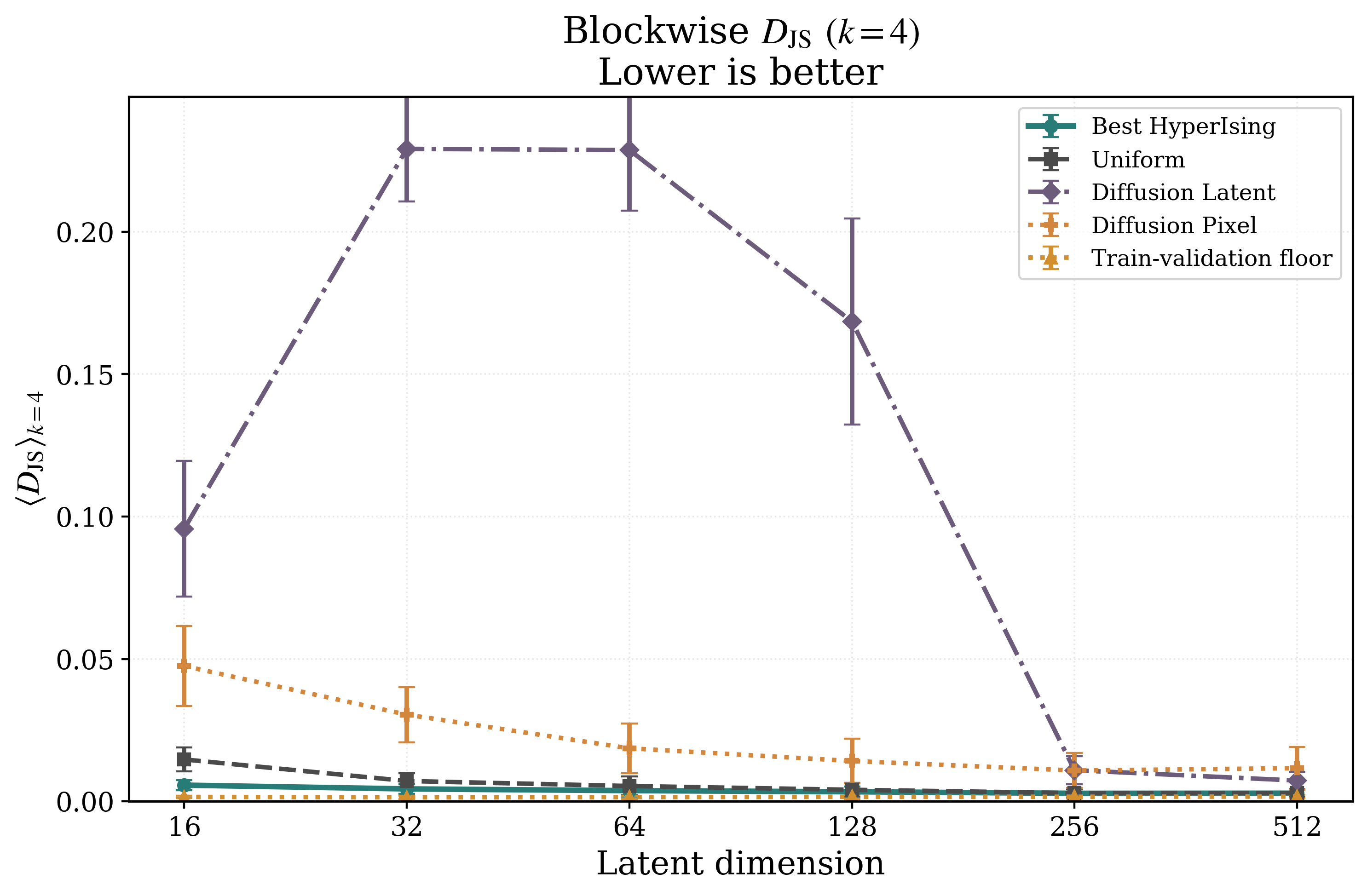}\end{minipage}\hfill \begin{minipage}[t]{.48\linewidth}\centering \textbf{6 bits per block}\par\smallskip \includegraphics[width=\linewidth]{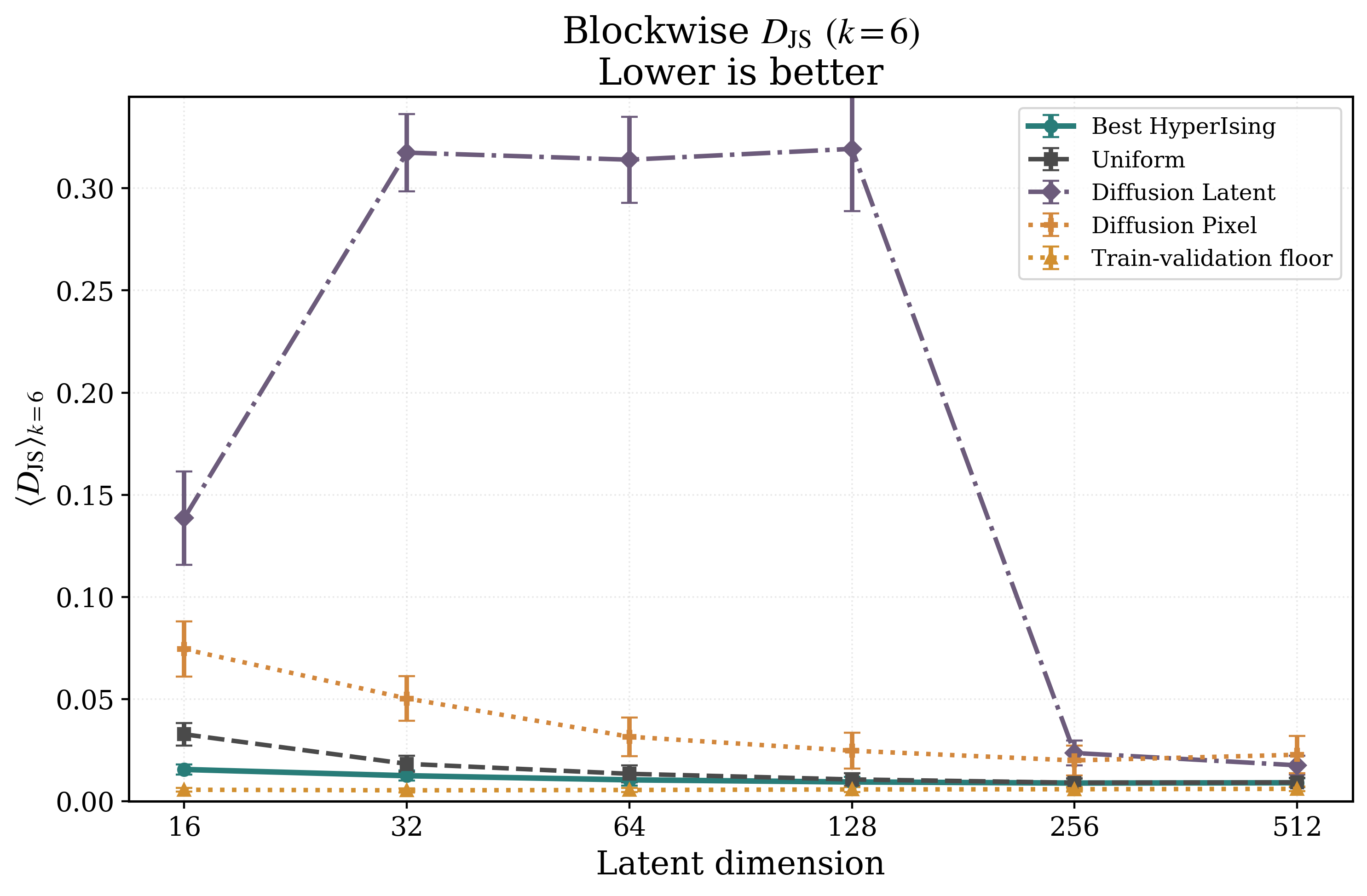}\end{minipage}\par\medskip
\begin{minipage}[t]{.48\linewidth}\centering \textbf{10 bits per block}\par\smallskip \includegraphics[width=\linewidth]{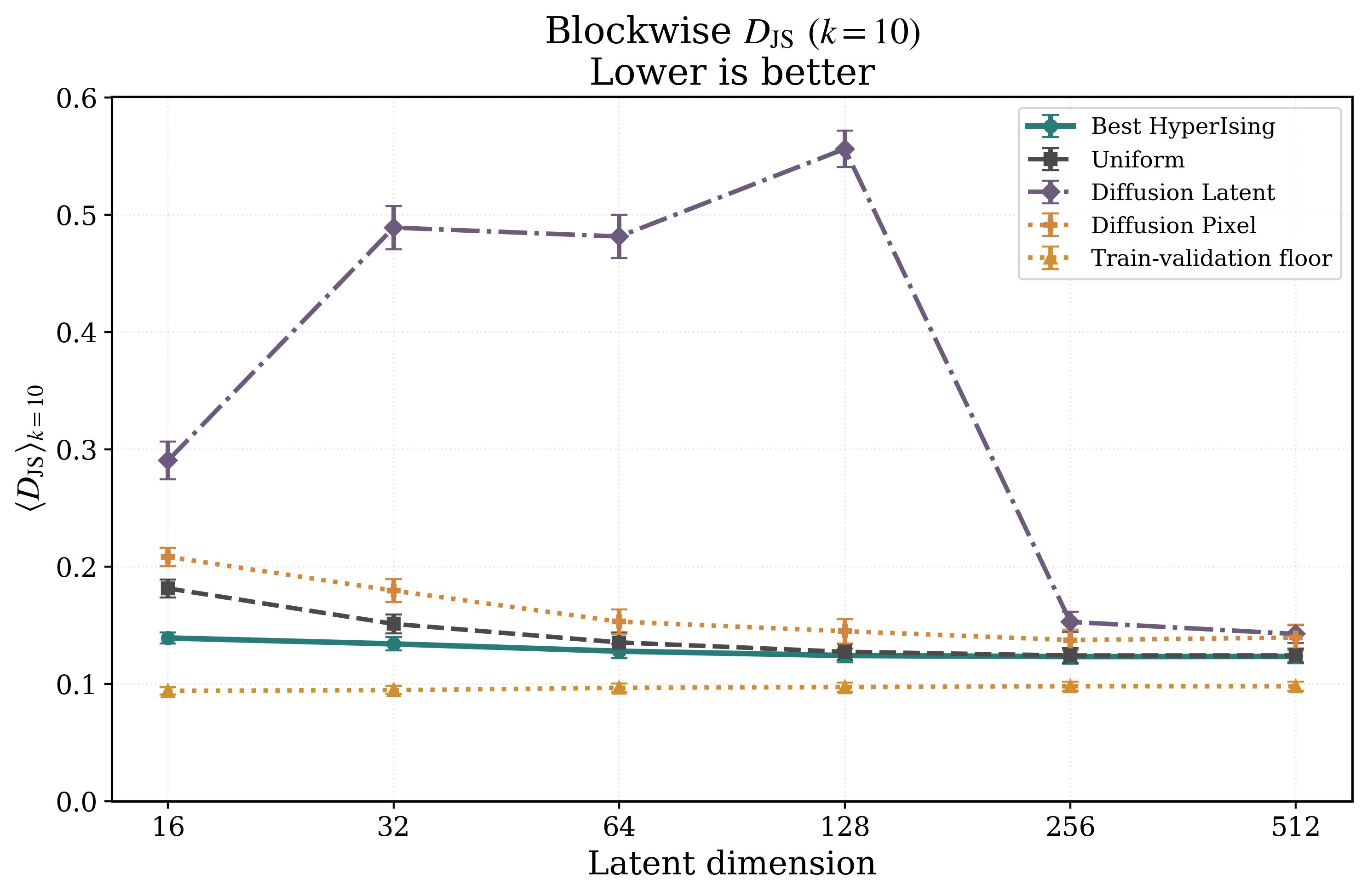}\end{minipage}\hfill \begin{minipage}[t]{.48\linewidth}\centering \textbf{12 bits per block}\par\smallskip \includegraphics[width=\linewidth]{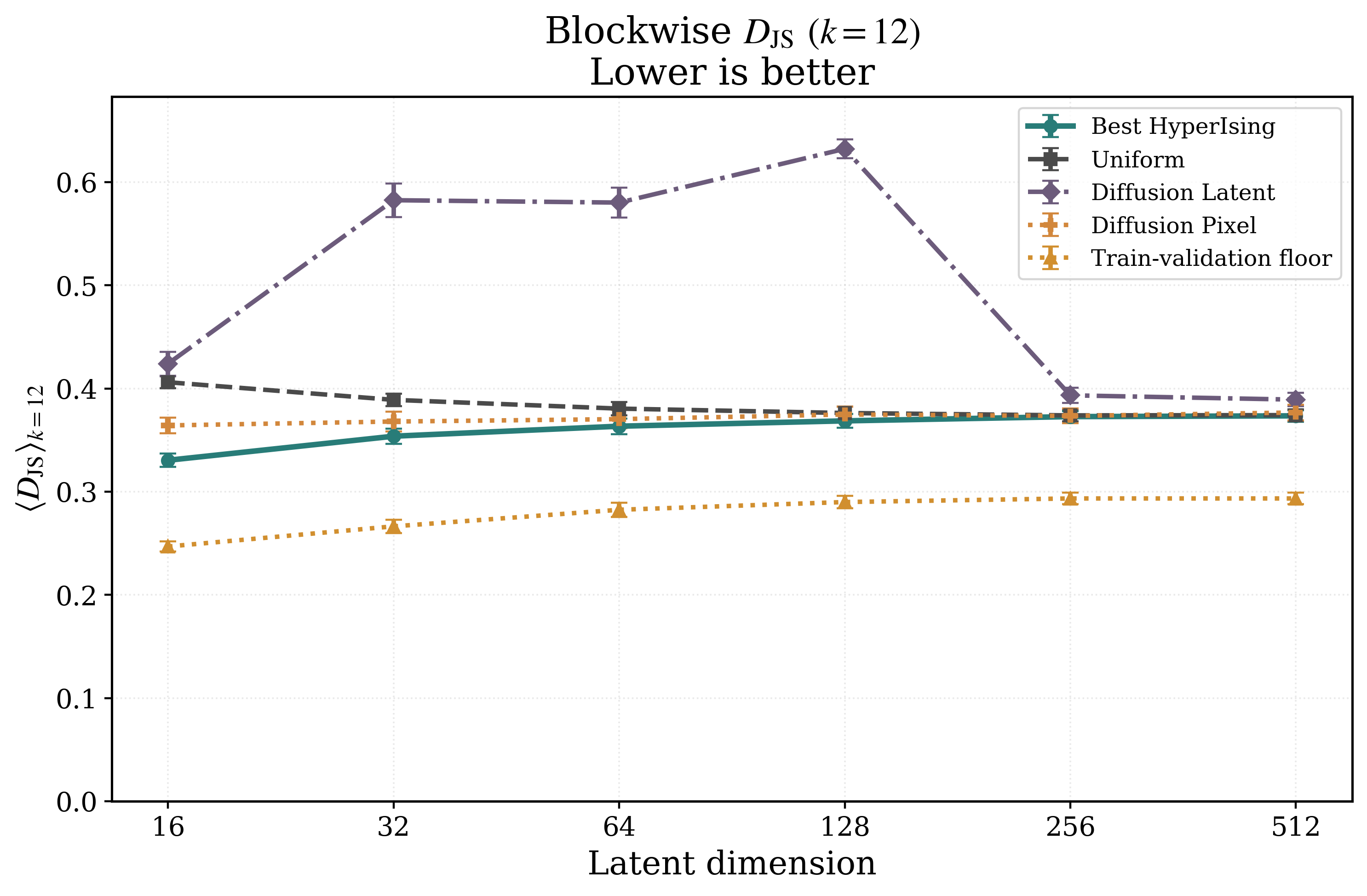}\end{minipage}
\caption{Blockwise Jensen--Shannon comparisons at alternative block sizes. The main text shows eight-bit blocks; these panels show changes in the result and its empirical scale as the block size changes. Error bars summarize variation across the sampled blocks. The selected HyperIsing settings may differ by panel.}\label{fig:cats-extra-js}\end{figure}

\begin{figure}[!tbp]\centering
\begin{minipage}[t]{.48\linewidth}\centering \textbf{Unique-code fraction}\par\smallskip \includegraphics[width=\linewidth]{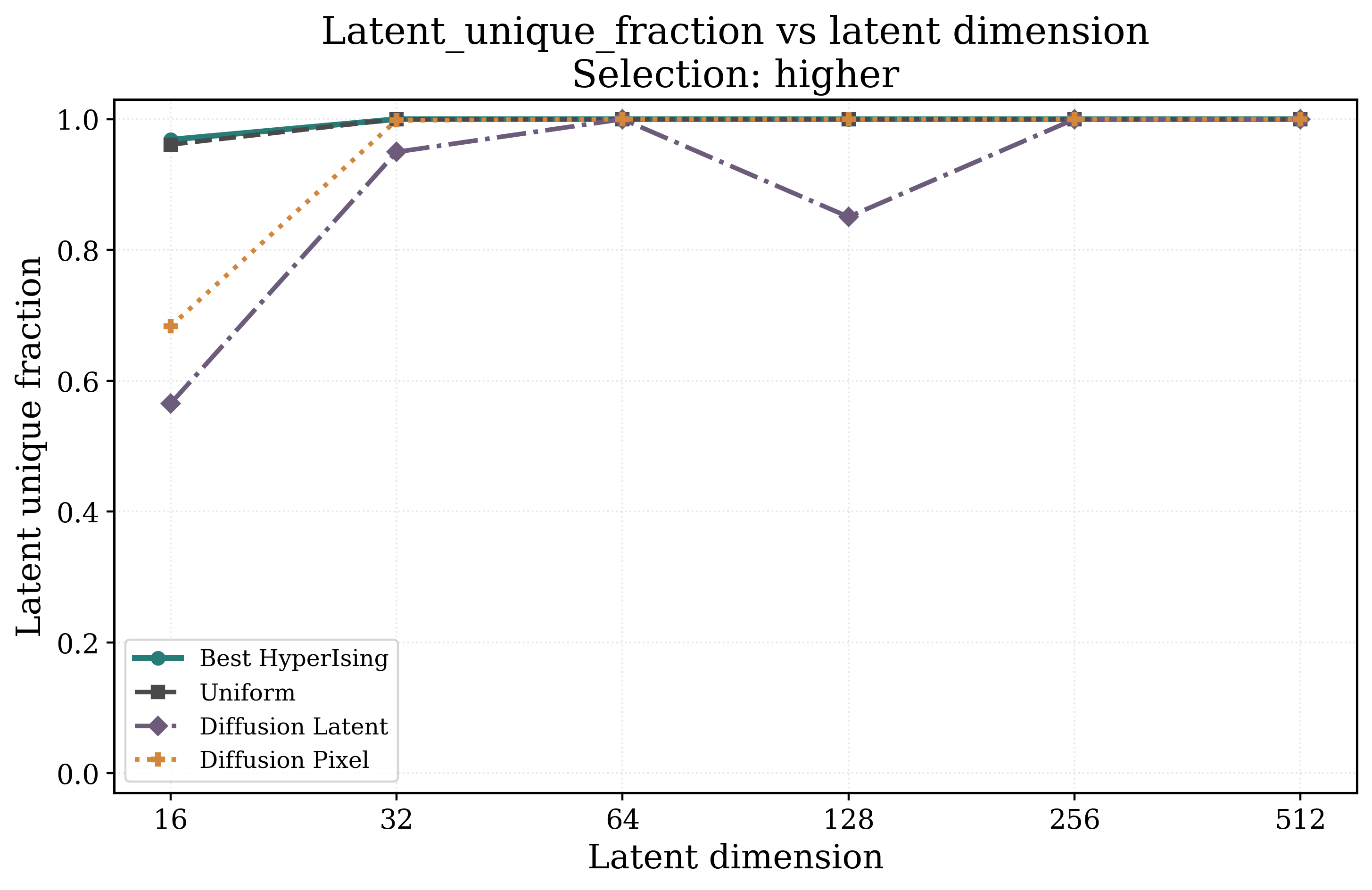}\end{minipage}\hfill \begin{minipage}[t]{.48\linewidth}\centering \textbf{Exact training-code matches}\par\smallskip \includegraphics[width=\linewidth]{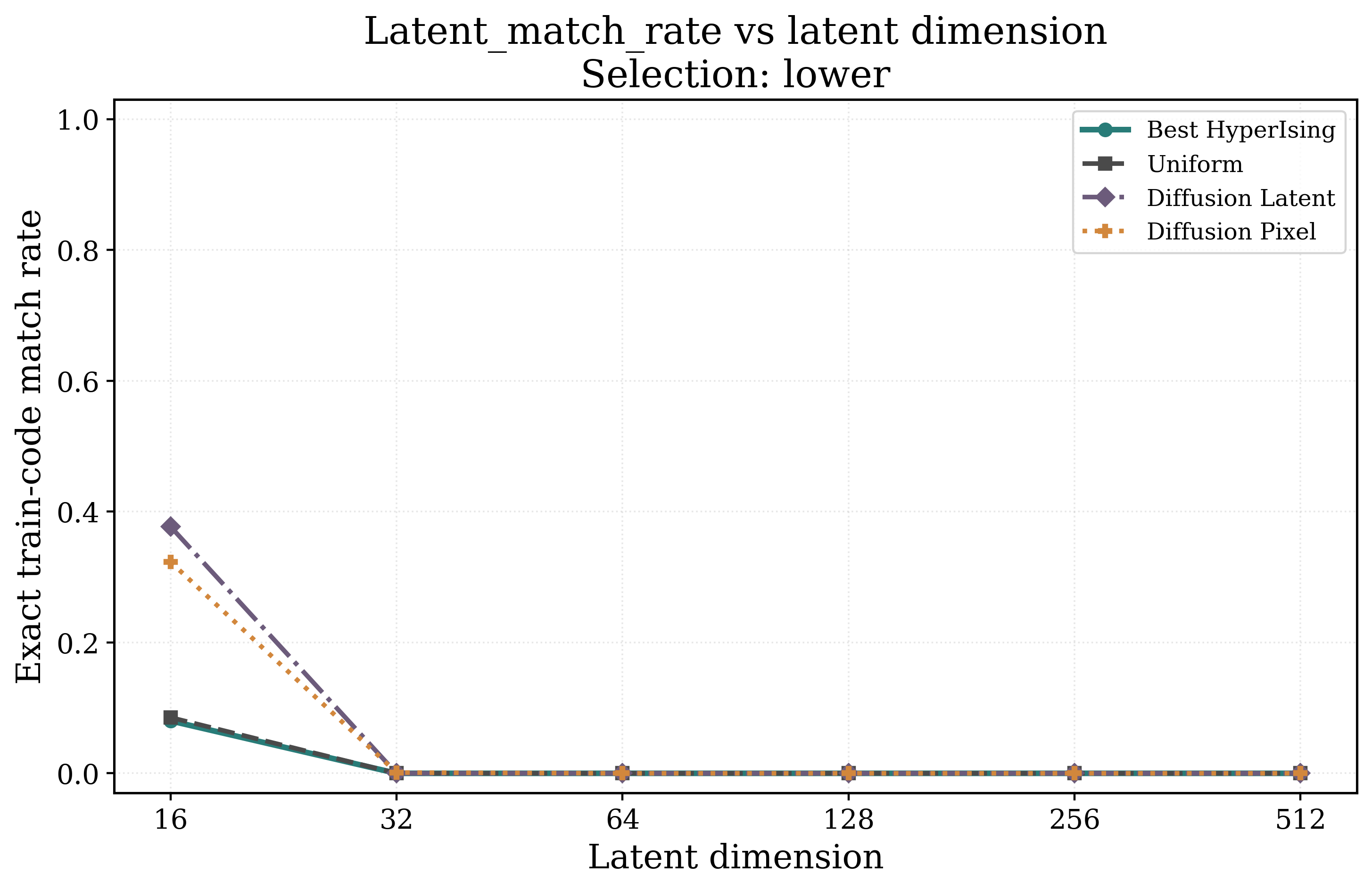}\end{minipage}\par\medskip
\begin{minipage}[t]{.48\linewidth}\centering \textbf{Within-set Hamming distance}\par\smallskip \includegraphics[width=\linewidth]{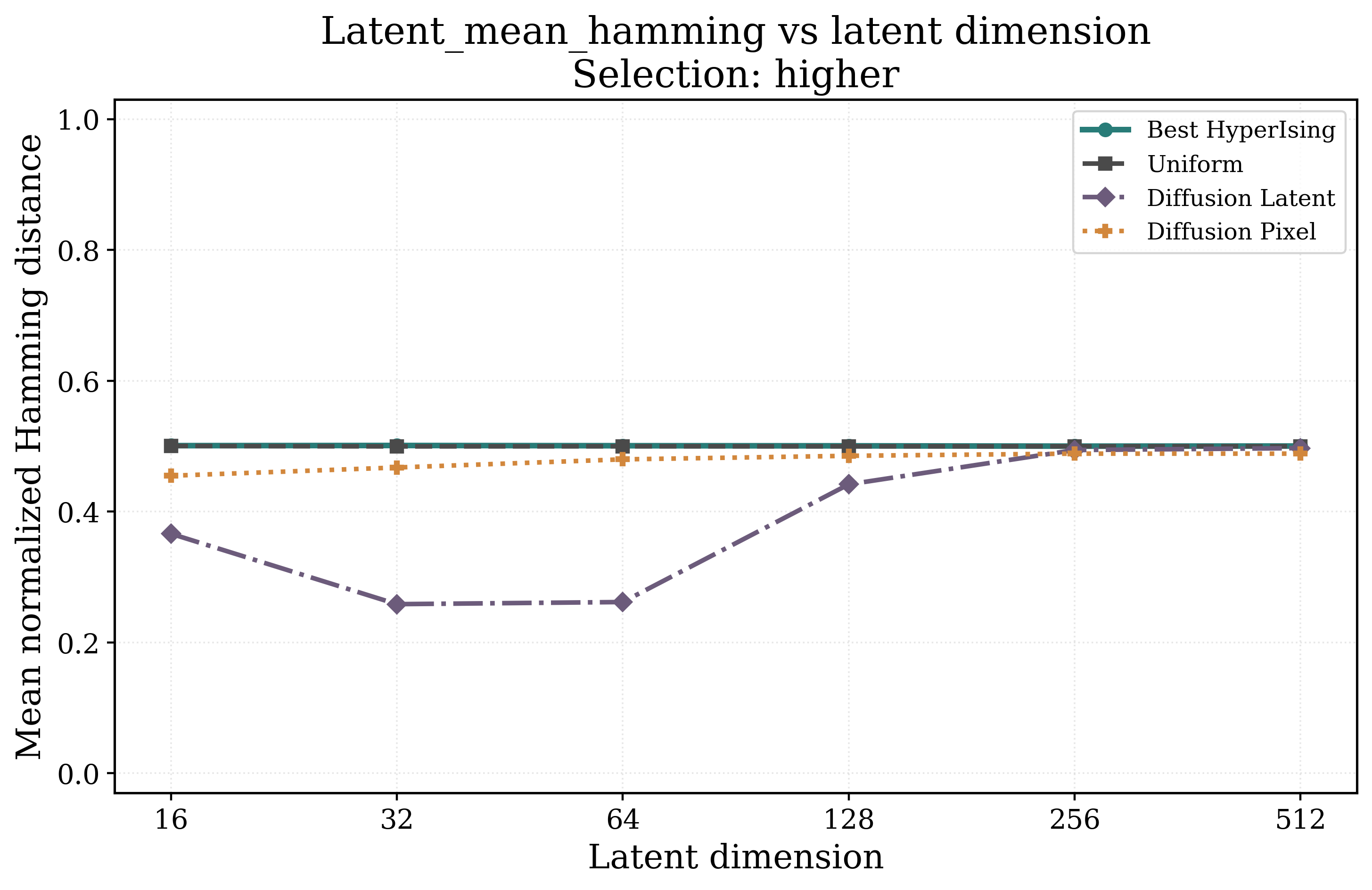}\end{minipage}\hfill \begin{minipage}[t]{.48\linewidth}\centering \textbf{Fraction of all possible codes}\par\smallskip \includegraphics[width=\linewidth]{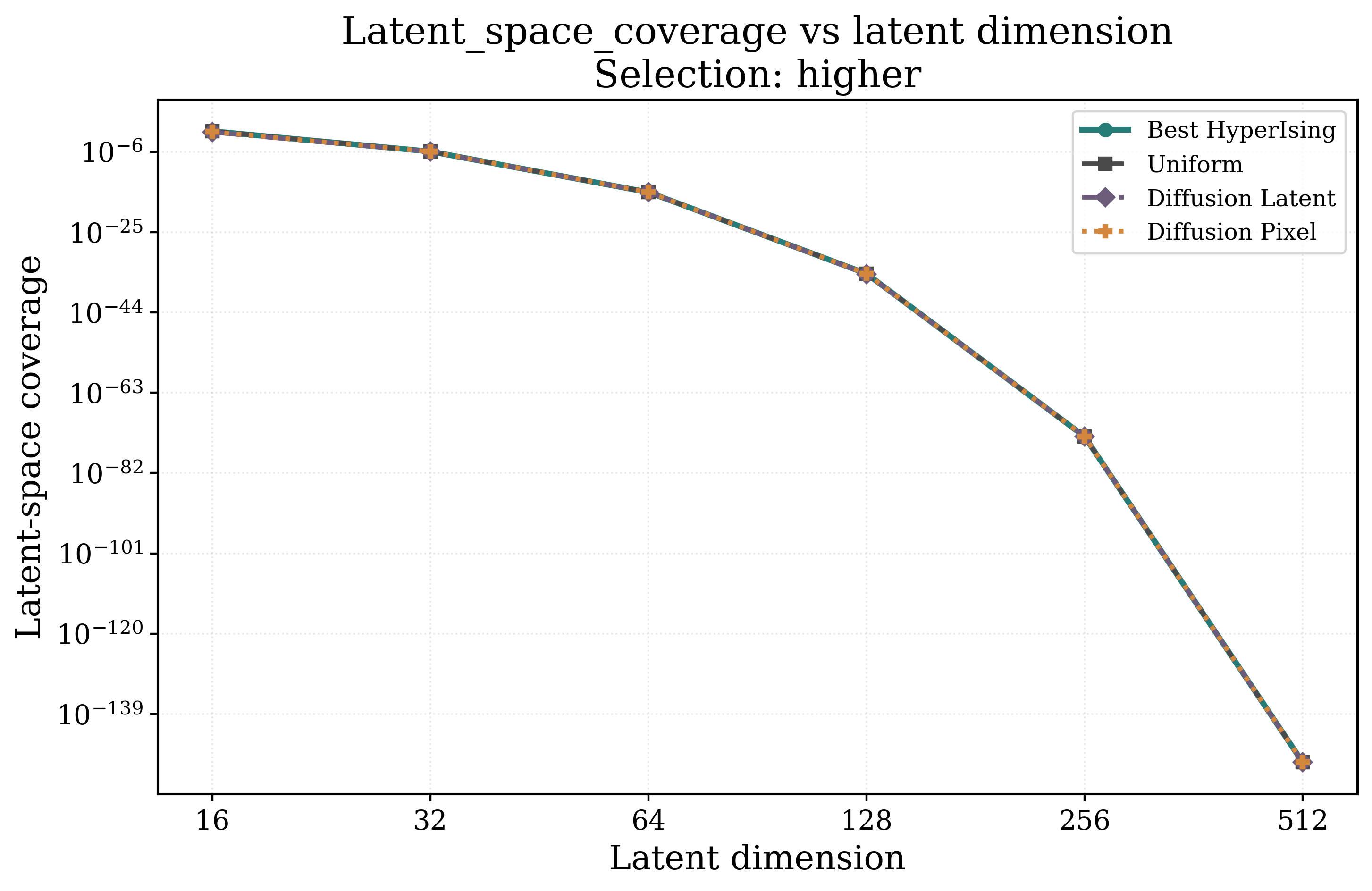}\end{minipage}
\caption{Cat latent-space diagnostics. Unique fraction counts distinct codes per sample; match rate counts generated codes equal to any training code; Hamming is the mean normalized bit difference between sampled pairs of generated codes; space fraction counts observed distinct codes divided by $2^n$. HyperIsing settings are chosen per panel.}\label{fig:cats-extra-diversity}\end{figure}

Figure~\ref{fig:cats-extra-diversity} reports distinct sample counts, exact training-code matches, and observed fractions of binary support. The support fraction divides the number of distinct sampled codes by $2^n$, giving a denominator that grows exponentially with code length.

\FloatBarrier
\subsection{Fixed-Sweep Temperature Scan at 512 Bits}
\begin{figure}[!tbp]\centering
\includegraphics[width=\linewidth,height=.40\textheight,keepaspectratio]{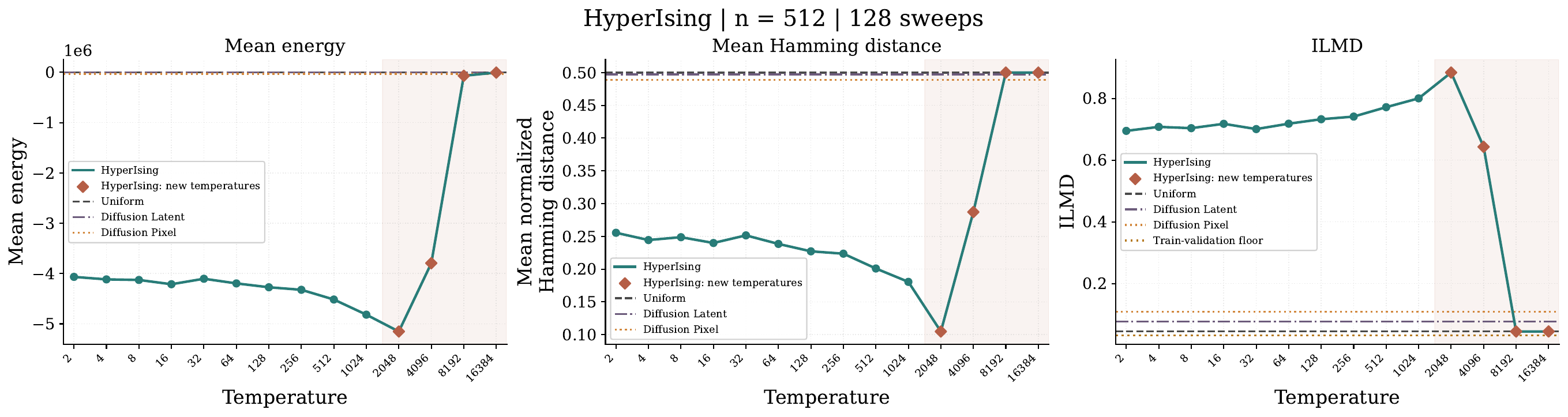}
\caption{Fixed 128-sweep scan at 512 latent bits. Panels report mean energy, mean normalized pairwise Hamming distance between generated codes, and ILMD against validation codes. Distinct markers identify the extension beyond the earlier temperature range.}
\label{fig:cats-fixed512}\end{figure}

\begin{figure}[!tbp]\centering
\includegraphics[width=\linewidth,height=.84\textheight,keepaspectratio]{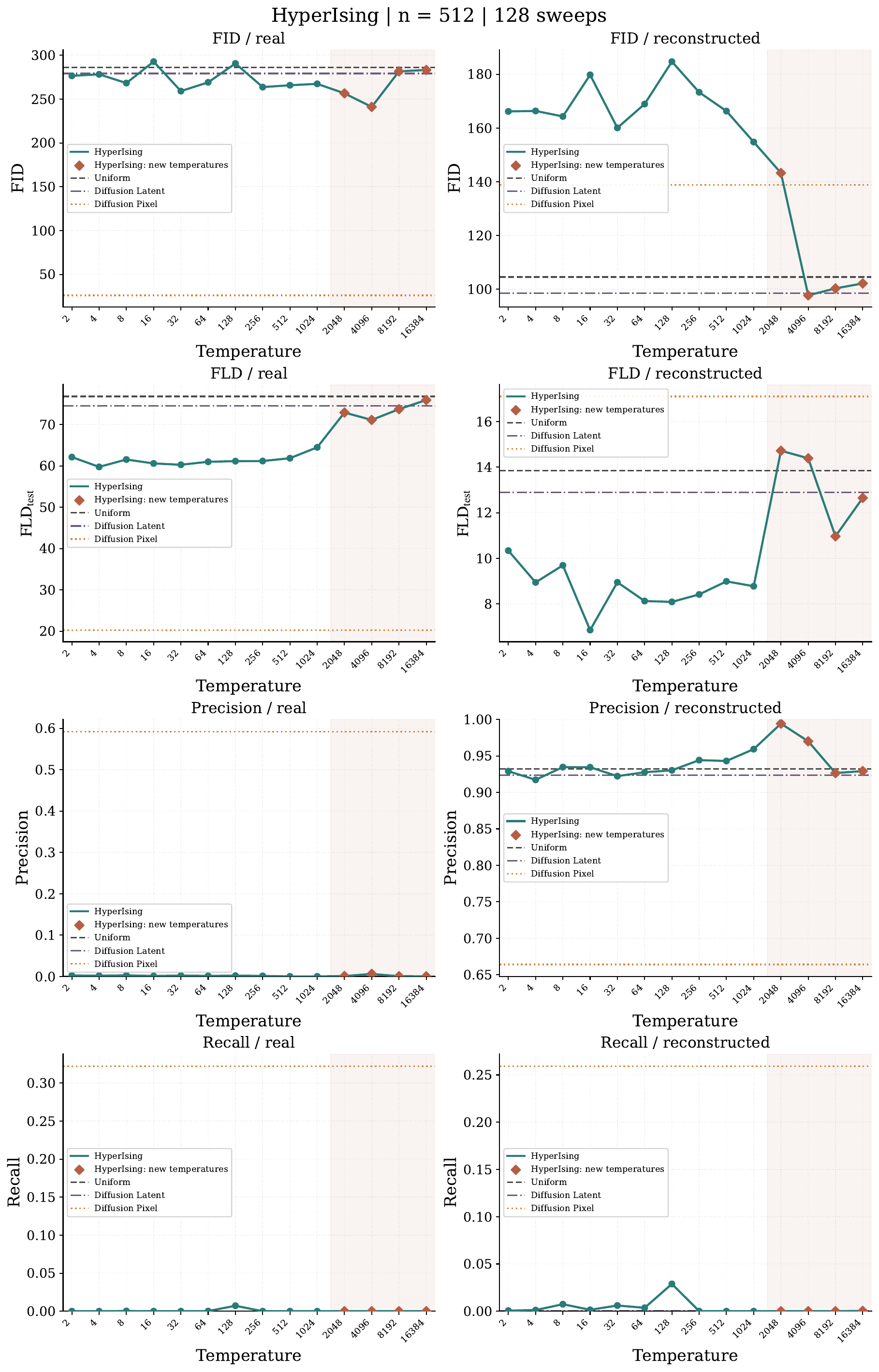}
\caption{Image-space results for the same fixed 128-sweep scan. Columns separate original images from autoencoder reconstructions; rows show Inception FID, validation FLD, generative precision, and generative recall. Lower is better for FID and FLD, and higher is better for precision and recall.}
\label{fig:cats-fixed512-visual}\end{figure}

Figures~\ref{fig:cats-fixed512} and~\ref{fig:cats-fixed512-visual} show latent and image-feature metrics over the 512-bit temperature scan at 128 sweeps.

\FloatBarrier

\FloatBarrier
\end{document}